\documentclass[12pt,amsmath,amssymb,longbibliography,nofootinbib,showkeys,eprint]{revtex4-2}
\usepackage[activate={true,nocompatibility},final,tracking=true,kerning=true,spacing=true]{microtype}
\usepackage{filecontents} 
\usepackage{epigraph}
\usepackage{graphicx}
\usepackage{epstopdf}
\usepackage{braket}
\usepackage{bm}
\usepackage{numprint}
\usepackage{float}
\usepackage[T1,T2A]{fontenc}
\usepackage[utf8]{inputenc}
\usepackage{xfrac}
\usepackage{amsthm}
\theoremstyle{definitions}

\usepackage{makecell}
\usepackage{tikz-timing}
\usepackage{circuitikzgit}
\usepackage{dsfont}
\usepackage{calc}
\usepackage{seqsplit}
\usepackage{hyperref}

\allowdisplaybreaks
\begin{document}
\preprint{V.M.}
\title{On The Radon--Nikodym Spectral Approach With Optimal Clustering}
\author{Vladislav Gennadievich \surname{Malyshkin}} 
\email{malyshki@ton.ioffe.ru}
\affiliation{Ioffe Institute, Politekhnicheskaya 26, St Petersburg, 194021, Russia}

\date{May, 31, 2019}

\begin{abstract}
\begin{verbatim}
$Id: RNSpectralMachineLearning.tex,v 1.717 2021/09/12 15:32:56 mal Exp $
\end{verbatim}
Problems of interpolation, classification, and clustering are considered.
In the tenets of Radon--Nikodym approach
$\Braket{f(\mathbf{x})\psi^2}/\Braket{\psi^2}$,
where the $\psi(\mathbf{x})$ is a linear function on input attributes,
all the answers are obtained
from a generalized eigenproblem $\Ket{f|\psi^{[i]}}=\lambda^{[i]}\Ket{\psi^{[i]}}$.
The solution to
the \hyperref[RNfsolutionpsi]{interpolation} problem 
is a regular Radon--Nikodym derivative.
The solution to the \hyperref[RNWfsolutionpsi]{classification} problem 
requires \hyperref[Pprior]{prior} and \hyperref[Pposterior]{posterior}
probabilities that are obtained using
the Lebesgue quadrature\cite{ArxivMalyshkinLebesgue} technique.
Whereas in a Bayesian approach new observations
change only outcome probabilities,
in the Radon--Nikodym approach not only 
 outcome probabilities
but also the probability space $\Ket{\psi^{[i]}}$
change with new observations.
This is a remarkable feature
of the approach: \textbf{both} the probabilities and the probability space
are constructed from the data.
The Lebesgue quadrature technique can be also applied to
the \hyperref[psiGclustersOpyatAZavor1e6]{optimal clustering}
problem. The problem
is solved by constructing
a Gaussian quadrature on the Lebesgue measure.
A distinguishing feature
of the Radon--Nikodym approach is the knowledge of the
invariant group:
all the answers are invariant relatively any
non--degenerated linear transform of input vector $\mathbf{x}$ components.
A software product implementing the algorithms
of interpolation, classification, and optimal clustering
\href{http://www.ioffe.ru/LNEPS/malyshkin/code_polynomials_quadratures.zip}{is available}
from the authors.

\end{abstract}

\maketitle

\newpage
\section{\label{intro}Introduction}

In our previous work\cite{ArxivMalyshkinLebesgue}
the concept of Lebesgue Integral Quadrature
was introduced and
subsequently
applied to the
problem of joint probability estimation\cite{ArxivMalyshkinJointDistribution}.
In this paper a different application of the Lebesgue Integral Quadrature is developed.
Consider a problem
where attributes vector $\mathbf{x}$ of $n$ components
is mapped to a single outcome $f$ (class label in ML) for $l=[1\dots M]$ observations:
\begin{align}
  (x_0,x_1,\dots,x_k,\dots,x_{n-1})^{(l)}&\to f^{(l)} & \text{weight $\omega^{(l)}$}  \label{mlproblem}
\end{align}
The data of this format is commonly available in practice.
There is a number of problems
of interest, e.g.:
\begin{itemize}
\item For a continuous attribute $f$ build optimal $\lambda_f^{[m]}$; $m=0\dots D-1$ discretization levels,
  a \href{https://en.wikipedia.org/wiki/Discretization_of_continuous_features}{discretization of continuous features} problem.
\item For a discrete $f$: construct a $f$--predictor for a given $\mathbf{x}$ input vector,
  \href{https://en.wikipedia.org/wiki/Statistical_classification}{statistical classification problem}, that  arise in ML, statistics, etc.
For a continuous $f$: predict it's value for a given $\mathbf{x}$.
\item For a given $\mathbf{x}$ estimate the support of the measure in (\ref{mlproblem}) problem,
  in the simplistic formulation it is: find the number of observations that are ``close enough''
  to a given $\mathbf{x}$. Find the $\mathrm{Coverage}(\mathbf{x})$.
  The Christoffel function is often used as a proxy for
  the coverage\cite{2015arXiv151107085G,lasserre2019empirical,beckermann2018perturbations},
  however a genuine  $\mathrm{Coverage}(\mathbf{x})$
  is a very important characteristics in ML.
\item
  Cluster the (\ref{mlproblem}) dataset according to $f$ separability
  (allocate $D\le n$ linear combinations $\psi_G^{[m]}(\mathbf{x})=\sum_{k=0}^{n-1}\alpha^{[m]}_k x_k$, $m=0\dots D-1$,
  that optimally separate the $f$
  in terms of $\Braket{f\psi^2}/\Braket{\psi^2}$).
  For a given $\mathbf{x}$ construct the probability
  distribution of $f$ to fall into the found $D$ clusters.
\end{itemize}
Currently used techniques typically construct a norm,
loss function,
penalty function,
\href{https://en.wikipedia.org/wiki/Metric_(mathematics)}{metric},
distance function,
etc. on $f$,
then perform an optimization minimizing the $f$--error according
to the norm chosen,
a typical example is the 
\href{https://en.wikipedia.org/wiki/Backpropagation}{backpropagation}.
The simplest approach of this type is linear
regression, $L^2$ norm  minimization:
\begin{align}
  \Braket{\left[f(\mathbf{x})-f_{LS}(\mathbf{x})\right]^2}&\rightarrow\min\label{norm2regrf}\\
  f_{LS}(\mathbf{x})&=\sum\limits_{k=0}^{n-1}\beta_k x_k \label{regrf}
\end{align}
As we have shown in \cite{malyshkin2015norm,2016arXiv161107386V}
the major drawback of
an approach of this type
is a difficulty to select a ``good''
norm,  this is especially the case for non--Gaussian data with spikes\cite{malyshkin2018spikes,MalMuseScalp}.

\section{\label{RNapproach}Radon--Nikodym Spectral Approach}
The Lebesgue integral quadrature\cite{ArxivMalyshkinLebesgue}
is an extension of Radon--Nikodym concept
of constructing a classifier of $\Braket{f\psi^2}/\Braket{\psi^2}$ form,
where the $\psi(\mathbf{x})$ is a linear function on input attributes,
to build the support weight as a quadratic function on  $x_k$.
It allows to approach many ML problems in a completely new, norm--free way,
this greatly increases practical applicability. The main idea is to convert (\ref{mlproblem}), a sample of $M$ observations, to a set of $n$ eigenvalue/eigenvector pairs,
subject to generalized eigenvalue problem:
\begin{align}
  \Ket{f\Big|\psi^{[i]}}&=\lambda^{[i]}\Ket{\psi^{[i]}} \label{GEVBracket}\\
  \sum\limits_{k=0}^{n-1} \Braket{x_j|f|x_k} \alpha^{[i]}_k &=
  \lambda^{[i]} \sum\limits_{k=0}^{n-1} \Braket{ x_j|x_k} \alpha^{[i]}_k
  \label{GEV} \\ 
 \psi^{[i]}(\mathbf{x})&=\sum\limits_{k=0}^{n-1} \alpha^{[i]}_k x_k
  \label{psiC}
\end{align}
Here and below the $\Braket{\cdot}$ is $M$ observations sample
averaging, for observations with equal weights $\omega^{(l)}=1$.
This is a plain sum:
\begin{subequations}
  \label{matrixfxx}
\begin{align}
\Braket{1}&=\sum\limits_{l=1}^{M}   \omega^{(l)}\label{mudef}\\ 
  F_{jk}&=\Braket{x_j|f|x_k}=\sum\limits_{l=1}^{M} x^{(l)}_jf^{(l)}x^{(l)}_k \omega^{(l)} \label{xfx}\\
  G_{jk}&=\Braket{x_j|x_k}=\sum\limits_{l=1}^{M} x^{(l)}_jx^{(l)}_k \omega^{(l)}    \label{xx}
\end{align}
\end{subequations}
Here and below we assume that Gram matrix $G_{jk}$ is a non--singular.
In case of a degenerated $G_{jk}$, e.g. in case of data redundancy
in (\ref{mlproblem}),
for example a situation when two input attributes are identical
$x_k=x_{k+1}$ for all $l$, a regularization procedure
is required. A regularization algorithm is presented in the Appendix \ref{regularization}.
Below
we consider the matrix $G_{jk}$ to be positively defined (a regularization is already applied).

Familiar $L^2$ least squares minimization (\ref{norm2regrf}) 
regression answer to (\ref{regrf})
is a linear system solution:
\begin{align}
   f_{LS}(\mathbf{x})&=\sum\limits_{j,k=0}^{n-1}x_jG^{-1}_{jk}\Braket{fx_k} \label{regrfsolution}
\end{align}
The Radon--Nikodym answer\cite{2016arXiv161107386V} is:
\begin{align}
  f_{RN}(\mathbf{x})&=\frac{\sum\limits_{j,k,l,m=0}^{n-1}x_jG^{-1}_{jk}F_{kl}G^{-1}_{lm}x_m}
  {\sum\limits_{j,k=0}^{n-1}x_jG^{-1}_{jk}x_k}
  \label{RNfsolution} \\
  1/K(\mathbf{x})&={\sum\limits_{j,k=0}^{n-1}x_jG^{-1}_{jk}x_k} \label{ChristoffelLike}
\end{align}
Here $G^{-1}_{kj}$ is Gram matrix inverse,
the $K(\mathbf{x})$ is a Christoffel--like function.
In case $x_k=Q_k(x)$, where $x$ is a continuous attribute
and $Q_k(x)$ is a polynomial of the degree $k$, 
the $G_{jk}$ and $F_{jk}$ matrices from (\ref{matrixfxx})
are  the $\Braket{Q_j|Q_k}$ and $\Braket{Q_j|f|Q_k}$ matrices
of Refs. \cite{2016arXiv161107386V,ArxivMalyshkinLebesgue},
and the Christoffel function is $1/K(x)=\sum_{j,k=0}^{n-1}Q_j(x)G^{-1}_{jk}Q_k(x)$.
The (\ref{mlproblem}) is a more general form,
the $x_k$ now can be of arbitrary origin,
an important generalization of
previously considered
a polynomial function of a continuous attribute.

The (\ref{GEV}) solution is $n$ pairs $(\lambda^{[i]},\psi^{[i]}(\mathbf{x}))$.
For positively defined $G_{jk}=\Braket{ x_j|x_k}$ the solution exists and is unique.
For normalized $\psi^{[i]}$ we have:
\begin{subequations}
  \begin{align}
    \delta_{ij}&= \Braket{\psi^{[i]}|\psi^{[j]}}
=\sum\limits_{m,k=0}^{n-1}\alpha^{[i]}_m \Braket{ x_m|x_k} \alpha^{[j]}_k
    \label{psinorm}\\
    \lambda^{[i]}\delta_{ij}&=\Braket{\psi^{[i]}|f|\psi^{[j]}}
=\sum\limits_{m,k=0}^{n-1}\alpha^{[i]}_m \Braket{ x_m|f|x_k} \alpha^{[j]}_k
    \label{fpsinorm}
  \end{align}
\end{subequations}
Familiar $L^2$ least squares minimization (\ref{norm2regrf})
regression  answer
and Radon--Nikodym  answers can be written
in $\psi^{[i]}$ basis.
The (\ref{regrfsolutionpsi}), (\ref{RNfsolutionpsi}), and (\ref{ChristoffelLikepsi})
are the (\ref{regrfsolution}), (\ref{RNfsolution}), and (\ref{ChristoffelLike})
written in the $\psi^{[i]}$ basis:
\begin{align}
  f_{LS}(\mathbf{x})&=\sum\limits_{i=0}^{n-1}\lambda^{[i]}\Braket{\psi^{[i]}}\psi^{[i]}(\mathbf{x}) \label{regrfsolutionpsi}\\
  f_{RN}(\mathbf{x})&=\frac{\sum\limits_{i=0}^{n-1}\lambda^{[i]}\left[\psi^{[i]}(\mathbf{x})\right]^2}{\sum\limits_{i=0}^{n-1}\left[\psi^{[i]}(\mathbf{x})\right]^2}  \label{RNfsolutionpsi} \\
  1/K(\mathbf{x})&={\sum\limits_{i=0}^{n-1}\left[\psi^{[i]}(\mathbf{x})\right]^2}
  \label{ChristoffelLikepsi}
\end{align}
The main result of \cite{ArxivMalyshkinLebesgue} is the construction
of the Lebesgue integral quadrature:
\begin{subequations}
  \label{lebesgueQ}
  \begin{align}
    f^{[i]}&=\lambda^{[i]} \label{fiLeb} \\
    w^{[i]}&= \Braket{\psi^{[i]}}^2 \label{wiLeb} \\
    \Braket{1}&=\sum\limits_{i=0}^{n-1}w^{[i]} \label{sumw} \\
    n&=\sum\limits_{i=0}^{n-1} \Braket{\left[\psi^{[i]}\right]^2} \label{sumw2}
  \end{align}
\end{subequations}  
The Gaussian quadrature groups sums by function argument; it can be viewed as a $n$--point discrete measure, producing the Riemann integral.
The Lebesgue quadrature groups sums by function value; it can be viewed as a $n$--point discrete distribution with $f^{[i]}$ support points (\ref{fiLeb})
and the weights $w^{[i]}$ (\ref{wiLeb}), producing the Lebesgue integral.
Obtained discrete distribution has the number of support points
equals to the
rank of $\Braket{x_j|x_k}$ matrix,
for non-degenerated basis it is equal to
the dimension $n$ of vector $\mathbf{x}$.
The Lebesgue quadrature is unique, hence the
\href{https://en.wikipedia.org/wiki/Principal_component_analysis#Properties_and_limitations_of_PCA}{principal component}
 spectral decomposition is also unique when written in
the Lebesgue quadrature
basis. Substituting (\ref{regrfsolutionpsi})
to (\ref{norm2regrf}) obtain PCA variation expansion:
\begin{align}
  \Braket{\left[f(\mathbf{x})-f_{LS}(\mathbf{x})\right]^2}&=
  \Braket{f^2}-\sum\limits_{i=0}^{n-1}\left(f^{[i]}\right)^2w^{[i]}
  =\Braket{\left(f-\overline{f}\right)^2}
  -\sum\limits_{i=0}^{n-1}\left(f^{[i]}-\overline{f}\right)^2w^{[i]}
  \label{evexpansionStdev}
\end{align}
Here  $\overline{f}={\Braket{f}}/{\Braket{1}}$.
The difference between (\ref{evexpansionStdev})
and regular principal components is that the basis $\Ket{\psi^{[i]}}$
(\ref{GEV})
of the Lebesgue quadrature is \textsl{unique}.
This removes the major limitation of the principal components method:
it's dependence
on the scale of $\mathbf{x}$ attributes.
The (\ref{evexpansionStdev})
does not require scaling and normalizing of input $\mathbf{x}$,
e.g. if $x_k$ attribute is a temperature in Fahrenheit,
when it is converted to Celsius or Kelvin --- the (\ref{evexpansionStdev}) expansion will
be identical. Due to (\ref{GEV}) invariance the variation expansion (\ref{evexpansionStdev})
will be the same for
arbitrary non--degenerated linear transform of $\mathbf{x}$ components:
$x^{\prime}_{j}=\sum_{k=0}^{n-1}T_{jk} x_k$.

In the basis of the Lebesgue quadrature Radon--Nikodym
derivative expression (\ref{RNfsolutionpsi})
is the eigenvalues weighted with (\ref{distY}) weights.
Such a solution is natural for interpolation type
of problem, however for a classification problem different
weights should be used.

\subsection{\label{NashOtvetBayesy}Prior and Posterior Probabilities}
Assume that in  (\ref{RNfsolutionpsi})
for some $\mathbf{x}$ only a single eigenfunction
$\psi^{[i]}(\mathbf{x})$ is non--zero,
then (\ref{RNfsolutionpsi}) gives the corresponding $f^{[i]}$
regardless the weigh $w^{[i]}$.
This is the proper approach to an interpolation problem,
where the $f$ is known to be
a deterministic function on $\mathbf{x}$.
When considering $f$ as random variable, a more reasonable approach
is to classify the outcomes according to overall weight.
Assume no information on $\mathbf{x}$ is available,
what is the best answer for estimation of outcomes probabilities of $f$ ?
The answer is given by the
\href{https://en.wikipedia.org/wiki/Prior_probability}{prior} probabilities
(\ref{Pprior}) that correspond to unconditional distribution of $f$ according
to (\ref{wiLeb}) weights.
\begin{subequations}
  \label{PriorPosterior}
\begin{align}
  \text{Prior weight for $f^{[i]}$:}\quad&  w^{[i]} \label{Pprior} \\
  \text{Posterior weight for $f^{[i]}$:}
   \quad &  w^{[i]}\mathrm{Proj}^{[i]}(\mathbf{x})= w^{[i]}
   \frac{\left[\psi^{[i]}(\mathbf{x})\right]^2}
        {\sum\limits_{j=0}^{n-1}\left[\psi^{[j]}(\mathbf{x})\right]^2}
        \label{Pposterior}
\end{align}
\end{subequations}
The  \href{https://en.wikipedia.org/wiki/Posterior_probability}{posterior}
distribution uses the same $\left[\psi^{[i]}(\mathbf{x})\right]^2$
probability as (\ref{RNfsolutionpsi})
adjusted to $f^{[i]}$  outcome prior weight $w^{[i]}$. The corresponding average
\begin{align}
  f_{RNW}(\mathbf{x})&=
  \frac{\sum\limits_{i=0}^{n-1}\lambda^{[i]}w^{[i]}\mathrm{Proj}^{[i]}(\mathbf{x})}
       {\sum\limits_{i=0}^{n-1}w^{[i]}\mathrm{Proj}^{[i]}(\mathbf{x})}
       =
   \frac{\sum\limits_{i=0}^{n-1}\lambda^{[i]}w^{[i]}\left[\psi^{[i]}(\mathbf{x})\right]^2}{\sum\limits_{i=0}^{n-1}w^{[i]}\left[\psi^{[i]}(\mathbf{x})\right]^2}
  \label{RNWfsolutionpsi}
\end{align}
is similar to (\ref{RNfsolutionpsi}), but uses the
posterior weights (\ref{Pposterior}). There are 
two distinctive cases of $f$ on $\mathbf{x}$ inference:
\begin{itemize}
  \item
    If $f$ is a deterministic function on $\mathbf{x}$,
    such as in an interpolation problem,
    then the probabilities of $f$ outcomes are not important,
    the only important characteristic is:
    how large is $\Ket{\psi^{[i]}}$ eigenvector
    at given $\mathbf{x}$;
    the weight is the $i$--th eigenvector projection (\ref{distY}).
    The best interpolation answer is then (\ref{RNfsolutionpsi}) $f_{RN}(\mathbf{x})$:
    the eigenvalues $\lambda^{[i]}$ weighted with the projections $\mathrm{Proj}^{[i]}(\mathbf{x})$
    as the weights.
  \item If $f$ (or some $x_k$) is a random variable,
    then inference answer depends on the distribution of $f$.
    The classification answer should include not only what the outcome $\lambda^{[i]}$
    corresponds to a given $\mathbf{x}$,
    but also how often the outcome $\lambda^{[i]}$ occurs; this is determined by the prior weights $w^{[i]}$.
    The best answer is then (\ref{RNWfsolutionpsi}) $f_{RNW}(\mathbf{x})$:
    the eigenvalues $\lambda^{[i]}$ weighted with the
    posterior  weights
    $w^{[i]}\mathrm{Proj}^{[i]}(\mathbf{x})$.
    An important characteristic is
    \begin{align}
      \mathrm{Coverage}(\mathbf{x})&=\sum\limits_{i=0}^{n-1}
      w^{[i]}\mathrm{Proj}^{[i]}(\mathbf{x})
      \label{coverage}
    \end{align}
    that is equals to Lebesgue quadrature weights $w^{[i]}$
    weighted with projections.
    For (\ref{lebesgueQ})
    the \href{https://en.wikipedia.org/wiki/Probability_space}{probability space}
    is $n$ vectors $\Ket{\psi^{[i]}}$ with the probabilities $w^{[i]}$.
    The coverage is a characteristic
    of how often given $\mathbf{x}$ occurs in the observations
    (here we assume that total sample space is projected
    to $\Ket{\psi^{[i]}}$ states).
    Entropy $S_f$ of a random variable $f$ can be estimated from prior
    probabilities:
    \begin{align}
    S_f  &= -\sum\limits_{i=0}^{n-1} \frac{w^{[i]}}{\Braket{1}} \ln\left(\frac{w^{[i]}}{\Braket{1}}\right)
    \label{EntropyLikeF}
    \end{align}
    It can be used as a measure of
    \href{https://en.wikipedia.org/wiki/Entropy_(information_theory)#Definition}{statistical dispersion} of $f$. Similarly, conditional entropy $S_{f|x}$
    can be obtained from prior and posterior
    probabilities (\ref{PriorPosterior}):
    \begin{align}
      S_{f|x}  &= -\sum\limits_{i=0}^{n-1}
      \frac{ w^{[i]}\mathrm{Proj}^{[i]}(\mathbf{x})}{\Braket{1}}
      \ln\left(\frac{w^{[i]}\mathrm{Proj}^{[i]}(\mathbf{x})}{\mathrm{Coverage}(\mathbf{x})}\right)
    \label{EntropyLikeFcondX}
    \end{align}
\end{itemize}
    
The $f_{RNW}$ can be interpreted as a Bayes style of answer.
An observation
$\mathbf{x}$
changes outcome probabilities from (\ref{Pprior}) to (\ref{Pposterior}).
Despite all the similarity there is a very important
difference between
\href{https://en.wikipedia.org/wiki/Bayesian_inference}{Bayesian inference}
and Radon--Nikodym approach.
In the Bayesian inference\cite{mosteller2012applied} the probability space
is \textsl{fixed}, new observations can adjust
only the probabilities of pre--set states.
In the Radon--Nikodym approach,
the probability space is the Lebesgue quadrature (\ref{lebesgueQ})
states $\Ket{\psi^{[i]}}$, the solution to (\ref{GEVBracket}) eigenproblem.
This problem is determined by two matrices
$\Braket{x_j|f|x_k}$ and $\Braket{x_j|x_k}$, that depend on the
observation sample themselves. The key difference
is that new observations coming to (\ref{mlproblem})
change not only outcome probabilities,
but also the \textbf{probability space} $\Ket{\psi^{[i]}}$.
This is a remarkable feature
of the approach: \textbf{both} the probabilities and the probability space
are constructed from the data.
For probability space of the Lebesgue quadrature (\ref{lebesgueQ})
this flexibility
allows us to solve the problem of optimal clustering.

\section{\label{BasisReduction}Optimal Clustering}
Considered in previous section two inference answers
(\ref{RNfsolutionpsi}) and (\ref{RNWfsolutionpsi})
use vector $\mathbf{x}$ of $n$ components as input attributes $x_k$.
In a typical ML setup the number
of attributes can grow quite substantially, and for a large
enough $n$ the problem of
\href{https://en.wikipedia.org/wiki/Overfitting}{data overfitting}
is starting to rise. This is especially the case
for norm--minimization approaches
such as (\ref{regrfsolutionpsi}), and is much less so for
Radon--Nikodym type of answer (\ref{RNfsolutionpsi}),
where  the answer  is a linear
superposition of the observed $f$ with \textsl{positive weight} $\psi^2(\mathbf{x})$
(the least squares answer is also a superposition
of the observed $f$, but the weight is not always positive).
However, for large enough $n$ the overfitting problem  also
arises in $f_{RN}$. The Lebesgue quadrature (\ref{lebesgueQ})
builds $n$ cluster centers, for large enough $n$ the (\ref{RNfsolutionpsi})
finds the closest cluster
in terms of $\mathbf{x}$ to $\psi^{[i]}$ distance,
this is the projection $\mathrm{Proj}^{[i]}(\mathbf{y})=\Braket{\psi_{\mathbf{y}}|\psi^{[i]}}^2$ to localized at $\mathbf{x}=\mathbf{y}$
state $\psi_{\mathbf{y}}(\mathbf{x})$:
\begin{align}
  \mathrm{Proj}^{[i]}(\mathbf{x})&=
  \frac{\left[\psi^{[i]}(\mathbf{x})\right]^2}
       {\sum\limits_{j=0}^{n-1}\left[\psi^{[j]}(\mathbf{x})\right]^2}
       =\Braket{\psi_{\mathbf{x}}|\psi^{[i]}}^2
       \label{distY} \\
       1&=\sum\limits_{i=0}^{n-1} \mathrm{Proj}^{[i]}(\mathbf{x})
       =\sum\limits_{i=0}^{n-1}\Braket{\psi_{\mathbf{x}}|\psi^{[i]}}^2
       \label{sim0} \\
       \psi_{\mathbf{y}}(\mathbf{x})&=\frac{\sum\limits_{i=0}^{n-1}\psi^{[i]}(\mathbf{y})\psi^{[i]}(\mathbf{x})}
           {\sqrt{\sum\limits_{i=0}^{n-1}\left[\psi^{[i]}(\mathbf{y})\right]^2}}
           =
           \frac{\sum\limits_{j,k=0}^{n-1}y_jG^{-1}_{jk}x_k}
           {\sqrt{\sum\limits_{j,k=0}^{n-1}y_jG^{-1}_{jk}y_k}}             
  \label{psiYlocalized}
\end{align}
and then uses corresponding $f^{[i]}$ as the result.
Such a special
cluster always exists for large
enough $n$, with $n$ increase the Lebesgue quadrature (\ref{lebesgueQ})
separates the $\mathbf{x}$ space on smaller and smaller
clusters in terms of (\ref{distY}) distance
as the square of wavefunction projection.

In practical applications a hierarchy of dimensions is required.
The number of sample observations $M$ is typically in a $1,000-100,000$ range.
The dimension $n$ of attributes vector $\mathbf{x}$
is at least ten times lower
than the $M$,  $n$ is typically $5-100$. The number of clusters
$D$, required to identify the data is several times lower than the $n$,
$D$ is typically $2-10$; the  $D\le n\le M$ hierarchy must be always held.

The Lebesgue quadrature (\ref{lebesgueQ}) gives us $n$ cluster centers,
the number of input attributes. We need to construct $D\le n$ clusters out of them,
that provide ``the best'' classification for a given $D$.
Even
the 
attributes selection problem (select $D$ ``best'' attributes out of $n$ available $x_k$)
is of combinatorial complexity\cite{maloldarxiv},
and can be solved only heuristically with a various degree of success.
The problem to construct $D$ attributes out of $n$ is even more
complex. The problem
is typically reduced to some optimization problem,
but the difficulty to chose a norm and computational complexity
makes it impractical.

In this paper an original approach is developed.
The reason for our success is the very specific
form of the Lebesgue quadrature weights (\ref{wiLeb})
$w^{[i]}= \Braket{\psi^{[i]}}^2$ that allows us to construct
a $D$--point Gaussian quadrature in $f$-- space,
it provides
the best $D$--dimensional separation of $f$,
and then to
convert obtained solution
to  $\mathbf{x}$ space!

A 
\href{https://en.wikipedia.org/wiki/Gaussian_quadrature}{Gaussian quadrature}
constructs
a set of nodes $f_G^{[m]}$ and weights $w_G^{[m]}$ such that
\begin{align}
  \Braket{g(f)}&\approx\sum\limits_{m=0}^{D-1} g(f_G^{[m]})w_G^{[m]}
  \label{intGaussF}
\end{align}
is exact for $g$ being a polynomial of a degree $2D-1$ or less.
The Gaussian quadrature can be considered as the optimal 
approximation of the distribution of $f$
by a $D$--point discrete measure.
With the measure $\Braket{\cdot}$ in the form of $M$ terms sample sum (\ref{matrixfxx})
no inference of $f$ on $\mathbf{x}$ can be obtained,
we can only estimate the distribution of $f$ (prior probabilities).

Now consider $D$--point
Gaussian quadrature
built on $n$ point discrete measure
of the Lebesgue quadrature (\ref{lebesgueQ}), $D\le n$.
Introduce the measure $\Braket{\cdot}_L$
\begin{align}
  \Braket{g(f)}_L&=\sum\limits_{i=0}^{n-1} g(f^{[i]})w^{[i]} \label{Lmeasure} \\
  \Braket{1}_L&=\Braket{1} \label{measureLmeasurematch}
\end{align}
and build Gaussian quadrature (\ref{intGaussF})
on the Lebesgue measure $\Braket{\cdot}_L$.
Select some polynomials $Q_k(f)$,
providing sufficient numerical stability,
the result is invariant with respect
to basis choice,  $Q_m(f)=f^m$ and $Q_m=T_m(f)$
give \textsl{identical} results, but numerical stability
can be drastically different\cite{beckermann1996numerical,2015arXiv151005510G}.
Then construct two matrices ${\mathcal F}_{st}$ and ${\mathcal G}_{st}$
(in (\ref{QfQg}) and (\ref{QQg}) the $f^{[i]}$ and $w^{[i]}$ are (\ref{fiLeb}) and (\ref{wiLeb})),
solve generalized eigenvalue problem (\ref{GEVgBracket}),
the $D$ nodes are $f_G^{[m]}=\lambda_G^{[m]}$ eigenvalues, the weights $w_G^{[m]}$,
$m=0\dots D-1$, are:
\begin{subequations}
\label{gaussQ}
\begin{align}
  {\mathcal F}_{st}&=\Braket{Q_s|f|Q_t}_L=\sum\limits_{i=0}^{n-1} Q_s(f^{[i]})Q_t(f^{[i]}) f^{[i]} w^{[i]} \label{QfQg}\\
  {\mathcal G}_{st}&=\Braket{Q_s|Q_t}_L=\sum\limits_{i=0}^{n-1} Q_s(f^{[i]})Q_t(f^{[i]}) w^{[i]} \label{QQg} \\
  \Ket{{\mathcal F}\Big|\psi_G^{[m]}}_L&=\lambda_G^{[m]}\Ket{{\mathcal G}\Big|\psi_G^{[m]}}_L \label{GEVgBracket} \\
   \sum\limits_{t=0}^{D-1}{\mathcal F}_{st} \alpha^{[m]}_t&=\lambda_G^{[m]}\sum\limits_{t=0}^{D-1}{\mathcal G}_{st} \alpha^{[m]}_t \label{GEVg} \\
   \psi_G^{[m]}(f)&=\sum\limits_{t=0}^{D-1} \alpha^{[m]}_t Q_t(f) \label{psiCg} \\
   f_G^{[m]}&=\lambda_G^{[m]} \label{fg}\\
   w_G^{[m]}&=\frac{1}{\left[\psi_G^{[m]}(\lambda_G^{[m]})\right]^2}  \label{wg}\\
   \Braket{1}_L&=\Braket{1}=\sum\limits_{m=0}^{D-1} w_G^{[m]}=\sum\limits_{i=0}^{n-1} w^{[i]}
\end{align}
\end{subequations}
The eigenfunctions $\psi_G^{[m]}(f)$ are polynomials
of $D-1$ degree that are equal (within a constant)
to  Lagrange interpolating polynomials $L^{[m]}(f)$
\begin{align}
  L^{[m]}(f)&=\frac{\psi_G^{[m]}(f)}{\psi_G^{[m]}(f_G^{[m]})}=
   \begin{cases}
   1        & \text{if}\: f=f_G^{[m]} \\
   0        & \text{if}\: f=f_G^{[s]}; s\ne m
\end{cases}
   \label{LagrangeG}
\end{align}
Obtained $D$ clusters in $f$--space
 are optimal in a sense they, as the Gaussian quadrature,
optimally approximate the \textsl{distribution} of $f$
among all $D$--points discrete distributions.
The greatest advantage of this approach is that attributes selection problem of combinatorial complexity
is now reduced to generalized eigenvalue problem (\ref{GEVg}) of dimension $D$!
Obtained solution is \textbf{more generic} than
typically used disjunctive conjunction or conjunctive disjunction  forms\cite{maloldarxiv}
because it is invariant with respect
to arbitrary non--degenerated linear transform of the input attribute components $x_k$.

The eigenfunctions $\psi_G^{[m]}(f)$ (\ref{GEVg}) are obtained in $f$--space.
Because the measure  $\Braket{\cdot}_L$ (\ref{Lmeasure})
was chosen with the Lebesgue quadratures weights $w^{[i]}= \Braket{\psi^{[i]}}^2$,
the $\psi_G^{[m]}(f)$ (\ref{psiCg})
can be converted to $\mathbf{x}$ basis, $m,s=0\dots D-1$:
\begin{align}
  \psi_G^{[m]}(\mathbf{x})&=\sum\limits_{i=0}^{n-1} \psi_G^{[m]}(f^{[i]}) \Braket{\psi^{[i]}} \psi^{[i]}(\mathbf{x})
  \label{psiGclustersOpyatAZavor1e6} \\
\delta_{ms} &=  \Braket{\psi_G^{[m]}(\mathbf{x})|\psi_G^{[s]}(\mathbf{x})}\label{psiGort}\\
\lambda_G^{[m]}\delta_{ms}&= \Braket{\psi_G^{[m]}(\mathbf{x})|f|\psi_G^{[s]}(\mathbf{x})} \label{psiGfort} \\
 w_G^{[m]}&=\Braket{\psi_G^{[m]}(\mathbf{x})}^2= \Braket{\psi_G^{[m]}(f)}^2_L \label{psiGortWeight}
\end{align}
The $\psi_G^{[m]}(\mathbf{x})$
is a function on $\mathbf{x}$, it is obtained from $\psi_G^{[m]}(f)$ basis
conversion (\ref{psiGclustersOpyatAZavor1e6}).
This became possible only because the Lebesgue quadratures weights $w^{[i]}= \Braket{\psi^{[i]}}^2$
have been used to construct the $\psi_G^{[m]}(f)$ in (\ref{GEVgBracket}).
The $\psi_G^{[m]}(\mathbf{x})$ satisfies the same orthogonality conditions  (\ref{psiGort})
and (\ref{psiGfort})
for the measure $\Braket{\cdot}$ as the $\psi_G^{[m]}(f)$ for the measure $\Braket{\cdot}_L$. Lebesgue quadrature weight for $\psi_G^{[m]}(\mathbf{x})$
is the same as Gaussian quadrature weight for $\psi_G^{[m]}(f)$,
Eq. (\ref{psiGortWeight}).

The (\ref{psiGclustersOpyatAZavor1e6}) is the solution to
\href{https://en.wikipedia.org/wiki/Cluster_analysis}{clustering problem}.
This solution optimally separates $f$-- space relatively
$D$ linear combinations of $x_k$ to construct\footnote{  
The (\ref{psiGclustersOpyatAZavor1e6}) defines $D$ clusters. If 1) $D=n$,
2) all Lebesgue quadrature nodes $f^{[i]}$ are distinct and 3) no weigh $w^{[i]}$
is equal to zero, 
then $\lambda_G^{[m]}=f^{[m]}$ and $\psi_G^{[m]}(\mathbf{x})=\psi^{[m]}(\mathbf{x})$.
  }
the separation weights $\psi^2(\mathbf{x})$
of $\Braket{f\psi^2}/\Braket{\psi^2}$ form.
In the Appendix \ref{regularization}
a regularization procedure is described, and the $1+\dim S^{d}$ linear
combinations of $x_k$ were constructed to have a  non--degenerated $G_{jk}$ matrix.
No information on $f$ have been used for that regularization.
In contrast, the functions (\ref{psiGclustersOpyatAZavor1e6})
 select $D\le n$ linear
combinations of $x_k$, that optimally partition the $f$--space.
The partitioning is performed according to the
distribution of $f$, the eigenvalue problem (\ref{GEVgBracket})
of the dimension $D$ has been solved
to obtain the optimal clustering. Obtained
$\psi_G^{[m]}(\mathbf{x})$
(they are linear combination of $x_k$)
should be used as input attributes in the
approach considered in the Section \ref{RNapproach} above,
Eq. (\ref{RNfsolutionpsi}) is directly applicable, the sum now contains
$D$ terms, the number of clusters\footnote{
One can also consider a ``hierarchical''
clustering similar to ``hidden layers''  of the neural networks.
The simplest approach is to take $n$ input $x_k$ and cluster them to $D_1$,
then cluster obtained result to $D_2$, then to $D_3$, etc., $n\le D_1\le D_2 \le D_3 \dots$.
Another option is to \textsl{initially group} the $x_k$ attributes (e.g. by
temporal or spatial closeness), perform Section \ref{BasisReduction} optimal
clustering for every group to some (possibly different for different groups) $D$, then use obtained $\psi_G^{[m]}(\mathbf{x})$
for all groups as input attributes for the ``next layer''.
}.
Familiar variation expansion (\ref{evexpansionStdev})
is also applicable,
total variation
$\Braket{f^2}-\sum\limits_{m=0}^{D-1}\left(\lambda_G^{[m]}\right)^2w_G^{[m]}$
is the same when clustering to any $D$ in the range $2\le D \le n$
and is equal to least
square norm $\Braket{\left[f(\mathbf{x})-f_{LS}(\mathbf{x})\right]^2}$
calculated in original attributes basis $\mathbf{x}$ of the dimension $n$,
Eq. (\ref{norm2regrf}).

\subsection{\label{UnsupervisedLearning}Optimal Clustering For Unsupervised Learning}
Obtained optimal clustering solution assumes that there is a scalar
function $f$, which can be put to (\ref{GEV})
to obtain $\Ket{\psi^{[i]}}$, then to construct the $\Braket{\cdot}_L$
measure and to obtain optimal clusters (\ref{psiGclustersOpyatAZavor1e6}).
For unsupervised learning a function $f$ does not exist and
the best what we can do is to put
the Christoffel function as
$f(\mathbf{x})=K(\mathbf{x})$:
\begin{align}
  \sum\limits_{k=0}^{n-1} \Braket{x_j|K(\mathbf{x})|x_k} \alpha^{[i]}_k &=
  \lambda_K^{[i]} \sum\limits_{k=0}^{n-1} \Braket{ x_j|x_k} \alpha^{[i]}_k
  \label{GEVK} \\ 
 \psi_K^{[i]}(\mathbf{x})&=\sum\limits_{k=0}^{n-1} \alpha^{[i]}_k x_k
 \label{psiCK} \\
 \|\rho_K\| &=
    \sum\limits_{i=0}^{n-1} \Ket{\psi^{[i]}_K} \lambda^{[i]}_K \Bra{\psi^{[i]}_K}
\label{rhoChristoffel} \\
\Braket{1}&=\sum\limits_{i=0}^{n-1} \lambda_K^{[i]} \label{theorem4} \\
S  &= -\sum\limits_{i=0}^{n-1} \frac{\lambda_K^{[i]}}{\Braket{1}} \ln\left(\frac{\lambda_K^{[i]}}{\Braket{1}}\right)
\label{EntropyLikeK}
\end{align}
The sum of all eigenvalues (\ref{theorem4}) is equal to total measure,
see Theorem 4 of \cite{ArxivMalyshkinLebesgue}.
The (\ref{EntropyLikeK}) is an entropy
of the distribution of $\mathbf{x}^{(l)}$,
it is similar to (\ref{EntropyLikeF}), but the
weights are now obtained only from $\mathbf{x}^{(l)}$.
\begin{figure}[t]
  \includegraphics[width=16cm]{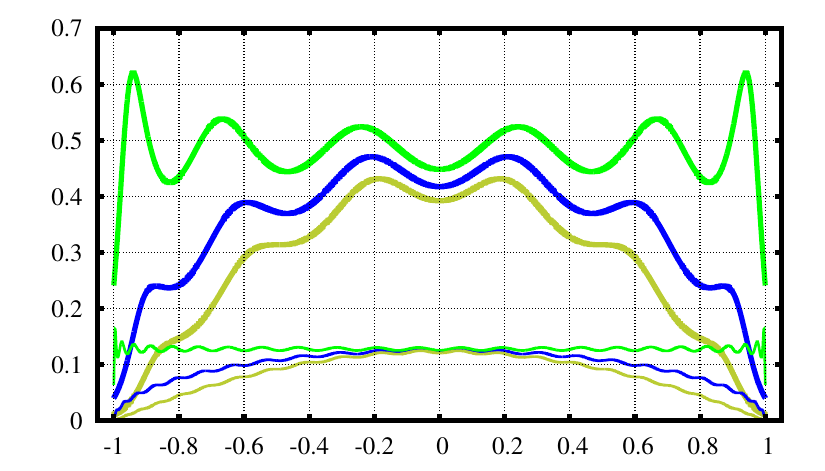}
  \caption{\label{Christoffel2Measure}
    The Christoffel function $K(x)$ for the measures $d\mu=dx$ (blue), $d\mu=dx/\sqrt{1-x^2}$ (green), and $d\mu=dx\sqrt{1-x^2}$ (olive)
    with $n=7$ and $n=25$ (thin). The $1/K(x)$ is a polynomial on $x$ of the degree $2n-2$.
    Christoffel function is determined by
integration measure and the basis used.
If one chooses
the harmonic basis: $\sfrac{1}{\sqrt{2}}$, \, $\sin(k\pi x)$, $\cos(k\pi x)$,
 $x\in [-1:1]$, $d\mu=dx$, $k=1,\dots,n-1$
then, in contradistinction to the blue line of this chart for $d\mu=dx$ in a polynomial basis,
the Christoffel function is exactly the constant $1/(n-0.5)$.
Christoffel function study for non--polynomial
bases may be an important direction of further research.
The first step in this direction
is numerical experiments:
from polynomial bases 
(where an extra degree gives one more basis function)
to harmonic basis
(where an extra degree gives two more basis functions),
following a transition to ``product'' attributes (\ref{xmulti}),
where the number of basis functions
growths with a degree as (\ref{nond}).
}
\end{figure}
In Fig. \ref{Christoffel2Measure} a demonstration of the Christoffel function in 1D case
is presented for the measures: $d\mu=dx$ and Chebyshev first and second kind $d\mu=dx/\sqrt{1-x^2}$ and $d\mu=dx\sqrt{1-x^2}$.
One can see from the figure that $K(x)$ for Chebyshev measure $d\mu=dx/\sqrt{1-x^2}$
is close to a constant,
this follows from the fact that all Gaussian quadrature weights are the
same for Chebyshev measure.
The operator $\|\rho_K\|$ allows us to construct a Chebyshev--like measure
for a multi--dimensional basis:
\begin{align}
  \|\rho_{TK}\|&=\sum\limits_{i=0}^{n-1} \Ket{\psi^{[i]}_K} \lambda_{TK}^{[i]} \Bra{\psi^{[i]}_K}
  \label{rhoChristoffelChebyshev} \\
  \lambda_{TK}^{[i]}&=\frac{\Braket{1}}{n} \label{lambdaTkChebyshev}
\end{align}
The operator $\|\rho_{TK}\|$ has the same eigenvectors as the $\|\rho_{K}\|$, but different
eigenvalues; all the eigenvalues are now the same (\ref{lambdaTkChebyshev}),
this is
a generalization from 1D Chebyshev measure.
For a large enough $n$ density matrix operator (\ref{rhoChristoffelChebyshev})
has similar to Chebyshev measure properties.
Note that the entropy (\ref{EntropyLikeK}) is maximal
for  (\ref{lambdaTkChebyshev}) distribution  (all weights are equal).
One may also consider to put entropy density
$s(\mathbf{x})= -K(\mathbf{x})\ln \big( K(\mathbf{x})/\Braket{1} \big)$
to Eq. (\ref{GEVK}) instead of $K(\mathbf{x})$ from Eq. (\ref{ChristoffelLike})
to obtain a ``spectral decomposition of the entropy''
as $S= \sum\limits_{i=0}^{n-1} \lambda_s^{[i]}$.
But it would be less convenient than the entropy (\ref{EntropyLikeK}),
where we construct a discrete distribution $\lambda_{K}^{[i]}$
and the entropy is then calculated in a usual way.
For a large enough $n$ these two approaches produce similar results.

The technique of an operator's eigenvalues adjustment
was originally developed in \cite{gsmalyshkin2017comparative}
and applied
to hydroacoustic signals processing:
first a covariation matrix is obtained and diagonalized,
second the eigenvalues (not the eigenvectors!)
are adjusted for an effective identification of weak hydroacoustic signals.
The (\ref{rhoChristoffelChebyshev}) is a transform of this type.

Before we go further let us take advantage of the basis $\Ket{\psi_K^{[i]}}$ uniqueness 
to obtain a familiar PCA variation expansion (\ref{evexpansionStdev})
but with the Christoffel function operator (\ref{rhoChristoffel}), the average
is defined as matrix Spur:
\begin{align}
  \mathrm{Spur}\left(\|\rho_K\| -\frac{\Braket{1}}{n}\|1\|\right)^2&=
  \sum\limits_{i=0}^{n-1}\left(\lambda^{[i]}_K -\frac{\Braket{1}}{n}\right)^2 \label{varexpansionK}
\end{align}
The (\ref{varexpansionK}) is invariant with respect to
an arbitrary non--degenerated linear transform of $\mathbf{x}$ components,
no scaling and normalizing is required, same as for (\ref{evexpansionStdev}).
One can select a few eigenvectors with a large $\lambda^{[i]}_K -\Braket{1}/n$
difference to capture ``most of variation''. However, our goal is not
to capture ``most of variation'' but to construct a basis of the dimension $D\le n$
that optimally separates the dataset.
Note that when the $\|\rho_{TK}\|$ operator is used in (\ref{varexpansionK})
the variation is minimal (zero).

We are interested not in variance
expansion, but in coverage expansion. If we sort eigenvalues in (\ref{theorem4})
\begin{align}
  \Braket{1}&=\sum\limits_{i=0}^{n-1} \lambda_K^{[i]}=\sum\limits_{i=0}^{n-1} \Braket{\psi_K^{[i]}|\rho_K|\psi_K^{[i]}}
  \label{coverageexpansion} 
\end{align}
is a sum of continuously decreasing terms,
by selecting a few eigenvectors we can create a projected state, that covers
a large portion of observations. This portion is minimal
for Chebyshev density matrix (\ref{rhoChristoffelChebyshev}),
where it is equal to the ratio of the number of taken/total eigenvalues.
As in the previous section
we are going 
to obtain $D\le n$ states that optimally separate the $\|\rho_K\|$
by constructing a Gaussian quadrature of the dimension $D$.
However,
in it's original form
there is an issue with the measure (\ref{Lmeasure}).

For $f(\mathbf{x})=K(\mathbf{x})$
a different separation criteria is required.
Consider the measure ``all eigenvalues are equal'', a typical one
used in random matrix theory, it is actually the Chebyshev density matrix (\ref{rhoChristoffelChebyshev}).
\begin{align}
  \Braket{g(f)}_E&=\sum\limits_{i=0}^{n-1} g(\lambda_K^{[i]}) \label{LmeasureK} \\
  \Braket{1}_E&=n \label{measureLmeasurematchK}
\end{align}
The measure (\ref{LmeasureK}) takes all eigenvectors of (\ref{GEV}) with equal weight,
the nodes are $\lambda_K^{[i]}$, the weight is 1 for every node.
If we now construct the Gaussian quadrature (\ref{gaussQ})
on the measure $\Braket{\cdot}_E$ instead of the $\Braket{\cdot}_L$,
the quadrature nodes
\begin{align}
  \lambda_G^{[m]}&=\frac{\Braket{\psi_G^{[m]}|f|\psi_G^{[m]}}_E}{\Braket{\psi_G^{[m]}|\psi_G^{[m]}}_E}
  =\frac{\sum\limits_{i=0}^{n-1} \lambda_K^{[i]} \left[\psi_G^{[m]}(\lambda_K^{[i]})\right]^2}
  {\sum\limits_{i=0}^{n-1} \left[\psi_G^{[m]}(\lambda_K^{[i]})\right]^2}
  &m=0\dots D-1
\end{align}
have a meaning of a weight per original eigenvalue\footnote{
If to use the Christoffel function average $\Braket{g(f)}_K=\sum_{i=0}^{n-1} \lambda_K^{[i]}g(\lambda_K^{[i]})$
the meaning of the nodes is unclear
${\sum\limits_{i=0}^{n-1} \left(\lambda_K^{[i]}\right)^2 \left[\psi_G^{[m]}(\lambda_K^{[i]})\right]^2}
\Big/{\sum\limits_{i=0}^{n-1} \lambda_K^{[i]}\left[\psi_G^{[m]}(\lambda_K^{[i]})\right]^2}$
}.
Then $m=0\dots D-1$ eigenfunctions $\psi_G^{[m]}(f)$ of (\ref{GEVg})
optimally cluster
the weight per eigenvalue, a ``density'' like function required
for unsupervised learning. The measure (\ref{LmeasureK})
does not allow to convert obtained optimal clustering
solution $\psi_G^{[m]}(f)$,
a pure state in $f$--space, to a pure state in $\mathbf{x}$--space $\psi_G^{[m]}(\mathbf{x})$,
however it can be converted to a density matrix state $\|\Psi_G^{[m]}\|$,
see Appendix C of \cite{ArxivMalyshkinLebesgue}.
While the $\psi_G^{[m]}(\mathbf{x})$ does not exist for a mixed state,
the $p^{[m]}(\mathbf{x})$,
an analogue of $\left[\psi_G^{[m]}(\mathbf{x})\right]^2$ that enters to
the solutions of 
Radon--Nikodym type, can always be obtained.
For the measure (\ref{LmeasureK}) the conversion is:
\begin{align}
  p^{[m]}(\mathbf{x})&=
  \sum\limits_{i=0}^{n-1}\left[\psi_K^{[i]}(\mathbf{x})\psi_G^{[m]}(\lambda_K^{[i]})\right]^2 &m=0\dots D-1
  \label{pmproj}
\end{align}
for a general case see Appendix C of \cite{ArxivMalyshkinLebesgue}.

In this section a completely new look to
unsupervised learning
PCA expansion is presented.
Whereas a ``regular'' PCA expansion
is attributes variation expansion,
which is scale--dependent and often does not have
a clear domain problem meaning\footnote{
  There is a situtation\cite{gsmalyshkin2017comparative}
  where the variation
  has a meaning of total energy $E=\sum_{j,k=0}^{n-1} x_j E_{jk} x_k$,
  the energy matrix $E_{jk}$ is determined by antenna design.},
the Christoffel function density matrix
expansion (\ref{coverageexpansion}) is coverage expansion:
every eigenvector covers some observations,
total sum of the eigenvalues is equal to total measure $\Braket{1}$,
the answer is invariant relatively any
non--degenerated linear transform of input vector $\mathbf{x}$ components.
In the simplistic form one can select a few eigenvectors with
a large $\lambda_K^{[i]})$ (e.g. use \texttt{\seqsplit{--flag\_replace\_f\_by\_christoffel\_function=true}} with the Appendix \ref{RN} software).
In a more advanced form $D\le n$ optimal clusters can be obtained
by constructing a Gaussian quadrature with the measure (\ref{LmeasureK})
and then converting
the result back to $\mathbf{x}$--space with (\ref{pmproj}) projections.

\section{\label{application}Selection of the Answer: $f_{RN}$ vs. $f_{RNW}$}
For a given input attributes vector we now have two answers:
interpolation $f_{RN}$ (\ref{RNfsolutionpsi})
and  classification $f_{RNW}$ (\ref{RNWfsolutionpsi}).
Both are the answers of Radon--Nikodym
$\Braket{f\psi^2}/\Braket{\psi^2}$ form,
that can be reduced to weighted eigenvalues
with $\mathrm{Proj}^{[i]}$ and $w^{[i]}\mathrm{Proj}^{[i]}$
weights respectively.
A question arise which one to apply.

For a deterministic function
$f(\mathbf{x})$, the $\mathrm{Proj}^{[i]}$ weights from (\ref{distY})
construct the state in $\Ket{\psi^{[i]}}$ basis that is the most close
to a given observation $\mathbf{x}$.
The $f_{RN}$ is a regular Radon--Nikodym derivative
of the measures $fd\mu$ and $d\mu$,
see Section II.C of \cite{ArxivMalyshkinLebesgue}.
This is a solution of interpolatory
type, see Appendix \ref{reproduce} below for a demonstration.

For a probabilistic $f$
the  $w^{[i]}\mathrm{Proj}^{[i]}$ weights, that include prior
probability of $f$ outcomes, is a preferable form
of outcome probabilities estimation, see Appendix \ref{NominalExample} below for a demonstration.
The $w^{[i]}\mathrm{Proj}^{[i]}$ posterior weights typically produce a
good classification even without
optimal clustering algorithm of Section \ref{BasisReduction}.
For a given scalar $f$
the solution to
\href{https://en.wikipedia.org/wiki/Supervised_learning}{supervised learning}
problem
is obtained in the form of (outcome,weight) 
posterior distribution (\ref{Pposterior}).

For \href{https://en.wikipedia.org/wiki/Unsupervised_learning}{unsupervised learning}  the function $f$
does not exist,
thus the eigenvalue problem (\ref{GEVBracket}) cannot be formulated.
However, we still want to obtain a unique basis
that is constructed from the data,
for example to avoid
PCA dependence on attributes scale.
For unsupervised learning
the Christoffel function should be used as $f(\mathbf{x})=K(\mathbf{x})$,
then
PCA expansion of coverage can be obtained,
this is 
an approach of Section \ref{UnsupervisedLearning}
to unsupervised learning.

\section{\label{firstOrderLogic}A First Order Logic Answer
  To The Classification Problem. Product Attributes.}
Obtained solutions to interpolation (\ref{RNfsolutionpsi})
and classification (\ref{Pposterior})
problems are more general than
a \href{https://en.wikipedia.org/wiki/Propositional_calculus#Terminology}{propositional logic} type of answer.
A regular basis function expansion (\ref{regrf})
is a local function of arguments, thus it 
can be considered as a ``propositional logic'' type of answer. 
Consider formulas including a quantor operator,
e.g. for a binary $x_k$ and $f$ in (\ref{mlproblem}) expressions like these:
\begin{align*}
\mathrm{if}\,\exists  x_k=1 & \text{ then $f=1$} \\
\mathrm{if}\, \forall  x_k=0 & \text{ then $f=1$}
\end{align*}
Similar expressions can be written for continuous $x_k$ and $f$,
the difference from the propositional logic is that
these expressions include a quantor--like operator that is a function
of several
$x_k$ attributes. The $\psi^2(\mathbf{x})$ expansion
includes the products of $x_jx_k$, thus the Radon--Nikodym representation
can be viewed
as a more general  form than a propositional logic.
The most straightforward approach to obtain
a ``true'' \href{https://en.wikipedia.org/wiki/First-order_logic}{first order logic}
answer from a propositional logic model
is to add all possible $Q_{k_0}(x_0)Q_{k_1}(x_1)\dots Q_{k_{n-1}}(x_{n-1})$ products
to the list of input attributes.
For a large enough ${\mathcal D}$ (\ref{countd})
we obtain a model with properties that are
very similar to a first order logic model.
The
attributes $x_{\mathbf{k}}$ are now polynomials
of $n$ variables with
\href{https://en.wikipedia.org/wiki/Multi-index_notation#Definition_and_basic_properties}{multi--index}
$\mathbf{k}$
of a degree ${\mathcal D}$;
they are constructed from  initial attributes $x_k$ with regular index $k$.
Multi--index degree (\ref{countd})
is \textbf{invariant}
relatively any linear transform of the attributes:
$x^{\prime}_{j}=\sum_{k=0}^{n-1}T_{jk} x_k$.
Because in the Radon--Nikodym 
approach
all the answers are invariant relatively any
non--degenerated linear transform of the basis,
we can construct similar to the first order logic
knowledge representation with known invariant group!
The situation is different with logical formulas of 
disjunctive conjunction or conjunctive disjunction,
where a basis transform change formula index\cite{maloldarxiv},
and the invariant group is either completely unknown or poorly understood;
a typical solution in this situation is to introduce a ``formula complexity''
concept to limit the formulas to be considered,
a mutli--index constraint (\ref{countd}) can be viewed as
a complexity of the formulas allowed.
The terms
\begin{align}
  x_{\mathbf{k}}&= x_0^{k_0}x_1^{k_1}\dots x_{n-1}^{k_{n-1}} \label{xmulti} \\
  \mathbf{k}&=(k_0,k_1,\dots,k_{n-1}) \label{kmultiindex} \\
  {\mathcal D}&=\sum\limits_{j=0}^{n-1}k_j \label{countd} \\
  {\mathcal N}(n,{\mathcal D}) &= C_{n+{\mathcal D}-1}^{\mathcal D} \label{nond}
\end{align}
are now
identified by a multi--index $\mathbf{k}$ and
added to (\ref{mlproblem}) as attributes\footnote{
  Note, that since the constant
  does always present in the original $x_k$ attributes (\ref{mlproblem})
  linear combinations,
  the $x_jx_k$ (and high order) products always include
  the $x_k$ (lower order products), what may produce
  a degenerated basis. The degeneracy can
  be removed either manually or by applying any regularization algorithm,
  such as the one from Appendix \ref{regularization}. 
  Unlike polynomials in a single variable,
  multidimensional polynomials cannot, in general, be factored\cite{hayes1982reducible,nieto1993fortran}.
}.
We will call the set of all possible (\ref{xmulti}) terms
used as ML attributes in (\ref{mlproblem}) ---
 the ``product'' attributes.
An individual (\ref{xmulti})
is called ``term'', see \cite{becker1993grobner,nalimov1968statistical,nalimov1971theoryOfExperiment}.
The number ${\mathcal N}(n,{\mathcal D})$ of ``product'' attributes
is the number of possible polynomial distinct terms
with multi--index not higher than ${\mathcal D}$,
it is equal to (\ref{nond}).
A few values:
${\mathcal N}(n,1)=n$,
${\mathcal N}(n,2)=(n+1)n/2$,
${\mathcal N}(7,7)=1716$, ${\mathcal N}(8,7)=3432$, etc.
In a typical ML setup such a transform to ``product'' attributes
is not a good idea because of:
\begin{itemize}
\item  A linear transform of input attributes
  produces a different solution, no gauge invariance.
\item Attributes offset and normalizing difficulty.
\item Data overfitting (\href{https://en.wikipedia.org/wiki/Curse_of_dimensionality}{curse of dimensionality}), as we now have a much bigger number of input attributes  ${\mathcal N}(n,{\mathcal D})$.
  A second complexity criteria (the first one is maximal multi--index (\ref{countd}))
  of constructed attributes is typically introduced
  to limit the number of input attributes.
  For example,
  a \href{https://en.wikipedia.org/wiki/Types_of_artificial_neural_networks}{neural network topology} can be considered as
  a variant of a complexity criteria.
\end{itemize}
The approach developed in this paper has these difficulties solved.
The invariant group is a non--degenerated linear transform $T_{jk}$
of input attributes components,
the $x_jx_k$ and $\sum_{j^{\prime},k^{\prime}=0}^{n-1}T_{jj^{\prime}}x_{j^{\prime}}T_{kk^{\prime}}x_{k^{\prime}}$
attributes
produce identical solutions;
for the same reason the terms (\ref{xmulti})
$Q_{k_0}(x_0)Q_{k_1}(x_1)\dots Q_{k_{n-1}}(x_{n-1})$
are $Q_k$ invariant, e.g. $Q_k(x)=x^k$ and $Q_k(x)=T_k(x)$
produce identical solutions.
The attributes offset and normalizing are not important since (\ref{GEV})
is invariant relatively any non--degenerated linear transform of $\mathbf{x}$ components.
The problem of data overfitting is not an issue since Section \ref{BasisReduction}
optimal clustering solution (\ref{psiGclustersOpyatAZavor1e6})
allows to reduce ${\mathcal N}(n,{\mathcal D})$ input attributes to a given number $D$
of their linear combinations that optimally separate the $f$.
The only cost to pay is that the Lebesgue quadrature
now requires a generalized eigenproblem of ${\mathcal N}(n,{\mathcal D})$ dimension
to be solved, but this is purely a computational complexity issue.
Critically important, that
we are now limited not by the data overfitting,
but by the computational complexity. Regardless
input attributes number
the optimal clustering solution (\ref{psiGclustersOpyatAZavor1e6})
selects given number $D\ll{\mathcal N}(n,{\mathcal D})$
of input attributes linear combinations
that optimally separate $f$ in terms of $\Braket{f\psi^2}/\Braket{\psi^2}$.

In the Appendix \ref{reproduce} a simple example
of usage of polynomial function of a single attribute $x$
as input attributes
was demonstrated
(\ref{mlproblem1D}).
Similarly, a polynomial of several variables (\ref{xmulti})
identified by the
multi--index (\ref{kmultiindex})
can be used to construct input attributes\footnote{\small
See numerical implementation of multi--index recursive
processing in \texttt{\seqsplit{com/polytechnik/utils/AttributesProductsMultiIndexed.java}}.
Due to invariant group of the Radon--Nikodym approach
 the ``product'' attributes (\ref{xmulti}) can be calculated in any basis.
For example these two solutions are identical:
\begin{itemize}
\item Take original basis, perform basis regularization of Appendix \ref{regularization},
 obtain ``product'' attributes (\ref{xmulti}) from $X_k$,
 then solve (\ref{GEV}) of ${\mathcal N}(n,{\mathcal D})$ dimension.
 Obtain the Lebesgue quadrature (\ref{lebesgueQ}).
\item In the previous step, after $X_k$ calculation, solve (\ref{GEV})
  of dimension $n$ to find $\psi^{[i]}(\mathbf{x})$ (\ref{psiC}),
  obtain ``product'' attributes (\ref{xmulti})
  from these $\psi^{[i]}(\mathbf{x})$,
  then solve (\ref{GEV}) of ${\mathcal N}(n,{\mathcal D})$ dimension.
  Obtain (\ref{lebesgueQ}).
\end{itemize}
See \texttt{\seqsplit{com/polytechnik/utils/TestDataReadObservationVectorXF.java:testAttributesProducts()}} for unit test example.
This result is also invariant to
input attributes ordering method.

For highly degenerated input attributes a direct application of  \texttt{\seqsplit{com/polytechnik/utils/AttributesProductsMultiIndexed.java}} algorithm to create ${\mathcal N}(n,{\mathcal D})$ ``product attributes'' and then regularize them all at once may not be the best approach
from computational stability point of view.
In this case it may be a better option to perform basis regularization
  incrementally, simultaneously with product attributes construction:
  obtain original basis regularized attributes ${\mathcal B}^{(1)}$,
  multiply them by itself (square), regularize the products to obtain the basis ${\mathcal B}^{(2)}$. Repeat the procedure: on each step multiply
  the basis ${\mathcal B}^{(d-1)}$ by ${\mathcal B}^{(1)}$ and do a regularization
  of products
  to obtain ${\mathcal B}^{(d)}$
  until the sought basis ${\mathcal B}^{({\mathcal D})}$
  is obtained.
}.
An increase of attributes number from $n$ to ${\mathcal N}(n,{\mathcal D})$
using ``product'' attributes (\ref{xmulti})
combined with subsequent attributes number decrease to $D$
by the clustering solution (\ref{psiGclustersOpyatAZavor1e6})
is a path to ML answers of the first order logic type:
$n$ original attributes (\ref{mlproblem})
$\rightarrow$ ${\mathcal N}(n,{\mathcal D})$ ``product'' attributes (\ref{xmulti})
$\rightarrow$ $D$ cluster attributes (\ref{psiGclustersOpyatAZavor1e6}).

\subsection{\label{LenaInterpolationMI}Lenna Image Interpolation Example. Multi--index Constraints Comparison.}
In \cite{2015arXiv151101887G} a two--dimensional image interpolation problem
was considered with multi--index $\mathbf{j}$ constraint
\begin{align}
  (x,y)^{(l)}&\to f^{(l)} & &\text{weight $\omega^{(l)}=1$}  \label{lenmlproblem} \\
  \mathbf{j}&=(j_x,j_y) \label{lenMI} \\
  0 &\le j_x \le n_x-1 \label{lenMIconstraintX} \\
  0 &\le j_y \le n_y-1 \label{lenMIconstraintY} \\
  \text{basis}&: x^{j_x}y^{j_y} & &\dim(\text{basis})=n_xn_y \label{lenBais}
\end{align}
of each multi--index component being in the $[0\dots n_{\{x,y\}}-1]$ range;
total number of basis functions is then $n_xn_y$ (\ref{lenBais}).
This is different from the constraint (\ref{countd}), where the sum of all
multi--index components is equal to ${\mathcal D}$;
total number of basis functions is then (\ref{basisImageC}).
Different basis functions produce
different interpolation,
let us compare the interpolation in these two bases.
Transform $d_x\times d_y$ image pixel coordinates $(x,y)$ ($x=0\dots d_x-1$; $y=0\dots d_y-1$)
and gray intensity $f$ to the data of (\ref{mlproblem}) form:
\begin{align}
  (x,y,1)^{(l)}&\to f^{(l)} & &\text{weight $\omega^{(l)}=1$}  \label{mlproblemImage} \\
  \mathbf{j}&=(j_x,j_y,j_c) \label{miImageC}\\
  {\mathcal D}&=j_x+j_y+j_c \label{DImage} \\
  \text{basis}&: x^{j_x}y^{j_y}=x^{j_x}y^{j_y}1^{j_c}  &  &\dim(\text{basis})={\mathcal N}(n,{\mathcal D})
  \label{basisImageC}
\end{align}
Input attributes vector $\mathbf{x}$ is of the dimension $n=3$: two pixel coordinates and const,
this way the (\ref{xmulti}) ``product'' attributes with the constraint (\ref{DImage})
include
all $x^{j_x}y^{j_y}$ terms with lower than ${\mathcal D}$ degree $j_x+j_y\le {\mathcal D}$. Observation index $l$ runs from $1$ to
the total number of pixels $M=d_x\times d_y$.

\begin{figure}
  \includegraphics[width=5.6cm]{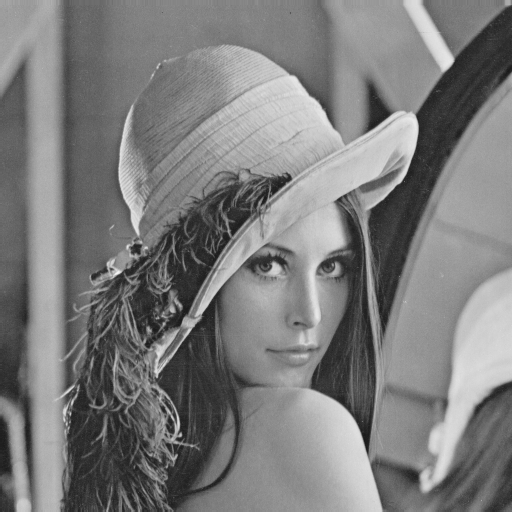}\\
  \includegraphics[width=5.6cm]{{lena512.bmp.LS.50.50.bs2500}.png}
  \includegraphics[width=5.6cm]{{lena512.bmp.LS.D.69.bs2485}.png} \\
  \includegraphics[width=5.6cm]{{lena512.bmp.RN.50.50.bs2500}.png}
  \includegraphics[width=5.6cm]{{lena512.bmp.RN.D.69.bs2485}.png}  
  \caption{\label{lennaimg}Top: original image.
    Middle: least squares in (\ref{lenBais}) basis (left)
    and  (\ref{basisImageC}) basis (right).
    Bottom: Radon--Nikodym  in (\ref{lenBais}) basis (left)
    and (\ref{basisImageC}) basis (right).
    The bases (\ref{lenBais}) and (\ref{basisImageC})
    are of 2500 elements ($n_x=n_y=50$)
    and 2485 elements ($n=3$, ${\mathcal D}=69$) respectively.
  }
\end{figure}

Let us compare \cite{2015arXiv151101887G}
$n_x=n_y=50; \dim(\text{basis})=n_xn_y=2500$  of
basis (\ref{lenBais})
with $n=3; {\mathcal D}=69; \dim(\text{basis})={\mathcal N}(n,{\mathcal D})=2485$
of basis (\ref{basisImageC}).
The value of ${\mathcal D}=69$ is selected
to have approximately the same
total number of basis functions. The bases are different:
$x^{67}y^{2}$, $x^{66}y^{2}$, etc. are among ``product'' attributes (\ref{basisImageC}),
but they are not among the (\ref{lenBais}) where the maximal degree
for $x$ and $y$ is $49$;
similarly the $x^{49}y^{49}$ is in (\ref{lenBais}), but it is not in (\ref{basisImageC}).
As in \cite{2015arXiv151101887G}
we choose 512x512
\href{https://en.wikipedia.org/wiki/Lenna}{Lenna}  grayscale image
as a testbed.
If you have \href{https://www.scala-lang.org/}{scala} installed run
\begin{verbatim}
 scala com.polytechnik.algorithms.ExampleImageInterpolation \
   file:dataexamples/lena512.bmp 50 50 chebyshev
\end{verbatim}
to reproduce \cite{2015arXiv151101887G} results using
(\ref{regrfsolution}) and (\ref{RNfsolution}) for least squares and Radon--Nikodym. Then run (note: this code is unoptimized and slow):
\begin{verbatim}
 java com/polytechnik/algorithms/ExampleImageInterpolation2 \
   file:dataexamples/lena512.bmp 50 50 69
\end{verbatim}
To obtain 4 files.
The files \texttt{\seqsplit{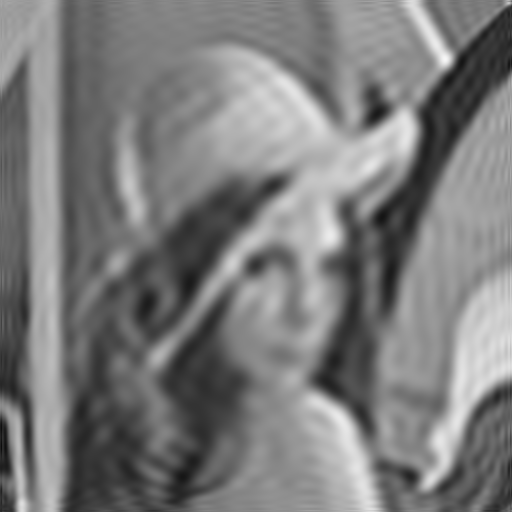}} and
\texttt{\seqsplit{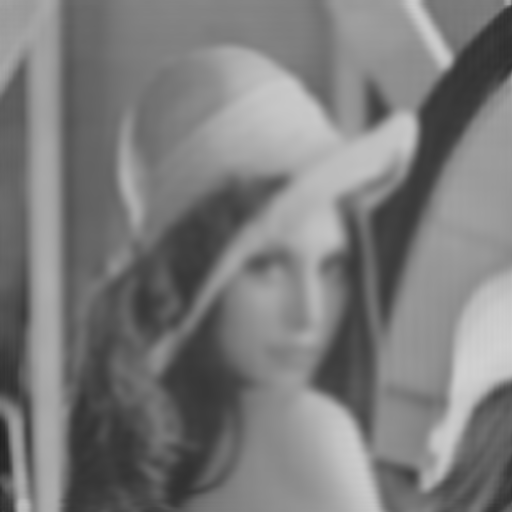}}
are obtained as (\ref{regrfsolutionpsi})
and (\ref{RNfsolutionpsi}) using (\ref{lenBais}) basis with $n_x=n_y=50$,
the result is
identical to \cite{2015arXiv151101887G}.
The files
\texttt{\seqsplit{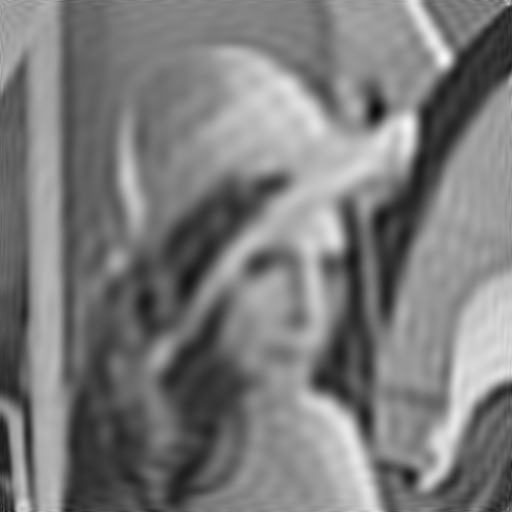}} and
\texttt{\seqsplit{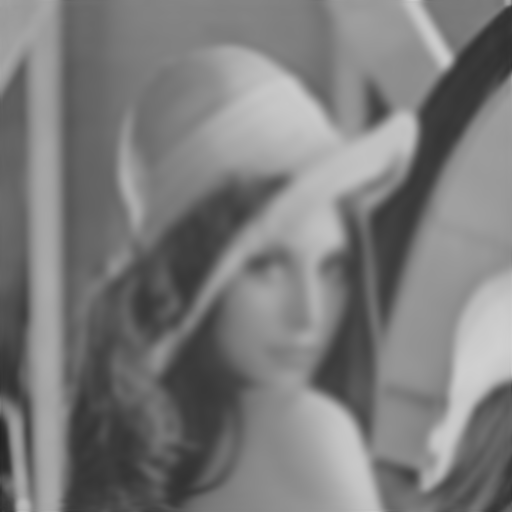}}
are obtained from (\ref{regrfsolutionpsi})
and (\ref{RNfsolutionpsi}) using (\ref{basisImageC}) basis with ${\mathcal D}=69$.
The images are presented in Fig. \ref{lennaimg}.
It was shown in \cite{2015arXiv151101887G} that
the Radon--Nikodym interpolation produces a
\href{https://en.wikipedia.org/wiki/Sfumato}{sfumato} type of picture
because it averages with always positive weight $\psi^2(\mathbf{x})$;
the (\ref{RNfsolutionpsi}) preserves the bounds of $f$:
if original gray intensity is $[0:1]$ bounded
then interpolated gray intensity is $[0:1]$ bounded as well;
this is an important difference from positive polynomials
interpolation\cite{bernard2009moments} where only
a low bound (zero) is preserved.
In contradistinction to Radon--Nikodym
the least squares interpolation
strongly oscillates near image edges
and may not preserve the bounds of gray intensity $f$.
In this section we compare not least squares vs. Radon--Nikodym
as we did in \cite{2015arXiv151101887G}
but the bases: (\ref{lenBais}) vs. (\ref{basisImageC})
as they have different multi--index constraints.
We observe that:
\begin{itemize}
\item The bases produce similar results.
  Basis differences in LS are more pronounced, than in RN;
  always positive weight makes the RN less sensitive to basis choice.
\item In RN a small difference is observed near image edges.
  With (\ref{lenBais}) RN still has small oscillations near edges,
  and with (\ref{basisImageC}) RN has oscillations completely suppressed.
\item The multi--index constraint (\ref{lenBais})
  is not invariant relatively
  a linear transform of input attributes,
  for example $x^{n_x-1}y^{n_y-1}$ relatively $x=x^{\prime}-y^{\prime}$, $y=x^{\prime}+y^{\prime}$,
  but the (\ref{basisImageC})
  is invariant.
\end{itemize}
This make us to conclude that the specific multi--index
constraint is not very important,
the results are similar.
Whereas in an interpolation problem
an explosion of
basis functions number
increases interpolation precision,
in a classification problem 
an explosion of
basis functions number leads to data overfitting.
The optimal clustering solution (\ref{psiGclustersOpyatAZavor1e6})
reduces the number of basis functions to a given $D$
thus
it solves
the problem of
data overfitting.
This reduction makes multi--index constraint
used for initial basis construction
even less important
for a classification problem than for an interpolation problem.

\subsection{\label{Lasserte}On The Christoffel Function Conditional Optimization}
All the solutions obtained in this paper have a distribution of $f$
as the answer:
the distribution
with posterior weights (\ref{Pposterior}),
optimal clustering (\ref{gaussQ}), etc.
Recently, a promising approach
to interpolation problem
has been developed \cite{marx2019tractable}.
In this subsection we consider a modification
of it to obtain, for a given $\mathbf{x}$,
not a single outcome of $f$, but a distribution.
Obtained weights can be considered as an alternative to the posterior weights
(\ref{Pposterior}).
A sketch of  \cite{marx2019tractable} theory:
\begin{itemize}
\item Introduce a vector $\mathbf{z}=(\mathbf{x},f)$ of the
  dimension $n+1$.
\item Construct ``product'' attributes (\ref{xmulti}) out of $\mathbf{z}$
  components with the degree equals to ${\mathcal D}$;
  because a constant always presents in $x_k$
  it is sufficient to consider the  degree \textsl{equals} to ${\mathcal D}$,
  lower order terms are obtained automatically as in (\ref{basisImageC}).
  There are ${\mathcal N}(n+1,{\mathcal D})$ ``product'' attributes
  obtained from $n+1$ components of $\mathbf{z}$.
\item Construct  Christoffel function (\ref{ChristoffelLike})
  from obtained ``product'' attributes $K(\mathbf{z})=K(\mathbf{x},f)$.
  Now the $1/K(\mathbf{z})$, for a given $\mathbf{x}$,
  is a positive polynomial on $f$ of the degree $2{\mathcal D}$.
\item
  For a given $\mathbf{x}$,
  the interpolation  \cite{marx2019tractable} of $f$ is the value
  providing the minimum of the polynomial $1/K(\mathbf{x},f)$;
  the value of $\mathbf{x}$ is fixed:
  \begin{align}
    K(\mathbf{x},f)\Big\vert_{\mathbf{x}}&
      \xrightarrow[{f}]{\quad }\max
    \label{ChristoffelFunMaxLassete}
\end{align}
\end{itemize}
As an extension of this approach
consider Christoffel function average, Appendix B of \cite{ArxivMalyshkinLebesgue}, but use the $K(\mathbf{z})=K(\mathbf{x},f)$
to calculate the moments of $f$:
\begin{align}
  \Braket{f^{m}}_{K(\mathbf{x},\cdot)}&=\Braket{f^{m} K(\mathbf{z})\Big\vert_{\mathbf{x}}}=
  \sum\limits_{l=1}^{M} \frac{\left(f^{(l)}\right)^m}{1/K(\mathbf{x},f^{(l)})}\omega^{(l)}
    \label{ChristoffelFunMeasureSum}
\end{align}
When one uses $\mathbf{x}=\mathbf{x}^{(l)}$ as Christoffel
function argument in the right hand side of (\ref{ChristoffelFunMeasureSum}),
the average is the Christoffel function average of
Ref.  \cite{ArxivMalyshkinLebesgue}
with the properties similar to regular average (\ref{matrixfxx});
the Gaussian quadrature built from the moments
obtained with the Christoffel function average
is similar to the one
built from the regular moments $\Braket{f^{m}}$,
and to the one built from (\ref{Lmeasure}) moments with $g=f^m$.
However, if to consider a \textsl{fixed} value of $\mathbf{x}$,
then the solution becomes similar to the approach of Ref. \cite{marx2019tractable},
the $K(\mathbf{x},f)$ is now used as a proxy to
joint distribution $\rho(\mathbf{x},f)$.
Because $1/K(\mathbf{x},f)$ at fixed $\mathbf{x}$
is a positive polynomial on $f$ of the degree $2{\mathcal D}$,
the moments $\Braket{f^{m}}_{K(\mathbf{x},\cdot)}$ do exist for at least $m=0\dots 2{\mathcal D}$.
A ${\mathcal D}$--point Gaussian quadrature can be
built from them,
exactly as (\ref{gaussQ}), but with the measure $\Braket{\cdot}_{K(\mathbf{x},\cdot)}$
instead of $\Braket{\cdot}_L$.
The result is ${\mathcal D}$ nodes (\ref{fg}) and weights (\ref{wg}).
The major difference from \cite{marx2019tractable} is that
instead of single $f$ we now obtained $i=0\dots {\mathcal D}-1$ (outcome,weight) pairs $(f_{K(\mathbf{x},\cdot)}^{[i]},w_{K(\mathbf{x},\cdot)}^{[i]})$
of the distribution of $f$ conditional to a given $\mathbf{x}$.
The most close to  \cite{marx2019tractable} interpolation answer is
to find the $f_{K(\mathbf{x},\cdot)}^{[i]}$, corresponding to the maximal $w_{K(\mathbf{x},\cdot)}^{[i]}$.
However, in ML the distribution of outcomes, not a single ``answer'', is of most interest.
From the Gaussian quadrature built on the $\Braket{\cdot}_{K(\mathbf{x},\cdot)}$
measure conditional distribution
characteristics can be obtained:
\begin{itemize}
\item
  The $\Braket{1}_{K(\mathbf{x},\cdot)}$
  is an analogue of $\mathrm{Coverage}(\mathbf{x})$ from (\ref{coverage}):
  how many observations are ``close enough'' to a given $\mathbf{x}$.
\item
  The Gaussian quadrature nodes and weights
  $(f_{K(\mathbf{x},\cdot)}^{[i]},w_{K(\mathbf{x},\cdot)}^{[i]})$  are
  an analogue of the posterior distribution (\ref{Pposterior}).
  However, in (\ref{ChristoffelFunMeasureSum}) approach
  both: the outcomes $f_{K(\mathbf{x},\cdot)}^{[i]}$ and the weights $w_{K(\mathbf{x},\cdot)}^{[i]}$
  depend on $\mathbf{x}$. In (\ref{Pposterior}) approach
  the outcomes are always the same $f^{[i]}$ and
  only posterior weights depend on $\mathbf{x}$ as $w^{[i]}\mathrm{Proj}^{[i]}(\mathbf{x})$.
  This distinction
  is similar to \cite{2015arXiv151107085G} with $\mathbf{x}$--dependent outcomes
  vs. \cite{2015arXiv151109058G} with $\mathbf{x}$--independent outcomes.
\item The approach (\ref{ChristoffelFunMeasureSum}) cannot provide
  an optimal clustering solution of (\ref{psiGclustersOpyatAZavor1e6}) type.
  Ideologically,
  $\mathbf{x}$--dependent outcomes make optimal clustering difficult.
  Technically,
 the $m=0\dots2{\mathcal D}$ moments
  $\Braket{f^{m}}_{K(\mathbf{x},\cdot)}$ cannot be reduced to a
  density matrix average of Appendix C of \cite{ArxivMalyshkinLebesgue}  
  or to a simple pure state average (\ref{wiLeb}).
\end{itemize}
  
\section{\label{vectorvalued}A Supervised Classification Problem With Vector--Valued Class Label}
In the ML problem (\ref{mlproblem}) the class label
$f$ is considered to be a scalar.
A problem with vector--valued class label $\mathbf{f}$
\begin{align}
  (x_0,x_1,\dots,x_k,\dots,x_{n-1})^{(l)}&\to
  (f_0,f_1,\dots,f_j,\dots,f_{m-1})^{(l)}
  & \text{weight $\omega^{(l)}$}  \label{mlproblemVector}
\end{align}
where an attributes vector $\mathbf{x}$ of the dimension $n$
is mapped to a class label vector $\mathbf{f}$ of the dimension $m$
is a much more interesting case.
For a vector  class label $\mathbf{f}$, the
most straightforward approach is to build
an individual model for every $f_j$ component.
However, constructed models
are often completely different
and obtained model set cannot be viewed as a probability space.
In addition, the invariant group of $\mathbf{f}$
(what transform of $f_j$ components does not change the prediction)
may become unknown and basis--dependent.
The situation is similar to the one of
our previous works\cite{2015arXiv151107085G,2015arXiv151109058G},
where the 
distribution regression
problem can be directly approached by the Radon--Nikodym technique,
however the distribution to distribution regression problem
is a much more difficult case.

Whereas the Christoffel function
maximization  approach (\ref{ChristoffelFunMaxLassete})
of Ref. \cite{marx2019tractable}
is interesting for a scalar $f$,
it becomes extremely promising for a vector class label $\mathbf{f}$.
Consider a vector $\mathbf{z}$ of the dimension $n+m$:
\begin{align}
  \mathbf{z}&=
  (x_0,x_1,\dots,x_k,\dots,x_{n-1},
  f_0,f_1,\dots,f_j,\dots,f_{m-1})^{(l)}
  & \text{weight $\omega^{(l)}$}  \label{ZVector}
\end{align}
The vector $\mathbf{z}$ mixes input attributes $\mathbf{x}$
with class label vector $\mathbf{f}$.
The ${\mathcal N}(n+m,{\mathcal D})$ ``product'' attributes $Z_i$
can be obtained out of $n+m$ $\mathbf{z}$ components
as in (\ref{xmulti}).
The ``product'' attributes $Z_i$ with the constraint (\ref{countd})
are the ones with the simplest invariant group:
the answer is invariant relatively any
non--degenerated linear transform of $\mathbf{z}$ components:
$z^{\prime}_{s}=\sum_{s^{\prime}=0}^{n+m-1}T_{ss^{\prime}} z_{s^{\prime}}$;
$s,s^{\prime}=0\dots n+m-1$\footnote{
In practical applications,
it is often convenient
to consider different degree ${\mathcal D}$
for $\mathbf{x}$ and $\mathbf{f}$,
e.g. to consider ${\mathcal D}>1$
only for $\mathbf{x}$
to obtain
${\mathcal N}(n,{\mathcal D})$
``product'' attributes
and, for the class label,
consider ${\mathcal D}=1$.
There are will be $m={\mathcal N}(m,1)$ attributes $f_j$,
total ${\mathcal N}(n,{\mathcal D})+m$ attributes $Z_i$.
Below we consider only the case of the constraint (\ref{countd}),
providing  ${\mathcal N}(n+m,{\mathcal D})$ attributes $Z_i$.
The transition to ``product'' attributes
extends the basis space,
but the $\Ket{\psi}$ still form a linear space \cite{bourass2005random}.
}.
The invariant group can be viewed
as a \href{https://en.wikipedia.org/wiki/Introduction_to_gauge_theory}{gauge transformations}
and is a critical insight into the ML model built.

From (\ref{ZVector}) $\mathbf{z}$ data 
construct ${\mathcal N}(n+m,{\mathcal D})$ ``product'' attributes $Z_i$
according to (\ref{countd}) (if necessary perform regularization
of the Appendix \ref{regularization}),
then, finally, construct the Christoffel function $K(\mathbf{z})$
according to (\ref{ChristoffelLike}).
Classification problem is to 
find $\mathbf{f}$--prediction
for a given $\mathbf{x}$.
When one puts $x_k, k=0\dots n-1$ part of vector $\mathbf{z}$
equal to a given $\mathbf{x}$ the
$K(\mathbf{x},\mathbf{f})$, for a fixed $\mathbf{x}$,
can be viewed as a a proxy to
joint distribution
 $\rho(\mathbf{x},\mathbf{f})$.
Find it's maximum over the vector $\mathbf{f}$:
\begin{align}
    K(\mathbf{x},\mathbf{f})\Big\vert_{\mathbf{x}}&
      \xrightarrow[{\mathbf{f}}]{\quad }\max
    \label{ChristoffelFunMaxVector}
\end{align}
to obtain  Ref.  \cite{marx2019tractable}
solution.
The
solution (\ref{ChristoffelFunMaxVector})
is exactly (\ref{ChristoffelFunMaxLassete}),
but with a
\textbf{vector} class label $\mathbf{f}$!

For a fixed $\mathbf{x}$ and a degree ${\mathcal D}$
the $1/K(\mathbf{x},\mathbf{f})\Big\vert_{\mathbf{x}}$
is a polynomial on $f_j$ of the degree $2{\mathcal D}$,
there are total ${\mathcal N}(m,{\mathcal D})$ distinct terms.
In applications it may be convenient
to minimize the polynomial $1/K(\mathbf{x},\mathbf{f})\Big\vert_{\mathbf{x}}$
instead of maximizing  the Christoffel function
(\ref{ChristoffelFunMaxVector}),
but these are implementation details.

Critically important, that, for a given $\mathbf{x}$,
we now obtained
a probability distribution of $\mathbf{f}$ as
$K(\mathbf{x},\mathbf{f})\Big\vert_{\mathbf{x}}$.
When a specific value of $\mathbf{f}$ is required,
it
can be estimated  from the distribution as:
\begin{itemize}
\item Christoffel function
maximum (\ref{ChristoffelFunMaxVector}).
\item
  The distribution of Christoffel function eigenvalues (\ref{GEVK})
\item
  The simplest one is to average $\mathbf{f}$
with $K(\mathbf{x},\mathbf{f})\Big\vert_{\mathbf{x}}$,
same as  
 (\ref{ChristoffelFunMeasureSum})
but with the vector $\mathbf{f}$ instead of $f^m$:
$\Braket{\mathbf{f} K(\mathbf{z})\Big|_{\mathbf{x}}}$
and similar generalizations.
\end{itemize}
The most remarkable feature is that the
$K(\mathbf{x},\mathbf{f})\Big\vert_{\mathbf{x}}$
approach is trivially applicable to a vector class label $\mathbf{f}$,
and the constructed model has a known
``gauge group''.

\subsection{\label{AttribsVectorF}A Vector--Valued Class Label: Selecting Solution Type}
While the idea \cite{marx2019tractable}
to combine input attributes $\mathbf{x}$
with class label vector $\mathbf{f}$
into a single vector $\mathbf{z}$ (\ref{ZVector})
with subsequent construction of
``product'' attributes $\mathbf{Z}$ (\ref{xmulti})
and finally to obtain Gram matrix $\Braket{Z_iZ_j}$
and Christoffel function $K(\mathbf{z})$ (\ref{ChristoffelLike})
is a very promising one,
it still has some limitations.

Consider a ${\mathcal D}=1$ example:
let a datasample (\ref{mlproblemVector})
has $f_0=x_0$ for all $l=1\dots M$.
Then Gram matrix $\Braket{z_iz_j}$ is degenerated.
When attributes regularization is applied --- it will remove
either $f_0$ or $x_0$ from $\mathbf{z}$,
thus the resulting $K(\mathbf{z})\Big|_{\mathbf{x}}$ depends
on attributes regularization:
a polynomial $1/K(\mathbf{z})\Big|_{\mathbf{x}}$ on $\mathbf{f}$
is different, thus $\Braket{\mathbf{f} K(\mathbf{z})\Big|_{\mathbf{x}}}$
produces the result depending on the regularization.
An ultimate example of this situation
is: for $k=0\dots n-1$, let $f_k=x_k$ for all $l=1\dots M$ with $n=m$.
In this case Gram matrix has two copies
of exactly the same attributes
and what combination
of them propagate to the final set of attributes
depends on regularization.
For example if $x_k$ are selected and $f_k$ are dropped then
$ K(\mathbf{z})\Big|_{\mathbf{x}}$
is a constant
and
$\Braket{\mathbf{f} K(\mathbf{z})\Big|_{\mathbf{x}}}$
is $\mathbf{x}$--independent.
Such a regularization--dependent answer
cannot be a solid foundation to ML classification problem,
a regularization--independent solution is required.

Consider two Gram matrices $\Braket{x_kx_{k^{\prime}}}$ and $\Braket{f_jf_{j^{\prime}}}$
with attributes possibly ``producted'' (\ref{xmulti})
to ${\mathcal D}_x$ and ${\mathcal D}_f$.
It's ``gauge transformation'' is:
\begin{subequations}
\label{gaugeXF}
\begin{align}
x^{\prime}_{k}&=\sum\limits_{k^{\prime}=0}^{n-1}T_{kk^{\prime}} x_{k^{\prime}} \label{gaugeX}\\
f^{\prime}_{j}&=\sum\limits_{j^{\prime}=0}^{m-1}T_{jj^{\prime}} f_{j^{\prime}} \label{gaugeF}
\end{align}
\end{subequations}
There are no $\mathbf{x}\Leftrightarrow\mathbf{z}$
``cross'' terms as when we were working with
the combined $\mathbf{z}$,
this makes the solution  regularization--independent.

Consider the simplest practical solution.
Let $x_k$ attributes being regularized
and ``producted'' (\ref{xmulti})
to a degree ${\mathcal D}$. The $\mathbf{f}$ attributes are untransformed.
The Radon--Nikodym
interpolation solution (\ref{RNfsolution})
is directly applicable:
\begin{align}
  \mathbf{f}_{RN}(\mathbf{x})&=\frac{\sum\limits_{l,j,k,i=0}^{n-1}x_lG^{-1}_{lj}\Braket{x_j|\mathbf{f}|x_k}G^{-1}_{ki}x_i}
  {\sum\limits_{j,k=0}^{n-1}x_jG^{-1}_{jk}x_k}
  \label{RNfsolutionVector}
\end{align}
This ``vector'' type of solution
to distribution to distribution regression problem
(that was obtained back in \cite{2015arXiv151109058G}) is just
(\ref{RNfsolution}) applied to every component of $\mathbf{f}$.
As we discussed in Section \ref{RNapproach} and demonstrated
in the Appendix \ref{NominalExample}, such a solution, while being
a good one to an interpolation problem, leads to data
overfitting when applied to a classification problem.
We need to use the posterior (\ref{Pposterior}) distribution
weights to obtain
an analogue of $f_{RNW}(\mathbf{x})$ (\ref{RNWfsolutionpsi}),
but \textbf{without} generalized eigenvalue problem on $f$,
as the $\mathbf{f}$ is now a vector.
This is feasible if we go from ``regular'' average
to Christoffel function average of Section \ref{UnsupervisedLearning}.
All density matrix averages posses the duality property\cite{ArxivMalyshkinLebesgue}:
\begin{align}
  \mathrm{Spur}\, \|f|\rho_K\| &=
  \sum\limits_{i=0}^{n-1} \lambda^{[i]}_K \Braket{\psi^{[i]}_K|f|\psi^{[i]}_K}=
  \sum\limits_{i=0}^{n-1} \lambda^{[i]}_f \Braket{\psi^{[i]}_f|\rho_K|\psi^{[i]}_f}
  \label{dualismA}
\end{align}
Thus, for a vector $\mathbf{f}$, where the pairs
$\left(\lambda^{[i]}_f;\Ket{\psi^{[i]}_f}\right)$
do not exist, obtain in $\Ket{\psi^{[i]}_K}$ basis:
\begin{align}
  \mathbf{f}_{RNW}(\mathbf{x})&=\frac{
    \sum\limits_{i=0}^{n-1} \lambda^{[i]}_K \left[\psi^{[i]}_K(\mathbf{x})\right]^2
    \Braket{\psi^{[i]}_K|\mathbf{f}|\psi^{[i]}_K}
  }
  {
    \sum\limits_{i=0}^{n-1} \lambda^{[i]}_K \left[\psi^{[i]}_K(\mathbf{x})\right]^2
  }
    \label{RNWfsolutionVector}
\end{align}
This is the simplest practical solution\footnote{
  One can also try the $\mathbf{f}_{RN}(\mathbf{x})$ from (\ref{RNfsolutionVector})
  with $\Braket{x_j|K(\mathbf{x})|x_k}$ and
  $\Braket{x_j|\mathbf{f}(\mathbf{x})K(\mathbf{x})|x_k}$
  used instead of $G_{jk}=\Braket{x_j|x_k}$ and $\Braket{x_j|\mathbf{f}|x_k}$.
  }
to a classification problem with vector class label $\mathbf{f}$.
It uses unsupervised learning basis $\Ket{\psi^{[i]}_K}$
of generalized eigenvalue problem (\ref{GEVK})
to solve the problem with a vector class label $\mathbf{f}$.
The solution (\ref{RNWfsolutionVector}) assumes
every component of vector $\mathbf{f}$
is diagonal in the basis $\Ket{\psi_K^{[i]}}$. This is not generally the case,
but allows to build a single classificator for a vector class label $\mathbf{f}$
instead of constructing an
individual classificator for every $f_j$ component.
The option
\texttt{\seqsplit{--flag\_assume\_f\_is\_diagonal\_in\_christoffel\_function\_basis=true}} of the provided software (see Appendix \ref{RN} below)
builds such a classifier.
This ``same $\Ket{\psi_K^{[i]}}$ basis for all $f_j$'' classifier typically has worse quality that
the one built in $\Ket{\psi^{[i]}}$ basis corresponding to
an individual scalar class label $f_j$

The approach of two
Gram matrices $\Braket{x_kx_{k^{\prime}}}$, $k,k^{\prime}=0\dots n-1$
and $\Braket{f_jf_{j^{\prime}}}$, $j,j^{\prime}=0\dots m-1$ 
without ``mixed'' terms  $\Braket{x_kf_j}$ in basis
allows to obtain a ``relative frequency'' characteristic,
a density of state type of solution.
Consider ${\mathcal R}$, the ratio of two Christoffel functions:
\begin{align}
  K(\mathbf{f}(\mathbf{x})) &={\mathcal R}\cdot K(\mathbf{x})
  \label{Rfrac}\\
    {\mathcal R}&=\frac{\sum\limits_{k,k^{\prime}=0}^{n-1}\alpha_k \Braket{x_k|K(\mathbf{f(\mathbf{x})})|x_{k^{\prime}}} \alpha_{k^{\prime}}}
  {\sum\limits_{k,k^{\prime}=0}^{n-1}\alpha_k \Braket{x_k|K(\mathbf{x})|x_{k^{\prime}}} \alpha_{k^{\prime}}}
  \label{Rquad}
\end{align}
which is an estimator of Radon--Nikodym derivative\cite{BarrySimon}.
The ${\mathcal R}$ is a dimensionless ``relative frequency'':
how often a given realization of vector $\mathbf{f}$
corresponds to a given realization of vector $\mathbf{x}$
in (\ref{mlproblemVector}) sample.
The $K(\mathbf{x})$ and $K(\mathbf{f})$
are Christoffel functions calculated on $\mathbf{x}$ and
$\mathbf{f}$ portion of (\ref{mlproblemVector}) data,
possibly regularized and ``producted''.
The $1/K(\mathbf{x})$ and $1/K(\mathbf{f})$
are positive polynomials on $x_k$ and $f_j$ components respectively.

To obtain the distribution of ${\mathcal R}$
multiply left- and right- hand side of (\ref{Rfrac}) by
$\psi^2(\mathbf{x})$ and integrate it over all $l=1\dots M$
observations of (\ref{mlproblemVector}) datasample, obtain (\ref{Rquad}).
The calculation of $\Braket{x_k|K(\mathbf{f(\mathbf{x})})|x_{k^{\prime}}}$
matrix elements is no different from
the one performed in (\ref{GEVK}): use (\ref{ChristoffelLike}) expression,
but now in $\mathbf{f}$--space.
A familiar generalized eigenvalue problem is then:
\begin{align}
  \sum\limits_{k^{\prime}=0}^{n-1}
  \Braket{x_k|K(\mathbf{f(\mathbf{x})})|x_{k^{\prime}}}
  \alpha^{[i]}_{k^{\prime}} &=
  \lambda_{\mathcal R}^{[i]} \sum\limits_{k^{\prime}=0}^{n-1} \Braket{x_k|K(\mathbf{x})|x_{k^{\prime}}}  \alpha^{[i]}_{k^{\prime}}
  \label{GEVKR} \\ 
 \psi_{\mathcal R}^{[i]}(\mathbf{x})&=\sum\limits_{k=0}^{n-1} \alpha^{[i]}_k x_k
 \label{psiCKR}
\end{align}
Obtained $\lambda_{\mathcal R}^{[i]}$ is a spectrum of ``relative frequency''.
In $\Ket{\psi_{\mathcal R}^{[i]}}$ state there are $\lambda_{\mathcal R}^{[i]}$ time
more $\mathbf{f}$ observations
than $\mathbf{x}$ observations.
The matrices $\Braket{x_k|K(\mathbf{f(\mathbf{x})})|x_{k^{\prime}}}$
and $\Braket{x_k|K(\mathbf{x})|x_{k^{\prime}}}$
are $n\times n$ matrices calculated from a training datasample.
The knowledge is accumulated in their spectrum.
When evaluating a testing dataset the simplest usage
of (\ref{Rquad}) is this: for a given $\mathbf{x}$,
how often/seldom we see an $\mathbf{f}$?
The answer is (\ref{Rquad}) with localized $\alpha_k=\sum_{k^{\prime}=0}^{n-1} G^{-1}_{kk^{\prime}} x_{k^{\prime}}$ or, when written in (\ref{psiCKR}) basis
\begin{align}
  {\mathcal R}(\mathbf{x})&=\frac{\sum\limits_{i=0}^{n-1}\lambda_{\mathcal R}^{[i]} \left[\psi_{\mathcal R}^{[i]}(\mathbf{x})\right]^2}{\sum\limits_{i=0}^{n-1} \left[\psi_{\mathcal R}^{[i]}(\mathbf{x})\right]^2}
  \label{RquadLoc}
\end{align}
While the (\ref{RNWfsolutionVector}) is $\mathbf{f}$--value
predictor, 
the ${\mathcal R}$ is ``relative frequency'' estimator, an important
characteristic when considering a vector--to--vector type of mapping.

\subsection{\label{ErrorF}A Vector--Valued Class Label: Error Estimation}
The vector--value estimators 
(\ref{RNfsolutionVector})
and (\ref{RNWfsolutionVector})
are an estimation of $\mathbf{f}$ by averaging class label
$\mathbf{f}^{(l)}=(f_0,f_1,\dots,f_j,\dots,f_{m-1})^{(l)}$ from (\ref{ZVector})
with a $\mathbf{x}$-- dependent positive weight $W_{\mathbf{x}}(\mathbf{x}^{(l)})$:
\begin{align}
  \mathbf{f}(\mathbf{x})&=\frac{\sum\limits_{l=1}^{M} W_{\mathbf{x}}(\mathbf{x}^{(l)}) \mathbf{f}^{(l)}}
  {\sum\limits_{l=1}^{M} W_{\mathbf{x}}(\mathbf{x}^{(l)})}
  \label{favefaiging} \\
  \Braket{1}_{W_{\mathbf{x}}}&=\sum\limits_{l=1}^{M} W_{\mathbf{x}}(\mathbf{x}^{(l)})
  \label{1W}
\end{align}
What is the best way to estimate an error of a solution of this type?
A ``traditional''  approach would be to consider a standard deviation
type of answer $\Braket{\left(f-\overline{f}\right)^2}$, a variation
of $\mathbf{f}$ components relatively their average value.
This solution can be obtained from Gram matrix in $\mathbf{f}$--space (with some complications
because of vector class label $\mathbf{f}$):
\begin{align}
  G_{jk}&=\Braket{f_jf_k}_{W_{\mathbf{x}}}=\sum\limits_{l=1}^{M} W_{\mathbf{x}}(\mathbf{x}^{(l)}) f^{(l)}_jf^{(l)}_k & j,k=0\dots m-1
  \label{Gf}
\end{align}
As we discussed in \cite{malyshkin2018spikes} and then earlier in this paper all
standard deviation error estimators cannot be applied
to non--Gaussian data, thus they have a limited applicability domain.
A much better estimator can be constructed from the Christoffel function.
Consider Christoffel function in $\mathbf{f}$--space $K_{W_\mathbf{x}}(\mathbf{f})$,
obtained from Gram matrix (\ref{Gf})
as $1/K_{W_\mathbf{x}}(\mathbf{f})={\sum_{j,k=0}^{m-1}f_jG^{-1}_{jk}f_k}$,
exactly as we did in (\ref{ChristoffelLike})
in $\mathbf{x}$--space\footnote{
  To calculate Christoffel function properly
  there always should be a constant present in the $(f_0,f_1,\dots,f_j,\dots,f_{m-1})$ basis space,
  if it does not have one -- add an attribute $f_{m}=1$ to the basis.  
  If $G_{jk}$ is degenerated the vector
  $(f_0,f_1,\dots,f_j,\dots,f_{m-1})$ should be regularized according to
  Appendix \ref{regularization} with the replacement $x_j \to f_j$. Described there regularization algorithms
  always add a constant to the basis if it does not have one.}.
Consider the best possible situation when (\ref{favefaiging})
has no variation, i.e. the averaging gives exact values. The support
of this measure is then a single point $\mathbf{f}$ from (\ref{favefaiging})
(compare with a Gaussian quadrature in case when a single node
has a dominantly large weight).
When a prediction is not perfect we have a variation of $\mathbf{f}^{(l)}$  around average.
Exactly as we did above, instead of considering a variation in $\mathbf{f}$--space,
consider the support of a measure, a ``Lebesgue'' style approach.
The total measure is $\Braket{1}_{W_\mathbf{x}}$, the support of $\mathbf{f}$--localized state is $K_{W_\mathbf{x}}(\mathbf{f})$,
their difference gives error estimation:
\begin{align}
  \mathrm{Error}&= \Braket{1}_{W_\mathbf{x}} - K_{W_\mathbf{x}}(\mathbf{f}) \label{Error} \\
  \mathrm{Error}_{rel}&=\frac{\mathrm{Error}}{\Braket{1}_{W_\mathbf{x}}}=1-\frac{K_{W_\mathbf{x}}(\mathbf{f})}{\Braket{1}_{W_\mathbf{x}}}
  \label{ErrorRel}
\end{align}
Error estimator (\ref{Error}) has a dimension of weight (number of observations).
It has the meaning of the difference
between total measure and the measure of $\mathbf{f}$--localized state.
It is gauge invariant relatively (\ref{gaugeXF}).

Even when a predictor (in a form of $\mathbf{x}$-- dependent positive weight $W_{\mathbf{x}}(\mathbf{x})$)
does not exist we can still obtain an information of how well
a vector in $\mathbf{f}$-space can be recovered
from $\mathbf{x}$-space.
In scalar case $\mathbf{f}=f$ the simplistic solution
to the problem is the aforementioned $L^2$ norm (\ref{norm2regrf}):
if standard deviation is zero then $f$ can be completely recovered from the value of
$\mathbf{x}$.
However, this solution, besides depending on the scale of $f$,
is problematically to generalize to a vector $\mathbf{f}$.

We can construct an original solution to vector $\mathbf{f}$ from three
matrices: $\Braket{f_{j^{\prime}}f_{k^{\prime}}}$ (the (\ref{Gf}) with $W_{\mathbf{x}}=1$),
$\Braket{x_{j}x_{k}}$, and $\Braket{x_jf_{k^{\prime}}}$. The first two are Gram matrices
in $\mathbf{f}$-
and $\mathbf{x}$-
space respectively:
\begin{align}
  G^{\mathbf{f}}_{j^{\prime}k^{\prime}}&= \Braket{f_{j^{\prime}}f_{k^{\prime}}} & j^{\prime},k^{\prime}=0\dots m-1 \label{GfW1} \\
  G^{\mathbf{x}}_{jk}&= \Braket{x_{j}x_{k}} & j,k=0\dots n-1 \label{Gx} \\
  G^{\mathbf{x}\mathbf{f}}_{jk^{\prime}}&= \Braket{x_jf_{k^{\prime}}} & j=0\dots n-1 ; k^{\prime}=0\dots m-1 \label{Gxf}
\end{align}
In scalar $f$ case we have  $m=2$ or greater:
\begin{align}
  \mathbf{f}&=(1,f) &
  f_0=1;
  f_1=f;
  m=2
  \label{scalarf} \\
    \mathbf{f}&=(1,f,f^2) &
  f_0=1;
  f_1=f;
  f_2=f^2;
  m=3 \nonumber \\
    \mathbf{f}&=(1,f,f^2,f^3) &
  f_0=1;
  f_1=f;
  f_1=f^2;
  f_3=f^3;
  m=4
  \nonumber
\end{align}
a constant should always present in the basis (both in $\mathbf{f}$ and $\mathbf{x}$).
A criterion of how well $\mathbf{f}$ can be recovered from $\mathbf{x}$
is to compare the matrices $\Braket{f_{j^{\prime}}f_{k^{\prime}}}$ and
$\Braket{f_{j^{\prime}}(\mathbf{x})f_{k^{\prime}}(\mathbf{x})}$;
the $f_{j^{\prime}}$ is exact value and
the $f_{j^{\prime}}(\mathbf{x})$ is obtained from (\ref{regrfsolution}) projection
of $\mathbf{f}$ on $\mathbf{x}$-space:
\begin{align}
  \mathbf{f}(\mathbf{x})&=\mathrm{Proj}^{(\mathbf{f\to x})} \mathbf{f} \label{fpoj} \\
  f_{j^{\prime}}(\mathbf{x})&=
  \sum\limits_{j,k=0}^{n-1}x_jG^{\mathbf{x};\,-1}_{jk}\Braket{f_{j^{\prime}}x_k}
  \label{fjregression} \\
  \Braket{f_{j^{\prime}}(\mathbf{x})f_{k^{\prime}}(\mathbf{x})}&=
  \sum\limits_{j,k=0}^{n-1}\Braket{f_{j^{\prime}}x_j}G^{\mathbf{x};\,-1}_{jk}\Braket{f_{k^{\prime}}x_k}
  =\sum\limits_{j,k=0}^{n-1}
  G^{\mathbf{x}\mathbf{f}}_{jj^{\prime}}
  G^{\mathbf{x};\,-1}_{jk}
  G^{\mathbf{x}\mathbf{f}}_{kk^{\prime}}
  \label{fxfx}
\end{align}
Here $G^{\mathbf{x};\,-1}_{jk}$ is an inverse of $G^{\mathbf{x}}_{jk}$ from (\ref{Gx}).
The non--negative $m\times m$ symmetric matrices\footnote{
  If the matrix $\Braket{f_{j^{\prime}}f_{k^{\prime}}}$ is not positive ---
  apply Appendix \ref{regularization} regularization first.
  }:
$\Braket{f_{j^{\prime}}(\mathbf{x})f_{k^{\prime}}(\mathbf{x})}$ (Eq. (\ref{fxfx}))
and $\Braket{f_{j^{\prime}}f_{k^{\prime}}}$ (Eq. (\ref{GfW1}))
coincide if $\mathbf{f}$ is a subspace of $\mathbf{x}$;
both represent the $\mathbf{f}$-space: the former is projected on $\mathbf{x}$,
the later is calculated directly.

Solve generalized eigenproblem with these two matrices
in left- and right- hand side respectively, exactly as in (\ref{GEV}):
\begin{align}
  \sum\limits_{k^{\prime}=0}^{m-1} \Braket{f_{j^{\prime}}(\mathbf{x})f_{k^{\prime}}(\mathbf{x})} \alpha^{[i]}_{k^{\prime}} &=
  \lambda^{[i]} \sum\limits_{k^{\prime}=0}^{m-1} \Braket{ f_{j^{\prime}}|f_{k^{\prime}}} \alpha^{[i]}_{k^{\prime}}
  \label{GEVfxf} \\
  \sum\limits_{k^{\prime}=0}^{m-1}
  \sum\limits_{j,k=0}^{n-1}
   G^{\mathbf{x}\mathbf{f}}_{jj^{\prime}}
      G^{\mathbf{x};\,-1}_{jk}
      G^{\mathbf{x}\mathbf{f}}_{kk^{\prime}}
      \alpha^{[i]}_{k^{\prime}}
      &=
      \lambda^{[i]} \sum\limits_{k^{\prime}=0}^{m-1} G^{\mathbf{f}}_{j^{\prime}k^{\prime}} \alpha^{[i]}_{k^{\prime}}
      \nonumber
\end{align}
If $\mathbf{f}$-space is a subspace of $\mathbf{x}$-space
then all $i=0\dots m-1$ eigenvalues $\lambda^{[i]}$ are equal to $1$ and their sum is equal to matrix $\Braket{ f_{j^{\prime}}|f_{k^{\prime}}}$ rank $m$.
Otherwise the difference represents an error: how big is the remaining error after
projecting $\mathbf{f}$-space on $\mathbf{x}$-space:
\begin{align}
  \mathrm{Error}_{rank}&= m - \sum\limits_{i=0}^{m-1} \lambda^{[i]}
  = m - \sum\limits_{j,k=0}^{m-1} \Braket{f_{j}(\mathbf{x})f_{k}(\mathbf{x})} G^{\mathbf{f};\,-1}_{kj}
   \label{ErrorRank}
\end{align}
This error is gauge--invariant relatively (\ref{gaugeXF}), it is dimensionless
and represents how well $\mathbf{f}$-space can be projected on  $\mathbf{x}$-space.
It can be viewed as a gauge--invariant ``squared multi--dimensional correlation''
  between $\mathbf{f}(\mathbf{x}^{(l)})$ and $\mathbf{f}^{(l)}$, $l=1\dots M$.
  If $n=m=2$ we have: $\mathbf{x}=(1,x)$; $\mathbf{f}=(1,f)$
  then (\ref{GEVfxf}) has
  the maximal eigenvalue $\lambda^{[1]}=1$ because a constant presents in both bases,
  and minimal eigenvalue is equal
  to regular correlation between $x$ and $f$ squared: $\lambda^{[0]}=\rho^2(x,f)$.

The (\ref{ErrorRank}) can also be calculated directly using matrix $\mathrm{Spur}$,
without solving a generalized eigenvalue problem.
It is a ``rank--difference'' error estimator
what makes it not always convenient in practical ML applications. The most convenient
error estimator in ML is of ``coverage'' type: how many observations are
correctly classified (or misclassified). This error can be obtained using (\ref{fjregression})
projection and Christoffel function technique we applied in Section \ref{ClusteringErrorF}
below to the Low-Rank Representation(LRR) problem.
The solution is straightforward:
\begin{itemize}
\item Construct a
$\psi_{\mathbf{g}}(\mathbf{f})$ state,
localized at $\mathbf{f}=\mathbf{g}$,
it is exactly (\ref{psiYlocalized})
with a replace $\mathbf{x}\to \mathbf{f}$ ; $\mathbf{y}\to \mathbf{g}$; $G\to G^{\mathbf{f}}$, see Eq. (\ref{psiGflocalizedAppendix}).
\item In every $\mathbf{g}=\mathbf{f}^{(l)}$ point
  we have $\Braket{\psi^2_{\mathbf{f}^{(l)}}}=1$,
  exactly as in full basis  expansion (\ref{fullbasisexpansion}).
\item If one, instead of $\psi_{\mathbf{g}}(\mathbf{f})$, take it's projection (\ref{fjregression})
  to $\mathbf{x}$-space --- the value (\ref{fxerrorWOnePoint})
  can be lower than $1$, similarly to (\ref{fullbasisexpansionpartial}).
  Then sum it over all $l=1 \dots M$ sample observations to obtain the number of covered points.
  The $\mathrm{Error}$
  is then:
  \begin{align}
    \varpi(\mathbf{g})&=
    \Braket{\left[\mathrm{Proj}^{(\mathbf{f\to x})} \psi_{\mathbf{g}}\right]^2}=
 \frac{\sum\limits_{j,k=0}^{n-1}\sum\limits_{s^{\prime},j^{\prime},k^{\prime},t^{\prime}=0}^{m-1}
      g_{s^{\prime}}
      G^{\mathbf{f};\,-1}_{s^{\prime}j^{\prime}}
      G^{\mathbf{x}\mathbf{f}}_{jj^{\prime}}
      G^{\mathbf{x};\,-1}_{jk}
      G^{\mathbf{x}\mathbf{f}}_{kk^{\prime}}
      G^{\mathbf{f};\,-1}_{k^{\prime}t^{\prime}}
      g_{t^{\prime}}
    }
         {
           \sum\limits_{j^{\prime},k^{\prime}=0}^{m-1} g_{j^{\prime}}G^{\mathbf{f};\,-1}_{j^{\prime}k^{\prime}}g_{k^{\prime}}
         } \label{fxerrorWOnePoint} \\    
    \mathrm{Error}&=\Braket{1}-\sum\limits_{l=1}^{M}
          \omega^{(l)}
    \varpi(\mathbf{f}^{(l)})
         \label{fxerrorW}
  \end{align}
\end{itemize}
The (\ref{fxerrorW}) is an analogue of (\ref{Error}) with no predictor available,
this is a characteristics of the data, not of a predictor,
the sum of basis projection successes
$\varpi(\mathbf{f}^{(l)})$ in every observation point $l$
with the weight $\omega^{(l)}$.
This expression can be generalized with
an  operator ${\mathcal U}$ in $\mathbf{x}$-space
converting $\psi_{\mathbf{x}^{(l)}}(\mathbf{x})$
to some other function in $\mathbf{x}$-space  $\Ket{\psi(\mathbf{x})}=
\Ket{{\mathcal U}|\psi_{\mathbf{x}^{(l)}}(\mathbf{x})}$
and only then projecting the result
to
actual realization $\psi_{\mathbf{f}^{(l)}}(\mathbf{f})$
in $\mathbf{f}$-space:
\begin{align}
  \mathrm{Error}&= \Braket{1}
  -\sum\limits_{l=1}^{M}
  \omega^{(l)}
  \left|\Braket{\psi_{\mathbf{f}^{(l)}}|{\mathcal U}|\psi_{\mathbf{x}^{(l)}}}\right|^2   
  \label{fxerrorWexpanded}
\end{align}
This error is
the number of misclassified observations
for specific predictor $\|{\mathcal U}\|$, it is always
greater than the error (\ref{fxerrorW}).
The (\ref{fxerrorW}) corresponds to
$\Ket{{\mathcal U}|\psi_{\mathbf{x}^{(l)}}}$ (a single vector in $\mathbf{x}$-space)
being replaced by
direct projection
to a full orthogonal basis $\Ket{\psi^{[i]}}$ in $\mathbf{x}$-space,
similar to (\ref{fullbasisexpansionpartial}) and (\ref{psiphicomplete}):
\begin{align}
  \varpi(\mathbf{g})&=
  \sum\limits_{i=0}^{n-1}
  \Braket{\psi_{\mathbf{g}}|\psi^{[i]}}^2
  &
    1 \ge \varpi(\mathbf{g})
  \label{fxerrorWOnePointFullBasis}
\end{align}
The $\varpi(\mathbf{g})$ determines how well a
localized in $\mathbf{f}$-space state $\psi_{\mathbf{g}}(\mathbf{f})$
can be projected to $\mathbf{x}$-space basis.
This criterion is then tested
for all $l=1\dots M$ observation points,
 For the reason of testing
the entire sample of $M$ points, not just $n$ basis functions,
the Error (\ref{fxerrorW}) is
an estimation of the best possible predictor performance,
thus it is useful as a bound (\ref{dpfcompare})
for a predictor of (\ref{fxerrorWexpanded}) form.

The Error can be spectrally expanded. Introduce
\begin{align}
  \Braket{f_{j}|K^{(\mathbf{f})}|f_{k}}
  &=\sum\limits_{l=1}^{M}
       \omega^{(l)}
  \frac{f^{(l)}_{j}f^{(l)}_{k}}
       { \sum\limits_{j^{\prime},k^{\prime}=0}^{m-1} f^{(l)}_{j^{\prime}}G^{\mathbf{f};\,-1}_{j^{\prime}k^{\prime}}f^{(l)}_{k^{\prime}}}
        &j,k=0\dots m-1 
          \label{KfffxerrorW}
\end{align}
Which is exactly Christoffel function matrix (\ref{GEVK}), but in
$\mathbf{f}$-space. Then (\ref{fxerrorW}) can be expressed
as matrix spur (\ref{ErrorKfsum}):
 \begin{align}
K^{(\mathbf{f\to x})}_{jk}&=
\sum\limits_{k^{\prime},t^{\prime},s^{\prime},j^{\prime}=0}^{m-1}
      G^{\mathbf{x}\mathbf{f}}_{kk^{\prime}}
      G^{\mathbf{f};\,-1}_{k^{\prime}t^{\prime}}
      \Braket{f_{t^{\prime}}|K^{(\mathbf{f})}|f_{s^{\prime}}}
      G^{\mathbf{f};\,-1}_{s^{\prime}j^{\prime}}
      G^{\mathbf{x}\mathbf{f}}_{jj^{\prime}}
       &j,k=0\dots n-1 \label{Kftoxsum}\\
\mathrm{Error}&=\Braket{1}-
\sum\limits_{j,k=0}^{n-1}
K^{(\mathbf{f\to x})}_{jk}G^{\mathbf{x};\,-1}_{kj} =
\Braket{1}-
\mathrm{Spur} K^{(\mathbf{f\to x})} G^{\mathbf{x};\,-1}
\label{ErrorKfsum}
 \end{align}
From which immediately follows, that if we
solve generalized eigenproblem
with $K^{(\mathbf{f\to x})}_{jk}$ and $G^{\mathbf{x}}_{jk}=\Braket{x_{j}x_{k}}$ matrices
in left- and right- hand side respectively,
the $\mathrm{Error}$ can be spectrally expanded:
\begin{align}
  \sum\limits_{k=0}^{n-1} K^{(\mathbf{f\to x})}_{jk}
  \alpha^{[i]}_{k} &=
  \lambda^{[i]} \sum\limits_{k=0}^{n-1} \Braket{x_{j}x_{k}} \alpha^{[i]}_{k}
\label{GEVKftoXfxf} \\
\mathrm{Error}&=\Braket{1}-\sum\limits_{i=0}^{n-1}\lambda^{[i]}
\label{ErrorKfsumSpectral}
\end{align}
The (\ref{ErrorKfsumSpectral}) is a spectral decomposition of (\ref{fxerrorW}),
it has at most $m$ non--zero eigenvalues (the rank of (\ref{Kftoxsum}) is $m$ or lower, we also assume $m\le n$).
If $\mathbf{f}$ belongs to a subspace of $\mathbf{x}$ then
the sum of these $m$ eigenvalues in (\ref{ErrorKfsumSpectral})
is equal to $\Braket{1}$.
The eigenvectors corresponding to a few ($m$ or  lower) maximal eigenvalues
of (\ref{GEVKftoXfxf}) is
the solution to vector class label classification problem
target basis (not the problem itself).

Consider a simple demonstrative solution. Let us project
$\psi_{\mathbf{g}}(\mathbf{f})$ to $\psi_{\mathbf{f}_{LS}(\mathbf{x})}(\mathbf{f})$
to obtain a joint probability estimator:
what is the probability\footnote{
  The coverage of the predictor (\ref{projestimation}) at $\mathbf{y}$ can be estimated from the value of $1/\mathrm{Norm}^2(\mathbf{y})$,
  similar to using Christoffel function $K(\mathbf{y})$
  for estimation of the support of the measure
  of localized at $\mathbf{x}=\mathbf{y}$ state.} of outcome $\mathbf{g}$
given input vector  $\mathbf{y}$
if $\mathbf{f}_{LS}(\mathbf{x})$ model is assumed.
\begin{align}
  \psi_{\mathbf{f}_{LS}(\mathbf{y})}(\mathbf{f})&=
  \frac{1}{\mathrm{Norm}(\mathbf{y})} \sum\limits_{j,k=0}^{n-1}\sum\limits_{j^{\prime},k^{\prime}=0}^{m-1}
  y_jG^{\mathbf{x};\,-1}_{jk} G^{\mathbf{x}\mathbf{f}}_{kj^{\prime}} G^{\mathbf{f};\,-1}_{j^{\prime}k^{\prime}} f_{k^{\prime}}
  \label{psiasymm} \\
  \mathrm{Norm}^2(\mathbf{y})&=
   \sum\limits_{j,k,s,t=0}^{n-1}\sum\limits_{j^{\prime},k^{\prime}=0}^{m-1}
    y_jG^{\mathbf{x};\,-1}_{jk} G^{\mathbf{x}\mathbf{f}}_{kj^{\prime}}
    G^{\mathbf{f};\,-1}_{j^{\prime}k^{\prime}} G^{\mathbf{x}\mathbf{f}}_{sk^{\prime}}
    G^{\mathbf{x};\,-1}_{st}y_t \label{normalizepsiasymm} \\
\mathrm{Prob}(\mathbf{g}|\mathbf{y})&=\Braket{\psi_{\mathbf{f}_{LS}(\mathbf{y})}(\mathbf{f})|\psi_{\mathbf{g}}(\mathbf{f})}^2
  =\frac{
    \left[
    \sum\limits_{j,k=0}^{n-1}\sum\limits_{j^{\prime},k^{\prime}=0}^{m-1}
    y_jG^{\mathbf{x};\,-1}_{jk} G^{\mathbf{x}\mathbf{f}}_{kj^{\prime}} G^{\mathbf{f};\,-1}_{j^{\prime}k^{\prime}} g_{k^{\prime}}
    \right]^2
  }{
    \mathrm{Norm}^2(\mathbf{y})
    \sum\limits_{j^{\prime},k^{\prime}=0}^{m-1} g_{j^{\prime}}G^{\mathbf{f};\,-1}_{j^{\prime}k^{\prime}}g_{k^{\prime}}  
  }
  \label{projestimation} \\
  \widetilde{\mathrm{Error}}&=\Braket{1}-
  \sum\limits_{l=1}^{M}
  \omega^{(l)}
  \mathrm{Prob}(\mathbf{f}^{(l)}|\mathbf{x}^{(l)})
  \label{fxerrorWJointProb} 
\end{align}
This solution has a form of conditional
probability (\ref{projestimation}) which can be used
to introduce a predictor-specific
error estimator $\widetilde{\mathrm{Error}}$.
Whereas the ``maximal coverage''
estimator (\ref{fxerrorW})
estimates data recoverability without constructing a predictor,
the estimator (\ref{fxerrorWJointProb}) estimates specific simple
prediction of least squares type;
usual least squares property holds:
it is zero if $\mathbf{f}$ is a subspace of $\mathbf{x}$.
This estimator can be spectrally decomposed
only at some given $\mathbf{x}$,
this makes it's properties (\ref{ChristoffelFunMaxVector}) related.
Introduce $b_{j^{\prime}}(\mathbf{y})$:
\begin{align}
b_{j^{\prime}}(\mathbf{y})&=
\frac{1}{\mathrm{Norm}(\mathbf{y})}
 \sum\limits_{j,k=0}^{n-1}
 y_jG^{\mathbf{x};\,-1}_{jk} G^{\mathbf{x}\mathbf{f}}_{kj^{\prime}}
 \label{bj} \\
 \sum\limits_{k^{\prime}=0}^{m-1}
 b_{j^{\prime}}(\mathbf{y})b_{k^{\prime}}(\mathbf{y}) \alpha^{[i]}_{k^{\prime}}&=
 \lambda^{[i]} \sum\limits_{k^{\prime}=0}^{m-1}
 G^{\mathbf{f}}_{j^{\prime}k^{\prime}}
 \alpha^{[i]}_{k^{\prime}}
 \label{bevproblem}
\end{align}
Then (\ref{bevproblem}) has a single non--zero eigenvalue
$\lambda^{[m-1]}= \sum_{j,k=0}^{m-1} b_{j}G^{\mathbf{f};\,-1}_{jk}b_{k}  
= 1$, which is the maximal value of (\ref{projestimation}).
While
vector--to--vector prediction models
are not implemented in the \href{http://www.ioffe.ru/LNEPS/malyshkin/code_polynomials_quadratures.zip}{provided software} yet, a reference
unit test for (\ref{projestimation}) and (\ref{fxerrorWJointProb})
is available therein;
it can be run with random data.
The calculations require only matrix algebra: the (\ref{projestimation}) is a ratio of a quadratic form squared and a product of two quadratic forms. Hence,
as with any Radon--Nikodym type of solution, it tends to a
constant (not to infinity like e.g. least squares)
when $\mathbf{y}\to\infty$ or $\mathbf{g}\to\infty$.
See \texttt{\seqsplit{SolutionVectorXVectorF.java:evaluateAt(final\, double\, [] X)}} for simple examples.
The (\ref{projestimation}) estimates conditional probability,
not the value of most probable outcome.
A familiar least squares (\ref{fjregression}) estimation of $\mathbf{f}$
given $\mathbf{x}$
can be obtained from:
\begin{align}
  &\mathbf{f}_{\mathrm{LS}}(\mathbf{x})=\mathrm{Norm}(\mathbf{x})\mathbf{b}(\mathbf{x})
  &
  f_{\mathrm{LS}\,j^{\prime}}(\mathbf{x})&=
   \sum\limits_{j,k=0}^{n-1}
 x_jG^{\mathbf{x};\,-1}_{jk} G^{\mathbf{x}\mathbf{f}}_{kj^{\prime}}  \label{fboutcome} \\
  &\mathrm{Prob}(\mathbf{f}_{\mathrm{LS}}(\mathbf{x})|\mathbf{x})=1
  \label{bmax}
\end{align}
The (\ref{projestimation}) is just a
simple example of conditional probability estimator,
a demonstration, that even with least squares na\"{\i}ve form
(\ref{fboutcome})
there exists
a big improvement when we consider
a conditional probability estimation
instead of typically considered value estimation.
A general form a ``unitary'' type of conditional probability
estimator 
is discussed below in Appendix \ref{nonunitarydynamics},

All considered estimators
are gauge--invariant relatively \ref{gaugeXF}).
The main idea behind these estimators is straightforward:
consider localized at $\mathbf{f}=\mathbf{g}$ state
$\psi_{\mathbf{g}}(\mathbf{f})$ (the (\ref{psiYlocalized}) in $\mathbf{f}$-space),
project it to some $\mathbf{x}$-dependent vector space
(in the simplistic case it is just (\ref{fjregression}) direct projection,
in most general case -- a unitary transformation (\ref{uinxbasisApp})
following a projection (\ref{vpierr})),
then sum it over the entire sample as in (\ref{fxerrorW}), (\ref{fxerrorWJointProb}), (\ref{ErrorUnsupervised}),
or (\ref{pql})
below to obtain the number of covered observations.

This approach can be deployed 
to estimate,
as the number of misclassified observations,
other vector--to--vector predictor systems
that
result in the value $\mathbf{f}(\mathbf{x})$,
not in conditional probability $\mathrm{Prob}(\mathbf{f}|\mathbf{x})$:
for example a distribution--to--distribution regression model,
a neural network with vector output, etc.
Take a projection\footnote{
  Note: this is a different concept from a typical consideration of how close are predicted and realized outcomes. 
  For an estimation of this type --- one can test how much the (\ref{GEVfxf}) eigenvalues
  are lower than $1$. The $\mathrm{Error}_{rank}$ from (\ref{ErrorRank}) is an aggregated estimator
  of this type.  }
of the state localized in realized outcome
$\psi_{\mathbf{f}^{(l)}}(\mathbf{f})$
to the state localized 
in predicted outcome $\psi_{\mathbf{f}(\mathbf{x}^{(l)})}(\mathbf{f})$,
obtain
an expression similar to (\ref{projestimation})
weighted over the entire sample:
\begin{align}
  \mathrm{Error}&=\Braket{1}-
  \sum\limits_{l=1}^{M}
  \omega^{(l)}
  \Braket{\psi_{\mathbf{f}^{(l)}}|\psi_{\mathbf{f}(\mathbf{x}^{(l)})}}^2 
  \label{fxerrorWJointProbValuePrediction} \\
\Braket{\psi_{\mathbf{f}}|\psi_{\mathbf{g}}}^2&=
  \frac{
    \left[
    \sum\limits_{j,k=0}^{m-1}
    f_j G^{\mathbf{f};\,-1}_{jk} g_{k}
    \right]^2
    }
       {
         \sum\limits_{j,k=0}^{m-1} f_{j}G^{\mathbf{f};\,-1}_{jk}f_{k}
         \sum\limits_{j,k=0}^{m-1} g_{j}G^{\mathbf{f};\,-1}_{jk}g_{k}
       }
       \label{projarbitrarypredictor}
\end{align}
This error estimator is outlier--stable,
it has the meaning of the number of misclassified observations.
In can be applied to any predictor of $\mathbf{f}(\mathbf{x})$
output type;
when least squares prediction
$\mathbf{f}_{\mathrm{LS}}(\mathbf{x}^{(l)})$
is put to (\ref{fxerrorWJointProbValuePrediction})
obtain (\ref{fxerrorWJointProb}).
These
are not bounded by
(\ref{fxerrorW})
as they are not of (\ref{fxerrorWexpanded}) form.

Another interesting option to consider is to put
$\mathbf{f}\equiv\mathbf{x}$,
then spectral decomposition (\ref{ErrorKfsumSpectral}) corresponds 
to ``coverage expansion'' (\ref{coverageexpansion}) above
and to LRR solution (\ref{UnsupervisedLearningErrorMin}) below with $D=n$.
Let us demonstrate an application of this technique
to the Low-Rank Representation problem.

\subsection{\label{ClusteringErrorF}A Christoffel Function Solution to
  Low-Rank Representation}
For an unlabeled data (no class label $f$ available)
consider the problem of 
clustering
to build a
\href{https://en.wikipedia.org/wiki/Low-rank_approximation#Basic_low-rank_approximation_problem}{Low-Rank Representation} (LRR).
Consider a data (\ref{mlproblem}) without $f$:
\begin{align}
  (x_0,x_1,\dots,x_k,\dots,x_{n-1})^{(l)}& & \text{weight $\omega^{(l)}$}  \label{mlproblemunsup}
\end{align}
the problem is to cluster vector space $\mathbf{x}$ of a dimension $n$
on a subspace of $D<n$ dimension.
A solution\cite{liu2012robust} is to introduce
a $n\times M$ matrix $x_k^{(l)}$ of the rank $n$ (we assume the problem is already regularized), and to represent it by $n\times M$ matrix ${\mathcal X}_k^{(l)}$ of lower rank $D<n$ and an ``error'' matrix $E_k^{(l)}$:
\begin{align}
  x_k^{(l)} &= {\mathcal X}_k^{(l)} +E_k^{(l)} \label{matrrepXE}
\end{align}
The problem is then to find a low-rank representation ${\mathcal X}_k^{(l)}$
from the given 
observation matrix $x_k^{(l)}$,
that allows to recover the given matrix
with a small enough error $E_k^{(l)}$.
The \cite{liu2012robust} authors consider
the following minimization problem:
\begin{align}
  \min\limits_{{\mathcal X},E}& \: \left[\mathrm{rank}({\mathcal X})+\widetilde{\lambda} \|E\|_F\right]
  \label{normmin}
\end{align}
where $\widetilde{\lambda}>0$ is a parameter and
$\|E\|_F$ is a norm, such as the squared \href{https://en.wikipedia.org/wiki/Matrix_norm#Frobenius_norm}{Frobenius norm}.
The main issue with (\ref{normmin}) minimization,
besides 
computational difficulities,
is that the solution is
not gauge invariant relatively (\ref{gaugeX}).

The (\ref{Error}) type of error estimator allows us to construct
a gauge invariant solution. Consider (\ref{psiYlocalized}) state
$\psi_{\mathbf{y}}(\mathbf{x})$
localized at $\mathbf{x}=\mathbf{y}$.
As a regular wavefunction,
when expanded in any full basis $\Ket{\psi^{[i]}}$ obtain:
\begin{align}
  1&=\sum\limits_{i=0}^{n-1} \Braket{\psi_{\mathbf{y}}|\psi^{[i]}}^2
  \label{fullbasisexpansion}
\end{align}
When, instead of a full basis $\Ket{\psi^{[i]}}$ of the dimension $n$,
a basis of lower dimension $D<n$ is used, this
can be for example $\psi_G^{[i]}(\mathbf{x})$ of the dimension $D<n$
from (\ref{psiGclustersOpyatAZavor1e6})
or any other lower dimension
basis $\Ket{\phi^{[i]}}$
orthogonal as $\delta_{ij}=\Braket{\phi^{[i]}|\phi^{[j]}}$,
the sum of squared projections
can be lower than $1$:
\begin{align}
  1&\ge\sum\limits_{i=0}^{D-1} \Braket{\psi_{\mathbf{y}}|\phi^{[i]}}^2
  \label{fullbasisexpansionpartial}
\end{align}
The (\ref{fullbasisexpansionpartial}) was obtained 
back in \cite{malyshkin2015norm} as Eq. (20) therein, where we
summed it over the entire sample.
Similarly, let us sum (\ref{fullbasisexpansionpartial})
with the weights $\omega^{(l)}$ over all $\mathbf{y}\in \mathbf{x}^{(l)}$, $l=1\dots M$ observations.
If all (\ref{fullbasisexpansionpartial}) terms are equal to $1$
then
the total measure $\Braket{1}$ is obtained.
Otherwise the difference
 is an estimation: how well the space $\Ket{\phi^{[i]}}$
of the dimension $D<n$ allows to recover the full space $x_k$ of the
dimension $n$. The error is:
\begin{align}
  \mathrm{Error}&= \Braket{1} - \sum\limits_{l=1}^{M}\omega^{(l)}
  \sum\limits_{i=0}^{D-1} \Braket{\psi_{\mathbf{x}^{(l)}}|\phi^{[i]}}^2
  \label{ErrorUnsupervised} \\
{\mathcal X}_k^{(l)} &=\sum\limits_{i=0}^{D-1}
        \Braket{x_k|\phi^{[i]}}\phi^{[i]}(\mathbf{x}^{(l)})
        \label{unsupclaster}
\end{align}
Unsupervised clustering solution
 is a $D$--dimensional
$\phi^{[i]}(\mathbf{x})$ basis minimizing the (\ref{ErrorUnsupervised}) error.
The solution to (\ref{ErrorUnsupervised}) minimization problem
can be readily obtained from $\Braket{\psi_{\mathbf{y}}|\phi}^2=K(\mathbf{y})\phi^2(\mathbf{y})$
and $\Ket{\psi_K^{[i]}}$ definition in (\ref{psiCK}):
\begin{align}
  \mathrm{Error}&= \Braket{1} - \sum\limits_{i=0}^{D-1} \lambda_K^{[i]}
  \label{UnsupervisedLearningErrorMin}
\end{align}
This is (\ref{ErrorUnsupervised}) written in a subset of $\Ket{\psi_K^{[i]}}$ basis. For $D=n$ this is previously obtained
coverage expansion (\ref{coverageexpansion}).
The Christoffel function clustering solution $\Ket{\phi^{[i]}}$
is then: the $D\le n$ vectors $\Ket{\psi_K^{[i]}}$ out of $n$
corresponding to $D$ largest $\lambda_K^{[i]}$. It can be converted
to $\mathbf{x}$ basis as (\ref{unsupclaster}).
The (\ref{unsupclaster}) is a low-rank representation of the data:
the matrix ${\mathcal X}_k^{(l)}$ of rank $D$
represents the original data matrix $x_k^{(l)}$ of rank $n$.
In contradistinction to (\ref{normmin}) solution,
the solution (\ref{UnsupervisedLearningErrorMin})
is gauge invariant relatively (\ref{gaugeX})
and unique if there is no $\lambda_K^{[i]}$ degeneracy.
This property enables a new range of availabilities
that are 
not practical (or even not possible) for other clustering
methods. The two most remarkable features ---
a possibility to use the ``product attributes'' (\ref{xmulti})
and the fact that
the ``coverage expansion''
solution (\ref{UnsupervisedLearningErrorMin})
is obtained from the expansion (\ref{rhoChristoffel}) of the Christoffel function,
that is small for a seldom observed $\mathbf{x}$.
This is important when input data (\ref{mlproblemunsup})
is a union of subspaces.
If $\mathbf{x}\in S_1$ and $\mathbf{y}\in S_2$
the union $S_1\cup S_2$ does not form a vector space ($a\mathbf{x}+b\mathbf{y}\in S_1\cup S_2$ iff $S_1 \subseteq S_2$ or $S_2 \subseteq S_1$).
The Christoffel function is small for the vectors not
in $S_1\cup S_2$, thus it serves as an indicator function of
a vector from subspaces direct
sum $S_1 \oplus S_2$ to belong to subspaces union $S_1\cup S_2$.

The option \texttt{\seqsplit{--flag\_replace\_f\_by\_christoffel\_function=true}} of Appendix \ref{RN} software makes the program to construct and output the
$\psi_K^{[i]}(\mathbf{x}^{(l)})$ matrix from read $x_i^{(l)}$ input
matrix of the dimensions: $i=0\dots n-1$; $l=1\dots M$.
Set option \texttt{\seqsplit{--flag\_print\_verbosity=3}}
to print all $\Braket{x_k|\psi_K^{[i]}}$ coefficients and $\psi_K^{[i]}(\mathbf{x}^{(l)})$ values
to obtain ${\mathcal X}_k^{(l)}$.
The error (\ref{UnsupervisedLearningErrorMin}) depends on how many 
$\Ket{\psi_K^{[i]}}$ are included in (\ref{unsupclaster}) as $\Ket{\phi^{[i]}}$,
the error is zero if all $\Ket{\psi_K^{[i]}}$ are included.

\subsection{\label{LRRInDynamics}An application of LRR representation solution to
 dynamic system identification problem.}
For an application  of LRR solution
to a dynamic system identification
 consider a linear stochastic dynamic system:
\begin{align}   
  \frac{x_j^{(l+1)}-x_j^{(l)}}{\tau}&\approx
  \frac{ dx_j^{(l)}}{dt}=
  \sum\limits_{k=0}^{n-1}M_{jk}x_k^{(l)} +\epsilon_j^{(l)}
  \label{mdyneq}
\end{align}
Here we assume that the dataset (\ref{mlproblemunsup})
is $l$--ordered (e.g. $l$ is time and all $\omega^{(l)}=1$).
The (\ref{mdyneq}) left--hand side is a discrete analogue of time--derivative, the $\bm{\epsilon}^{(l)}$
is a noise with some distribution (not necessary Gaussian). The problem: to determine the matrix $M_{jk}$ for a given observation set $x_k^{(l)}$, $k=0\dots n-1; l=1\dots M$.

This problem has a trivial ``projection'' solution,
similar to (\ref{fjregression}) projection with a replace $f_k \to dx_k/dt$:
\begin{align}
  M_{jk}&=\sum\limits_{i=0}^{n-1}\Braket{\frac{dx_j}{dt}|x_i}G^{-1}_{ik}
  \label{mijtrivial}
\end{align}
corresponding to a direct projection of $dx_j/dt$ vectors to $\mathbf{x}$-space;
it has zero error when $\bm{\epsilon}^{(l)}=0$.
This solution is formally applicable even when
$\mathbf{x}$ and $d\mathbf{x}/dt$ spaces are of different dimension,
e.g. $dx_j/dt$,  $j=0\dots n-1$, are original attributes derivatives,
and  $x_{\mathbf{k}}$ are product attributes (\ref{xmulti})
with a multi--index $\mathbf{k}$; there are
${\mathcal N}(n,{\mathcal D})$ product attributes (\ref{nond}).
Then the matrix $M_{jk}$ is of the dimension
$n\times {\mathcal N}(n,{\mathcal D})$ and the matrix
$G^{-1}_{ik}$ is of the dimension
${\mathcal N}(n,{\mathcal D}) \times {\mathcal N}(n,{\mathcal D})$
The selection of a space to project is the key element of
any approach, a direct use of the full $\mathbf{x}$-space
(even more so for product attributes space)
typically produces poor results.

The $\mathbf{x}$ is a
\href{https://en.wikipedia.org/wiki/Phase_space}{phase space}
of the dynamic system (\ref{mdyneq}),
for a mechanical system it is coordinates and momentums $\mathbf{x}=(q,p)$.
Dynamic system equation determines the evolution of
a point in the phase space.
The biggest practical problem with a dynamic system identification is
that the phase space can be of a very large dimension. We need a
low--dimensional subset
that captures most of the dynamic features.

In case of a stationary dynamic system (\ref{mdyneq}) our solution
is straightforward: apply Section \ref{ClusteringErrorF} LRR
solution to the phase space matrix $x_k^{(l)}$, $k=0\dots n-1; l=1\dots M$:
Construct the $K(\mathbf{x})$,
perform (\ref{GEVK}) coverage expansion in $\mathbf{x}$--space,
then select $D\le n$
maximal eigenvalues
(according to (\ref{UnsupervisedLearningErrorMin}) error condition),
new basis functions $\phi^{[i]}$, $i=0\dots D-1$
are corresponding to them eigenvectors (\ref{psiCK}).
Then study the system dynamics in $\phi^{[i]}$ basis of dimension $D\le n$:
\begin{align}   
  \frac{d \phi^{[i]}}{dt}&=
  \sum\limits_{k=0}^{D-1}\widetilde{M}_{ik}\phi^{[k]} +\epsilon_i
  & i,k=0\dots D-1
  \label{mdyneqPhi} \\
  \phi^{[i]}&=\sum\limits_{j=0}^{n-1}\alpha_j^{[i]}x_j \label{phidef}
\end{align}
Instead of the original problem
to identify the matrix $M$ of the dimension $n$
the problem became to identify
the matrix $\widetilde{M}$ of the dimension $D\le n$.

The (\ref{mdyneqPhi}) is a ``projected'' dynamic equation.
One can use (\ref{unsupclaster}) to obtain the dynamics in original
variables $x_j$ and $dx_j/dt$.
The LRR solution of Section \ref{ClusteringErrorF}
constructs the $\Ket{\phi^{[i]}}$ basis of the dimension $D$,
this basis is the optimal one to recover the
dynamics of (\ref{mdyneq}) in the form (\ref{mdyneqPhi})
among all $D$-dimensional bases.

\subsection{\label{DynamicEqPsi}Localized states
  $\Ket{\psi_{\mathbf{y}}}$ dynamics.}
A dynamic equation of (\ref{mdyneq}) form
is written in $\mathbf{x}$-space directly.
It is equivalent to a recurrent relation:
\begin{align}
  x_j^{(l+1)}&=\sum\limits_{k=0}^{n-1} {\mathcal M}_{jk} x_k^{(l)} +\epsilon^{(l)}_j
  \label{xjlxkl1diff}
\end{align}
with ${\mathcal M}_{jk}=\delta_{jk}+\tau M_{jk}$ being evolution matrix
and a renormalized noise.
This equation determines the dynamics of a point
in the original
\href{https://en.wikipedia.org/wiki/Phase_space}{phase space}
$\mathbf{x}$
of the system.
Existing dynamics techniques typically use a variant of
\href{https://en.wikipedia.org/wiki/Kalman_filter}{Kalman filter}\cite{kalman1960new}
approach, which is a linear quadratic estimation (LQE).
The central concept of these approaches is
the covariance matrix, a ``glorified standard deviation''
concept.
The technique developed in this paper is based on
using a wavefunction $\psi(\mathbf{x})=\sum_{k=0}^{n-1}\alpha_k x_k$
and obtaining the results by averaging with the $\psi^2(\mathbf{x})$ weight.
For this reason,
instead of considering the dynamic of a point itself,
we
are going to consider the dynamics
of a
wavefunction
localized at some point of the phase space:
not the dynamics of $\mathbf{x}^{(l)}$
but of a state $\psi_{\mathbf{x}^{(l)}}(\mathbf{x})$, localized at
$\mathbf{x}=\mathbf{x}^{(l)}$;
it is the state $\psi_{\mathbf{y}}(\mathbf{x})$ from (\ref{psiYlocalized})
with $\mathbf{y}=\mathbf{x}^{(l)}$.

The transition $\mathbf{x}^{(l)} \rightarrow \mathbf{x}^{(l+1)}$
corresponds to localized wavefunction
transition $\Ket{\psi_{\mathbf{x}^{(l)}}} \rightarrow \Ket{\psi_{\mathbf{x}^{(l+1)}}}$:
\begin{align}
  \psi_{\mathbf{x}^{(l+1)}}(\mathbf{x})&=
  {\mathcal U} \psi_{\mathbf{x}^{(l)}}(\mathbf{x}) + \epsilon
  \label{dyneqpsi} \\
  \Ket{\psi_{\mathbf{x}^{(l+1)}}}&=
  \Ket{{\mathcal U}|\psi_{\mathbf{x}^{(l)}}} + \Ket{\epsilon}
  \nonumber
\end{align}
Here the $\|{\mathcal U}\|$ is a \textbf{unitary} operator (to preserve normalizing)
converting
$\psi_{\mathbf{y}}(\mathbf{x})$ from (\ref{psiYlocalized})
from $\mathbf{y}=\mathbf{x}^{(l)}$ to $\mathbf{y}=\mathbf{x}^{(l+1)}$;
in the simplest stationary case it can be considered $l$--independent,
and $\Ket{\epsilon}$ is a noise vector.
The (\ref{dyneqpsi}) is written in two types of notation;
it can be projected to any orthogonal basis
$\psi^{[i]}$ (for example (\ref{psiC}) with any $f$, Christoffel basis (\ref{psiCK}), regularized basis $X_i$ from the Appendix \ref{regularization}, etc.)
to be written in the matrix form:
\begin{align}
  s_i^{(l)}&=\Braket{\psi_{\mathbf{x}^{(l)}}|\psi^{[i]}}
  =\frac{\psi^{[i]}(\mathbf{x}^{(l)})}
  {\sqrt{\sum\limits_{j=0}^{n-1}
    \left|\psi^{[j]}(\mathbf{x}^{(l)})\right|^2
  }}
  &1=\sum\limits_{i=0}^{n-1}\left|s_i^{(l)}\right|^2
  \label{projSil} \\
  s_j^{(l+1)}&=\sum\limits_{k=0}^{n-1} {\mathcal U}_{jk} s_k^{(l)} +\epsilon
  \label{dyneqMatrixForm}
\end{align}  
The (\ref{dyneqMatrixForm}) is the dynamic equation
for the projections $\Braket{\psi_{\mathbf{x}^{(l)}}|\psi^{[i]}}$.

The dynamic system identification problem,
for a given observation set $x_k^{(l)}$, $k=0\dots n-1; l=1\dots M$,
instead of
determining evolution matrix ${\mathcal M}_{jk}$ of the dimension $n\times n$
that transforms $\mathbf{x}^{(l)}$ to $\mathbf{x}^{(l+1)}$
now became:
to determine a unitary operator ${\mathcal U}_{jk}$ of the dimension $n\times n$
that transforms
$\psi_{\mathbf{x}^{(l)}}$ to $\psi_{\mathbf{x}^{(l+1)}}$.
If one apply (\ref{mijtrivial})
solution to (\ref{dyneqMatrixForm}) this will be incorrect\footnote{
  It is also incorrect
  to consider  time evolution  operator as an ``average''
  of observed state transitions:
  $\|\widetilde{{\mathcal U}}\|=\sum_{l=1}^{M}\Ket{\psi_{\mathbf{x}^{(l+1)}}}\Bra{\psi_{\mathbf{x}^{(l)}}}$
  with subsequent ``unitarization'' procedure
  (e.g. SVD followed by setting $\Sigma_{jk}=\delta_{jk}$
  we deployed in Eq. (\ref{svdadjusted})
  for numerical optimization)
  because 
  identical dynamics must be obtained under transform  $\psi_{\mathbf{x}^{(l)}}\to\exp(i\varphi^{(l)})\psi_{\mathbf{x}^{(l)}}$ with arbitrary phases $\varphi^{(l)}$, $l=1\dots M$;
  this invariance is satisfied only in (\ref{Unormax}).
  }:
because the (\ref{mijtrivial}) is a equation for a point in phase space.
It corresponds to minimizing
predicted/observed differences which is
the $L^2$ norm error applied to (\ref{xjlxkl1diff}):
\begin{align}
  \sum\limits_{l=1}^{M}
\omega^{(l)}
  \left[x_j^{(l+1)}-\sum\limits_{k=0}^{n-1} {\mathcal M}_{jk} x_k^{(l)}
       \right]^2 &
     \xrightarrow[{{\mathcal M}_{jk}}]{\quad }\min
     & j=0\dots n-1
     \label{l2normmin}
\end{align}
This result in linear system solution with
$\sum\limits_{l=1}^{M} x_j^{(l+1)} x_k^{(l)}\omega^{(l)}$
determining linear system right part
and Gram matrix (\ref{xx}) determining linear systems matrix.

The (\ref{dyneqMatrixForm})
is a equation for wavefunction, e.g. if one apply a $l$-dependent
transform $s^{(l)}_i \to\exp(i\varphi^{(l)}) s^{(l)}_i$, $i=0\dots n-1$,
the result should be identical;
similarly ${\mathcal U}_{jk}$ and $-{\mathcal U}_{jk}$
should provide identical dynamics  (compare with ${\mathcal M}_{jk}\to - {\mathcal M}_{jk}$).
Were we study a quantum system time evolution operator
can be readily obtained as
\href{https://en.wikipedia.org/wiki/Hamiltonian_(quantum_mechanics)}{Hamiltonian} related:
\begin{align}
  {\mathcal U}&=
  \exp \left[-i\frac{t}{\hbar} H \right] \label{Uquantum} \\
   \Ket{\psi^{(t)}}&=\Ket{{\mathcal U}|\psi^{(t=0)}} \label{unitaryPsiEvolution}
\end{align}
Now, however, we are trying to construct the operator ${\mathcal U}$
from the data. The functional\footnote{
  In (\ref{Unormax}) the
  $|\cdot|$ denote absolute value, not an operator. Here 
  $\Big|\Braket{\psi_{\mathbf{x}^{(l+1)}}|{\mathcal U}|\psi_{\mathbf{x}^{(l)}}}\Big|^2=
  \Braket{\psi_{\mathbf{x}^{(l+1)}}|{\mathcal U}|\psi_{\mathbf{x}^{(l)}}}
  \Braket{\psi_{\mathbf{x}^{(l+1)}}|{\mathcal U}|\psi_{\mathbf{x}^{(l)}}}^*$
  is $[0\dots 1]$ bounded value having the meaning of conditional probability
  and determining how well
  the $\psi_{\mathbf{x}^{(l+1)}}$ is recovered from $\psi_{\mathbf{x}^{(l)}}$
  using (\ref{dyneqpsi}).
 }
\begin{align}
  \sum\limits_{l=1}^{M}
  \omega^{(l)}
\Big|\Braket{\psi_{\mathbf{x}^{(l+1)}}|{\mathcal U}|\psi_{\mathbf{x}^{(l)}}}\Big|^2  &
 \xrightarrow[{{\mathcal U}}]{\quad }\max
     \label{Unormax}
\end{align}
determines how well $\psi_{\mathbf{x}^{(l+1)}}$
is reconstructed  from $\psi_{\mathbf{x}^{(l)}}$ by a unitary operator ${\mathcal U}$
when system dynamics takes the form of a sequence of unitary transformations (\ref{dyneqpsi}) of a wavefunction.
It can be interpreted as a density matrix dynamics: consider
localized pure state density matrix $\|\rho_{\mathbf{x}}\|=\Ket{\psi_{\mathbf{x}}}\Bra{\psi_{\mathbf{x}}}$. Then  $\|\widetilde{\rho}_{\mathbf{x}^{(l+1)}}\|=\|{\mathcal U}|\rho_{\mathbf{x}^{(l)}}|{\mathcal U}^{\dagger}\|$ and the criterion (\ref{Unormax}) determines
the difference between realized $\|\rho_{\mathbf{x}^{(l+1)}}\|$
and predicted 
$\|\widetilde{\rho}_{\mathbf{x}^{(l+1)}}\|$
density matrices:
$\sum_{l=1}^{M}\omega^{(l)}\mathrm{Spur}\|\rho_{\mathbf{x}^{(l+1)}}|{\mathcal U}|\rho_{\mathbf{x}^{(l)}}|{\mathcal U}^{\dagger}\|$.
If there is a perfect recovery $\|\rho\|=\|\widetilde{\rho}\|$ for all $l$ -- then, as for pure states $\mathrm{Spur}\|\rho^2\|=1$,
total coverage $\Braket{1}$ is obtained, the difference is an error.
The problem is: to find a unitary transformation ${\mathcal U}$
maximizing (\ref{Unormax}).
In (\ref{projSil}) basis the  (\ref{Unormax}) is:
\begin{align}
  &S_{jk;j^{\prime}k^{\prime}}=
  \sum\limits_{l=1}^{M} \omega^{(l)} s^{(l+1)}_js^{(l)}_k s^{(l+1)\,*}_{j^{\prime}}s^{(l)\, *}_{k^{\prime}}
    \label{Sdef}\\
    &\sum\limits_{j,k,j^{\prime},k^{\prime}=0}^{n-1}
    {\mathcal U}_{jk}S_{jk;j^{\prime}k^{\prime}}{\mathcal U}^*_{j^{\prime}k^{\prime}}
    \xrightarrow[{\mathcal U}]{\quad }\max  \label{optimmatrix}\\
    &\sum\limits_{k^{\prime}=0}^{n-1}{\mathcal U}_{jk^{\prime}} {\mathcal U}^*_{kk^{\prime}}=\delta_{jk}
    \label{optimmatrixConstraint}\\
      &S_{jk;j^{\prime}k^{\prime}}=S^*_{j^{\prime}k^{\prime};jk}
  \label{SSymmetric}
\end{align}
The optimization problem (\ref{optimmatrix}) is
considered for a matrix ${\mathcal U}_{jk}$ satisfying
unitarity constraint (\ref{optimmatrixConstraint});
the $S_{jk;j^{\prime}k^{\prime}}$ is a Hermitian tensor (\ref{SSymmetric})
obtained from the data sample,
in an orthogonal basis it takes the form (\ref{Sdef});
for $S_{jk;j^{\prime}k^{\prime}}=\delta_{jj^{\prime}}\delta_{kk^{\prime}}$ Eq. (\ref{optimmatrix})
becomes (\ref{optimmatrixConstraintScalar}).
A complex unitary matrix ${\mathcal U}_{jk}$ of dimension $n$
is determined by
$n^2$ real parameters (a complex Hermitian matrix
of full rank is determined by $n^2$
real parameters,
a unitary matrix is obtained from it's complex exponent,
similar to (\ref{Uquantum})).
Were the constraint (\ref{optimmatrixConstraint})
be of scalar type
$\sum_{j,k,k^{\prime}=0}^{n-1}{\mathcal U}_{jk^{\prime}} {\mathcal U}^*_{kk^{\prime}}=n$
or, even better, the squared \href{https://en.wikipedia.org/wiki/Matrix_norm#Frobenius_norm}{Frobenius norm} of ${\mathcal U}$:
\begin{align}
    &\sum\limits_{j,k=0}^{n-1}{\mathcal U}_{jk} {\mathcal U}^*_{jk}=n
    \label{optimmatrixConstraintScalar}
\end{align}
which is the sum of all (\ref{optimmatrixConstraint}) diagonal components,
then
Eq. (\ref{optimmatrix}) can be considered as a quadratic
form with a vector of $n^2$ dimension
obtained from matrix elements of operator ${\mathcal U}_{jk}$
row by row; the (\ref{optimmatrixConstraintScalar})
is a regular Euclidean scalar product for this vector,
the \href{https://en.wikipedia.org/wiki/Frobenius_inner_product}{Frobenius inner product}.
Remarkably, that (\ref{optimmatrix}) solution  with the constraint (\ref{optimmatrixConstraintScalar})
instead of (\ref{optimmatrixConstraint})
can be obtained as  a regular eigenproblem solution, however it
does not produce the matrix ${\mathcal U}_{jk}$ that is exactly unitary,
nevertheless it may be a good starting point for a numerical method.

For exact unitary constraint optimization problem
(\ref{optimmatrix}) can be approached using Lagrange multipliers
technique where it takes the form (\ref{variatelagrangetovariate}), similar to an eigenvalue problem:
\begin{align}
  S {\mathcal U} &= \lambda {\mathcal U}
  \label{eigenvaluesLikeProblem}
\end{align}
but $S$ is now a \textbf{Hermitian tensor}, 
``eigenvector'' ${\mathcal U}$ is a \textbf{unitary matrix},
and ``eigenvalues''  $\lambda$ is a \textbf{Hermitian matrix}
(\ref{newLambdaSol});
functional (\ref{Unormax}) extremal value is equal to  $\lambda$ spur.

While a complete mathematical structure of this problem
requires a separate study, it's portion
required for 
a dynamic system identification:
find a unitary matrix ${\mathcal U}_{jk}$
maximizing
(\ref{optimmatrix}), can be readily 
solved numerically, see Appendix \ref{NumericalSilutionUnitaryMatrix}
below.

When performing
realtime analysis of (\ref{mlproblemunsup}) data
at any given moment $l$ only the data of $1\dots l$ interval
is available, not $1\dots M$ as required in (\ref{xx}) and (\ref{Sdef})
for calculation of $G_{jk}$ and $S_{jk;j^{\prime}k^{\prime}}$.
In this case the $G_{jk}$ and $S_{jk;j^{\prime}k^{\prime}}$ should be calculated
on  $1\dots l$ sample, thus all the calculations
start having ``sliding'' $G_{jk}$ and $S_{jk;j^{\prime}k^{\prime}}$,
e.g. every new observation coming add one more $\omega x_jx_k$ term
to $G_{jk}$; a weight such as $\omega(t)=\exp\left(-(t_{now}-t)/\tau\right)$
allows recurrently adjust the sum without re-calculating aggregates
of previously
observed  sample.
An example of sliding
$G_{jk}$ technique can be found in \cite{MalMuseScalp}.
Moreover, in this case a ``secondary'' Hilbert space
can be  constructed from some
\textsl{calculated} at $t=l$ value (such as the maximal eigenvalue of operator
$I=dV/dt$, the number
of shares traded per unit time; a highly singular function \cite{malyshkin2018spikes})
treating it as it were plain \textsl{observed} at $t=l$ with the weight $\omega^{(l)}$.
For marker dynamics this allows to separate price changes
that occurred on rising and falling execution flow $I=dV/dt$.
As only the former ones have predictive power, this allows us to construct
a ``scalp'' price: the sum of price changes occurred on rising execution rate.

In this section a new approach to dynamic system identification
is developed. Instead of considering a trajectory in
phase space we convert a sequence of phase space observations
$\mathbf{x}^{(l)}$ to a sequence of probability states $\psi_{\mathbf{x}^{(l)}}(\mathbf{x})$ (wavefunctions)
localized at $\mathbf{x}^{(l)}$.
Then system dynamics is considered
as a sequence of unitary transformations of the wavefunction.
The approach allows to write the dynamics of these probability states;
quality criterion (\ref{optimmatrix}) estimates the number of
correctly predicted outcomes.
The probability of the next outcome $\mathbf{x}^{(l+1)}$ being equal $\mathbf{y}$ given currently observed outcome equal $\mathbf{x}^{(l)}$ is:
\begin{align}
  P(\mathbf{x}^{(l+1)}
  =\mathbf{y})\Big|_{\mathbf{x}^{(l)}}&=
  \Big|\Braket{\psi_{\mathbf{y}}|{\mathcal U}|\psi_{\mathbf{x}^{(l)}}}\Big|^2
  \label{Poutcome}
\end{align}
The approach can be readily generalized to density matrix states,
however a unitary form (\ref{unitaryPsiEvolution}) of the dynamics
has limitations in data analysis (e.g. in application to 
the data of
\href{https://en.wikipedia.org/wiki/Markov_chain}{Markov chain}
type), this requires to approach the problem of state decoherence,
see Applendix \ref{sysdecoherence} below.
In this section we solved the problem of determining
evolution operator  ${\mathcal U}_{jk}$ from a ``sequence of wavefunctions''
$\psi_{\mathbf{x}^{(l)}}(\mathbf{x})$ that are
obtained from a sequence of observation points in phase space
$\mathbf{x}^{(l)}$.
The key element for this success is
the (\ref{Unormax}) form of quality criteria.
This criterion satisfies wavefunction unobservability,
a fundamental characteristic
of a quantum system: whereas Schr\"{o}dinger equations
is written for a wavefunction, the
wavefunction itself is not observable, only it's absolute square
can be measured.  The (\ref{Unormax})
is invariant if all $l=1\dots M$ observations
has the wavefunction defined within an arbitrary phase shifts:
$\psi_{\mathbf{x}^{(l)}}\to\exp(i\varphi^{(l)})\psi_{\mathbf{x}^{(l)}}$;
similarly two time--evolution operators $\|{\mathcal U}\|$
produce identical dynamics if they transform a wavefunction
within a phase shift.
One may ask a question: given a sequence
of quantum mechanical wavefunctions,
can this approach identify a quantum system?
The answer is definitely yes 
if only time--evolution operator
(\ref{Uquantum}) is required
(Appendix \ref{NumericalSilutionUnitaryMatrix} optimization problem).
If the Hamiltonian,
not just time evolution operator,
is required then the formal answer is yes, but practically this
requires taking a logarithm of a unitary matrix,
what is a complex problem required a separate consideration\cite{loring2014computing}.

Another important topic to discuss is allowed
transformation of a $\Ket{\psi}$ state.
Whereas for quantum systems only unitary transformation (\ref{unitaryPsiEvolution}) determined by a unitary matrix ${\mathcal U}_{jk}$ is allowed,
in data analysis it can possibly be of a non--unitary form.
We see ``non--unitary dynamics'' as an important
direction of further research, see Appendix
\ref{nonunitarydynamics}
discussing unitary transformations following by a projection
and Appendix
\ref{sysdecoherence}
discussing \href{https://en.wikipedia.org/wiki/Quantum_channel#Pure_channel}{quantum channel} type of transformation (\ref{KrausOperator}).

\section{\label{conclusion}Conclusion}

In this work the support weight of Radon--Nikodym form $\psi^{2}(\mathbf{x})$,
with $\psi(\mathbf{x})$ function to be  a linear
function on $x_k$ components
was considered and applied to interpolation, classification,
and optimal clustering problems.
The most remarkable feature of the Radon--Nikodym approach is that
input attributes $x_k$ are used not for constructing the $f$,
but for constructing
a probability density (support weight) $\psi^2(\mathbf{x})$,
which is then used for
evaluation of the value $f=\Braket{f(\mathbf{x})\psi^2}/\Braket{\psi^2}$
or conditional probability.
This way we can avoid using a norm in $f$--space, what
greatly increases
practical applicability of the approach.

A distinguishing feature
of the developed approach is knowledge of the predictor's
invariant group. Given (\ref{mlproblem}) dataset, what $\mathbf{x}$
basis transform does not change the solution?
Typically in ML (neural networks, decision tree, SVM, etc.)
the invariance is either completely unknown or poorly understood.
The invariance is known for linear regression (and a few other linear models),
but linear regression has an unsatisfactory knowledge representation.
Developed in this paper Radon--Nikodym approach
has 1) known invariant group (non--degenerated linear transform of $\mathbf{x}$ components)
and
2) advanced knowledge representation in the form of matrix spectrum;
even an answer of the \hyperref[firstOrderLogic]{first order logic} type
becomes feasible.
The knowledge is extracted by applying projection
operators, thus completely avoiding using a norm
in the solution to
interpolation (\ref{RNfsolutionpsi}),
classification (\ref{RNWfsolutionpsi}),
and optimal clustering (\ref{psiGclustersOpyatAZavor1e6})
problems.

The developed approach, while being mostly completed
for the case of a scalar class label $f$,
has a number of unsolved problems
in case of a vector class label $\mathbf{f}$.
As the most intriguing one
we see the question:
whether the optimal clustering solution of Section
\ref{BasisReduction}
can be generalized to vector--valued class label
approach of Section \ref{vectorvalued}: the solutions
(\ref{RNfsolutionVector}) and  (\ref{RNWfsolutionVector})
have no basis dimension reduction feature,
and the conditional probability solution (\ref{projestimation}) currently
always sets clusters number to be equal to the dimension of vector class label.
For our first try
to construct a subspace with an arbitrary number of $D\le n$ clusters 
see optimization problem (\ref{uconstrNUcontributingSubspace})
 below.

\appendix
\section{\label{regularization}Regularization Example}
An input vector $\mathbf{x}= (x_0,x_1,\dots,x_k,\dots,x_{n-1})^{(l)}$
from (\ref{mlproblem})
may have redundant data, often highly redundant.
An example of a 
redundant data is the situation
when two attribute components are equal e.g. $x_k=x_{k+1}$ for all $l$.
In this case the $G_{jk}=\Braket{x_j|x_k}$ matrix
becomes degenerated and the generalized eigenvalue problem (\ref{GEV})
cannot be solved directly, thus a regularization is required.
A regularization process consists
in selection of such $x_k$ linear combinations that remove the redundancy,
mathematically the problem is equivalent to
\href{https://en.wikipedia.org/wiki/Rank_(linear_algebra)}{finding the rank}
of a symmetric matrix.

All the theory of this paper is invariant with respect to any non--degenerated
linear transform of $\mathbf{x}$ components.
For this reason we may consider the vector $\widetilde{\mathbf{x}}$
with equal to zero average, as this transform improves the numerical stability of $\Braket{x_j|x_k}$ calculation.
Obtain $\Braket{\widetilde{x}_j|\widetilde{x}_k}$ matrix (it is plain \href{https://en.wikipedia.org/wiki/Covariance_matrix}{covariance matrix}):
\begin{align}
\widetilde{\mathbf{x}}&=(x_0-\overline{x}_0,x_1-\overline{x}_1,\dots,x_k-\overline{x}_k,\dots,x_{n-1}-\overline{x}_{n-1})\label{xaverageeq0}\\
\overline{x}_k&=\frac{\Braket{x_k}}{\Braket{1}} \label{xkaverage} \\
\widetilde{G}_{jk}&=\Braket{\widetilde{x}_j|\widetilde{x}_k} \label{wtildeXX} \\
\sigma_k&=\sqrt{\frac{\widetilde{G}_{kk}}{\Braket{1}}}\label{stdev{xk}}
\end{align}
For each $k=0\dots n-1$ consider standard deviation $\sigma_k$ of $x_k$, select the
set $S$ of indexes $k$, that have standard deviation greater that a given $\varepsilon$, determined by
computer's numerical precision. Then construct the matrix $\widetilde{G}_{jk}$ with the indexes \textsl{in the set obtained}: $j,k\in S$.
The new matrix $\widetilde{G}_{jk}$
is obtained by removing $x_k$ components that are equal to a constant,
but it still can be degenerated.

We need to regularize the problem by removing the redundancy.
The  criteria is like a \href{https://en.wikipedia.org/wiki/Condition_number#Matrices}{condition number}
in a linear system problem, but because we deploy generalized eigenproblem
anyway, we can do it straightforward.
Consider generalized eigenproblem (\ref{GEVcond}) with the right hand side matrix equals
to diagonal components of $\widetilde{G}_{jk}$.
\begin{align}
  j,k\in& S \label{jkinS}\\
  \widetilde{G}^d_{jk}&=\delta_{jk}\widetilde{G}_{kk} \label{diagGG}\\
   \sum\limits_{k\in S} \widetilde{G}_{jk} \alpha^{[i]}_k &=
  \lambda^{[i]} \sum\limits_{k\in S} \widetilde{G}^d_{jk} \alpha^{[i]}_k
  \label{GEVcond} \\
  S^{d}:& \text{a set of $i$, such that:}\lambda^{[i]}>\varepsilon \label{setselSd} \\
  X_{S^{d}}&=\sum\limits_{k\in S}\alpha^{[S^{d}]}_k \left(x_k-\overline{x}_k\right)
  \label{Xsd}
\end{align}
By construction of the $S$ set the right hand side diagonal matrix
$\widetilde{G}^d_{jk}$ has only positive terms,
that are not small, hence the (\ref{GEVcond}) has a unique solution.
The eigenvalues $\lambda^{[i]}$ of the problem (\ref{GEVcond})
have a meaning of a ``normalized standard deviation''. Select (\ref{setselSd}) set:
the indexes $i$,
such that the $\lambda^{[i]}$ is greater than a given
$\varepsilon$, determined by
computer's numerical precision. Obtained $S^{d}$ set determines
regularized
basis (\ref{Xsd}). The matrix $\Braket{X_i|X_m}$ with $i,m\in S^{d}$ is non--degenerated.
After the constant component $X=1$ is added to the basis (\ref{Xsd})
the $\mathbf{X}=(\dots X_i \dots, 1 )$
can be used in (\ref{mlproblem})
instead of the $\mathbf{x}=(\dots x_k \dots )$.
This algorithm is implemented in \texttt{\seqsplit{com/polytechnik/utils/DataReadObservationVectorXF.java:getDataRegularized\_EV()}}.

Alternatively to (\ref{setselSd}), a regularization
can be performed without solving the eigenproblem (\ref{GEVcond}),
using an approach similar to
\href{https://en.wikipedia.org/wiki/Pivot_element#Partial_and_complete_pivoting}{Gaussian elimination with pivoting}
in a linear system problem.
This algorithm is implemented  in \texttt{\seqsplit{com/polytechnik/utils/DataReadObservationVectorXF.java:getDataRegularized\_LIN()}}.
Which regularization method to be used depends on the parameter
\texttt{\seqsplit{--regularization\_method=}} supplied to
\texttt{\seqsplit{com/polytechnik/utils/RN.java}} driver,
see Appendix \ref{RN} below.

A \href{https://en.wikipedia.org/wiki/Singular_value_decomposition}{singular value decomposition}
is often used as a regularization method. However,
for a symmetric matrix considered in this appendix, without
\href{https://en.wikipedia.org/wiki/Moore%E2%80%93Penrose_inverse}{pseudoinverse}
required,
a regularization method based on symmetric eigenproblem (\ref{GEVcond})
provides the same result with lower computational complexity.
Moreover, even a ``Gaussian elimination  with pivoting''
type of regularization
provides the result of about the same quality.

Regardless the regularization details,
for a given input data in the basis $x_k$,
 different regularization methods
produce the same number of  $\mathbf{X}$ components,
formed
\href{https://en.wikipedia.org/wiki/Vector_space}{vector space}
is the same regardless the regularization used;
the dimension of it is the rank of $\Braket{ x_j|x_k}$ matrix.
Important, that because the developed theory
is ``gauge invariant'' relatively (\ref{gaugeXF}),
all inference results are identical
regardless regularization method used,
see \texttt{\seqsplit{com/polytechnik/utils/TestDataReadObservationVectorXF .java:testRegularizations()}} unit test for a demonstration.
It is important to stress that:
\begin{itemize}
\item No any information on $f$ have been used in the regularization of $G_{jk}=\Braket{x_j|x_k}$.
\item All ``standard deviation`` type of thresholds
  were compared with a given $\varepsilon$, determined by
  the computer's numerical precision. No ``standard deviation`` is used
  in solving the inference problem itself.
\end{itemize}
The result of this appendix is a new basis $\mathbf{X}=(\dots X_i \dots, 1)$
of $1+\dim S^{d}$ elements ((\ref{Xsd}) and const,
the rank of $\Braket{ x_j|x_k}$)
that now can be used in (\ref{mlproblem}) instead of original $\mathbf{x}=(\dots x_k \dots)$.
Obtained basis provides a non--degenerated
Gram matrix $\Braket{X_i|X_m}$ (\ref{xx}).

\section{\label{RN}RN Software Usage Description}
The \href{http://www.ioffe.ru/LNEPS/malyshkin/code_polynomials_quadratures.zip}{provided software} is written in java.
The source code files of interest are
\texttt{\seqsplit{com/polytechnik/utils/\{RN,RadonNikodymSpectralModel,DataReadObservationVectorXF,AttributesProductsMultiIndexed\}.java}}.
The class \texttt{\seqsplit{DataReadObservationVectorXF}}
reads input data (\ref{mlproblem}) from a comma--separated file
and stores the observations.
The methods \texttt{\seqsplit{getDataRegularized\_EV()}}
or
 \texttt{\seqsplit{getDataRegularized\_LIN()}}
perform Appendix \ref{regularization} data regularization
and return an object of \texttt{\seqsplit{DataRegularized}} type that contains the matrices $\Braket{X_j|X_k}$ and $\Braket{X_j|f|X_k}$ in the regularized basis $\mathbf{X}$.
The method \texttt{\seqsplit{getRadonNikodymSpectralModel()}} of this object
creates Radon--Nikodym spectral model of Section \ref{RNapproach}, it returns an object of \texttt{\seqsplit{RadonNikodymSpectralModel}} class. The method \texttt{\seqsplit{getRNatXoriginal(double\enspace[]{\enspace}xorig)}} of this object
evaluates an observation at a
$\mathbf{xorig}$
in the original basis (\ref{mlproblem}) and returns an object of
\texttt{\seqsplit{RadonNikodymSpectralModel.RNPointEvaluation}} type;
this object has the methods \texttt{\seqsplit{getRN()}}, \texttt{\seqsplit{getRNW()}}, and \texttt{\seqsplit{getPsikAtX()}}
that, for a $\mathbf{xorig}$ given, calculate the (\ref{RNfsolutionpsi}), (\ref{RNWfsolutionpsi}),
and $\psi^{[i]}(\mathbf{xorig})$ components.
An object of \texttt{\seqsplit{RadonNikodymSpectralModel}} type
has a method \texttt{\seqsplit{reduceBasisSize(int\enspace D)}}
that performs optimal clustering of Section \ref{BasisReduction}
and returns \texttt{\seqsplit{RadonNikodymSpectralModel}} object
with the basis chosen as the optimal dimension \texttt{\seqsplit{D}}
clusterization of $f$. The documentation
produced by
\href{https://docs.oracle.com/en/java/javase/13/javadoc/javadoc.html#GUID-7A344353-3BBF-45C4-8B28-15025DDCC643}{javadoc}
is bundled with the \href{http://www.ioffe.ru/LNEPS/malyshkin/code_polynomials_quadratures.zip}{provided software}.

The 
\texttt{\seqsplit{com/polytechnik/utils/RN.java}} is a driver
to be called from a command line.
The driver's arguments are:
\begin{itemize}
\item \verb+--data_file_to_build_model_from=+ The input file name
  to read (\ref{mlproblem}) data and build a Radon--Nikodym model from it.
  The file is comma--separated,
  if the first line starts with the \texttt{|\#} --- it is considered to be
  the column names, otherwise the
  column names are created from their indexes. Empty lines and the lines starting with the \texttt{|}
  are considered comments. All non--comment lines must have
  identical number of columns. 
\item \verb+--data_file_evaluation=+ The input files (multiple options with multiple files possible)
  to evaluate the model built. The same format.
\item \verb+--data_cols=+ The description of the
  input files data columns. The format is \texttt{\seqsplit{--data\_cols=numcols:xstart,xend:f:w:label}},
  where \texttt{\seqsplit{numcols}} is the total number
  of columns in the input file,  \texttt{\seqsplit{xstart,xend}}
  are the columns to be used for $x_k$, e.g. the columns
  \texttt{\seqsplit{(xstart,xstart+1,\dots,xend-1,xend)}} are used
  as the $(x_0,x_1,\dots,x_k,\dots,x_{n-1})$ in (\ref{mlproblem}) input.
  The \texttt{\seqsplit{f}} and \texttt{\seqsplit{w}} are the columns
  for class label $f$ and weight $\omega$,
  if weight column index $\mathtt{w}$ is set to negative then all weights $\omega$
  are set to $1$.
  The \texttt{\seqsplit{label}} is column index of observation identification string
  (uniquely identifies a data row in the input data file, a typical identification is: row number \texttt{12345}, $x\times y$ image pixel id \texttt{132x15}, customer id \texttt{johnsmith1990}, etc.), it is copied without modification (or set to \texttt{??} if \texttt{\seqsplit{label}} is negative)
  from input data file
  to the first column of output file.
  All column identifiers are
  integers, base 0 column index. 
  For example input file \texttt{\seqsplit{dataexamples/runge\_function.csv}} of Appendix \ref{reproduce}
  has $9$ columns, the $x_k$ are in the first $7$ columns, then $f$ and $\omega$ columns follow, the $x_1$ is
  used as observation string label of input file row. This corresponds to \texttt{\seqsplit{--data\_cols=9:0,6:7:8:1}}
\item \verb+--clusters_number=+  The value of $D$. If presents Section \ref{BasisReduction}
  optimal clustering is performed with this $D$ and the output is of this dimension.
  Otherwise all $n$ input components are used
  to construct the $\psi^{[i]}(\mathbf{x})$ from (\ref{psiC})
  and the dimension of the output is the rank of $\Braket{ x_j|x_k}$ matrix.
\item \verb+--regularization_method=+ Data regularization method to be used,
  possible values: \verb+NONE+, \verb+EV+ (default), and \verb+LIN+, see Appendix \ref{regularization} for algorithms description.
\item \verb+--max_multiindex=+ The value of ${\mathcal D}$. If presents then
  ${\mathcal N}(n,{\mathcal D})$ ``product''
  attributes $X_0^{k_0}X_1^{k_1}\dots X_{n-1}^{k_{n-1}}$
  are constructed (\ref{xmulti}) in regularized basis (using recursive algorithm)
  with the multi--index $\mathbf{k}$ lower or equal than the ${\mathcal D}$,
  these ``product'' attributes 
  are then used instead of $n$ original attributes  $x_k$,
  see Section \ref{firstOrderLogic} above.
  For a large enough ${\mathcal D}$ the problem may become numerically unstable.
  For ${\mathcal N}(n,{\mathcal D})\ge 500$ used eigenvalue routines
  may be very slow\footnote{
  For eigenproblem routines
  one can use \href{https://docs.oracle.com/javase/8/docs/technotes/guides/jni/}{JNI} interface
   \texttt{\seqsplit{com/polytechnik/lapack/Eigenvalues\_JNI\_lapacke.java}}
  to
  \href{http://www.netlib.org/lapack/}{LAPACK}
  instead of java code, see \texttt{\seqsplit{com/polytechnik/utils/EVSolver.java}} for selection.}.  
  The option is intended to be deployed together
  with \texttt{\seqsplit{--clusters\_number=}}
  with the goal
  to obtain a model of a ``first order logic'' type.
\item \verb+--flag_print_verbosity=+ By default is \texttt{\seqsplit{2}}. Set \texttt{\seqsplit{--flag\_print\_verbosity=1}} to suppress the
  output of $\psi^{[i]}(\mathbf{x}^{(l)})$ values or set
  \texttt{\seqsplit{--flag\_print\_verbosity=3}} to output the projections
  $\Braket{x_k|\psi^{[i]}}$ in expansion
  $x_k^{(l)}=\sum_{i=0}^{n-1}\Braket{x_k|\psi^{[i]}}\psi^{[i]}(\mathbf{x}^{(l)})$.
  Usefult for obtaining LRR ${\mathcal X}_k^{(l)}$ matrix (\ref{unsupclaster}) from
  printed $\psi^{[i]}(\mathbf{x}^{(l)})$ values.
\item \verb+--flag_replace_f_by_christoffel_function=+ By default is \texttt{\seqsplit{false}}. If set to \texttt{\seqsplit{true}} then, after regularization
  of the Appendix \ref{regularization},
  the Christoffel function (\ref{ChristoffelLike}) is calculated for every observation
  and used instead of $f$; datafile read values of $f$ are discarded.
  Useful for unsupervised learning.
  While mathematical result does not depend on $f$, the specific basis used
  may affect numerical stability
  because of initial regularization;
  in this situation a good heuristic is to use observation number as the $f$,
  this removes class label degeneracy and makes the basis more stable.
\item \verb+--flag_assume_f_is_diagonal_in_christoffel_function_basis=+ By default is false. If set to true then $f$ is  considered to be diagonal in $\Ket{\psi_K^{[i]}}$ basis (\ref{psiCK}).
  Sampled matrix $\Braket{x_j|f|x_k}$ is converted to $\Braket{\psi_K^{[j]}|f|\psi_K^{[k]}}$, all off--diagonal elements are removed,
  then the matrix diagonal in $\Ket{\psi_K^{[i]}}$ basis is converted back to $x_i$ basis.
  This can be viewed as \cite{gsmalyshkin2017comparative} type
  of transform: $\|f\|\approx \sum\limits_{i=0}^{n-1} \Ket{\psi^{[i]}_K} \Braket{\psi^{[i]}_K|f|\psi^{[i]}_K} \Bra{\psi^{[i]}_K}$. This is an experimental option
  to vector class label classification problem of Section \ref{AttribsVectorF}.
\item \verb+--output_files_prefix=+ If set all output files are prefixed by this string.
  A typical usage is to save output to some directory, such as
  \texttt{\seqsplit{--output\_files\_prefix=/tmp/}}.
\end{itemize}

The program reads the data, builds Radon--Nikodym model
from \texttt{\seqsplit{--data\_file\_to\_build\_model\_from=}}
then evaluates it on itself and on all
\texttt{\seqsplit{--data\_file\_evaluation=}} files.
The output file has the same filename with the
\texttt{\seqsplit{.RN.csv}} extension appended.
In the comments section it prints data statistics
(filename, observations number, and the Lebesgue quadrature (\ref{lebesgueQ})).
Column data description
is presented in the column header.
Every output row corresponds to an input file row.
An output row has a number of columns.
The first column is observation string label,
then $n+2$ columns follow: observation original input attributes $x_k$,
observation class label $f$, and observation weight $\omega$.
Calculated data
is put into additional columns of the same row.
The columns are: \texttt{\seqsplit{f\_RN}} (\ref{RNfsolutionpsi}),
\texttt{\seqsplit{f\_LS}} (\ref{regrfsolution}),
\texttt{\seqsplit{Christoffel}} (\ref{ChristoffelLikepsi}),
\texttt{\seqsplit{f\_RNW}}  (\ref{RNWfsolutionpsi})
\texttt{\seqsplit{Coverage}} (\ref{coverage}),
and, unless \texttt{\seqsplit{--flag\_print\_verbosity=1}},
the $\psi^{[i]}(\mathbf{x}^{(l)})$ (\ref{psiC}) $D$ components.
Here the $D$ is either the rank of $\Braket{ x_j|x_k}$ matrix, or
the parameter \texttt{\seqsplit{--clusters\_number=}} if specified.
For all output files the following relations are held for the columns:
\begin{align}
  \mathtt{f\_RN}^{(l)}&=
  \frac{\sum\limits_{i=0}^{D-1} f^{[i]}\left[\psi^{[i]}(\mathbf{x}^{(l)})\right]^2}
         {\sum\limits_{i=0}^{D-1} \left[\psi^{[i]}(\mathbf{x}^{(l)})\right]^2}
         \label{ProgfRN} \\
\mathtt{Christoffel}^{(l)}&=
\frac{1}{\sum\limits_{i=0}^{D-1}\left[\psi^{[i]}(\mathbf{x}^{(l)})\right]^2}
\label{ProgChristoffel} \\
\mathtt{f\_RNW}^{(l)}&=\frac{\sum\limits_{i=0}^{D-1} f^{[i]}w^{[i]}
           \left[\psi^{[i]}(\mathbf{x}^{(l)})\right]^2}
         {\sum\limits_{i=0}^{D-1} w^{[i]}\left[\psi^{[i]}(\mathbf{x}^{(l)})\right]^2}
         \label{ProgfRNW} \\
\mathtt{Coverage}^{(l)}&=\frac{\sum\limits_{i=0}^{D-1} w^{[i]}\left[\psi^{[i]}(\mathbf{x}^{(l)})\right]^2}
       {\sum\limits_{i=0}^{D-1} \left[\psi^{[i]}(\mathbf{x}^{(l)})\right]^2}
       \label{ProgCoverage}
\end{align}
For the file the model is built from (learning data)
a few additional relations are held ($i,m=0\dots D-1$):
\begin{align}
  w^{[m]}&=\left[\sum\limits_{l=1}^{M}\psi^{[m]}(\mathbf{x}^{(l)})\omega^{(l)}\right]^2 \label{ProgW} \\
    f^{[m]}\delta_{im}&=\sum\limits_{l=1}^{M}\psi^{[i]}(\mathbf{x}^{(l)})\psi^{[m]}(\mathbf{x}^{(l)})f^{(l)}\omega^{(l)} \label{ProgfNorm} \\
  \delta_{im}&=\sum\limits_{l=1}^{M}\psi^{[i]}(\mathbf{x}^{(l)})\psi^{[m]}(\mathbf{x}^{(l)})\omega^{(l)} \label{ProgNorm}
\end{align}
Obtained $D$ states $\psi^{[m]}(\mathbf{x})$
(for $D<\mathrm{rank\,of}\Braket{ x_j|x_k}$
these are the $\psi_G^{[m]}(\mathbf{x})$
from  (\ref{psiGclustersOpyatAZavor1e6}),
$w^{[m]}=w_G^{[m]}$ from (\ref{psiGortWeight}),
and $f^{[m]}=\lambda_G^{[m]}$)
provide the optimal clustering of class label $f$ among all
$D$--point discrete measures.

\subsection{\label{install} Software Installation And Testing}

\begin{itemize}
\item Install \href{https://www.oracle.com/technetwork/java/javase/downloads/jdk11-downloads-5066655.html}{java 11} or later.
\item Download the source code
\href{http://www.ioffe.ru/LNEPS/malyshkin/code_polynomials_quadratures.zip}{\texttt{\seqsplit{code\_polynomials\_quadratures.zip}}}
from \cite{polynomialcode}.
\item Decompress and recompile the program. Run a selftest.
\begin{verbatim}
unzip code_polynomials_quadratures.zip
javac -g com/polytechnik/*/*java
java com/polytechnik/utils/TestDataReadObservationVectorXF
\end{verbatim}

\item Run the program with bundled deterministic
 data file (Runge function (\ref{rungeF})).
\begin{verbatim}
java com/polytechnik/utils/RN --data_cols=9:0,6:7:8:1 \
   --data_file_to_build_model_from=dataexamples/runge_function.csv \
   --data_file_evaluation=dataexamples/runge_function.csv
\end{verbatim}
Here, for usage demonstration, we evaluate the model twice.
The file \texttt{\seqsplit{runge\_function.csv.RN.csv}}
will be created (the same file is written twice,
because the built model is then test--evaluated on the same input \texttt{\seqsplit{dataexamples/runge\_function.csv}}).
See Appendix \ref{reproduce} below for interpolation results obtained from the output.
\item
  Run the program with the constructed $\psi^{[i]}(\mathbf{x}^{(l)})$ (\ref{psiC})
  as input. They are in the columns with the index $15$ to $21$ of
  the file \texttt{\seqsplit{runge\_function.csv.RN.csv}} ($22$ columns total).
\begin{verbatim}
java com/polytechnik/utils/RN --data_cols=22:15,21:8:9:0 \
   --data_file_to_build_model_from=runge_function.csv.RN.csv
\end{verbatim}
The file \texttt{\seqsplit{runge\_function.csv.RN.csv.RN.csv}}
will be created. Because the input $x_k$ are now
selected as $\psi^{[k]}(\mathbf{x})$,
with this input,
the Radon--Nikodym approach of Section \ref{RNapproach}
produce exactly the input $x_k$
as the result $\psi^{[k]}(\mathbf{x})$, possibly with $\pm 1$ factor.
There are $7$ nodes/weights of the Lebesgue quadrature (\ref{lebesgueQ})
for input data file \texttt{\seqsplit{dataexamples/runge\_function.csv}}:
\begin{equation}
  \begin{aligned}
f^{[0]}&= 0.042293402383175485 & w^{[0]}&= 0.2453611587632685 \\
f^{[1]}&= 0.043621284685679745 & w^{[1]}&= 0 \\
f^{[2]}&= 0.06535351052058812 & w^{[2]}&= 0.5222926033815862 \\
f^{[3]}&= 0.07864169617926474 & w^{[3]}&= 0 \\
f^{[4]}&= 0.16469273913045052 & w^{[4]}&= 0.6710343400073819 \\
f^{[5]}&= 0.28493524789476266 & w^{[5]}&= 0 \\
f^{[6]}&= 0.7025238747369117 & w^{[6]}&= 0.5613118978475747
\end{aligned}
\label{quadrNumRunge}
\end{equation}

Some of the Lebesgue weights are $0$. This may happen
with (\ref{wiLeb}) definition. The weights sum is equal to total measure,
for (\ref{totalmeasurerungeF}) it is equal to $2$.
\item The dimension of the Lebesgue quadrature is $n$,
it is the number of input attributes $x_k$. 
When we start to increase the $n$, the Lebesgue quadrature
starts to partition the $\mathbf{x}$ space
on smaller and smaller elements.
The
(\ref{RNfsolutionpsi}) type of answer
will eventually start to exhibit
data overfitting effect. Radon--Nikodym is much less prone to it
than a direct expansion of $f$ in $x_k$, a (\ref{regrf}) type of answers,
but for a large enough $n$ even the $\Braket{f\psi^2}/\Braket{\psi^2}$
type of answer is starting to overfit the data.
We need to select $D\le n$ linear combinations
of $x_k$ that optimally separate the $f$.
Optimal clustering is described in Section \ref{BasisReduction}.
Run the program
\begin{verbatim}
java com/polytechnik/utils/RN --data_cols=9:0,6:7:8:1 \
   --data_file_to_build_model_from=dataexamples/runge_function.csv \
   --clusters_number=4
\end{verbatim}
Running with \texttt{\seqsplit{--clusters\_number}} equals to
$5$, $6$, or $7$ may fail to construct a Gaussian quadrature (\ref{GEVgBracket})
as the number of the measure (\ref{Lmeasure}) support points should be
greater or equal than
the dimension of Gaussian quadrature built on this measure.
For \texttt{\seqsplit{--clusters\_number=4}} the obtained
quadrature gives exactly the (\ref{quadrNumRunge}) nodes with zero
weights removed: the optimal approximation of the measure with
four support points by a four points discrete measure
is the measure itself.
\begin{equation}
  \begin{aligned}
f^{[0]}&= 0.04229340238319568 & w^{[0]}&= 0.24536115876382128  \\
f^{[1]}&= 0.065353510520606 & w^{[1]}&= 0.5222926033810373 \\
f^{[2]}&= 0.1646927391304516 & w^{[2]}&= 0.6710343400073585 \\
f^{[3]}&= 0.7025238747369116 & w^{[3]}&= 0.5613118978475746
  \end{aligned}
  \label{quadrNumRungeN4}
\end{equation}
A more interesting case is to set \texttt{\seqsplit{--clusters\_number=3}}
\begin{verbatim}
java com/polytechnik/utils/RN --data_cols=9:0,6:7:8:1 \
   --data_file_to_build_model_from=dataexamples/runge_function.csv \
   --clusters_number=3
\end{verbatim}
\begin{equation}
 \begin{aligned}
f^{[0]}&= 0.0553329558917533 & w^{[0]}&= 0.737454390130916 \\
f^{[1]}&= 0.16285402990411255 & w^{[1]}&= 0.701183615381193 \\
f^{[2]}&= 0.7025131758981266 & w^{[2]}&= 0.5613619944877021
\end{aligned}
\label{quadrNumRungeN3}
\end{equation}
The (\ref{quadrNumRungeN3}) is the optimal approximation
of the measure (\ref{quadrNumRunge}) with 4 support points
by a 3--point discrete distribution, this is
a typical application of Gaussian quadrature.
The $n$--point Gaussian quadrature requires $0\dots 2n-1$
distribution moments for calculation,
the measure must have at least $n$ support points.
The distribution moments
of $f$ can be obtained using a different method,
for example using the sample sum (\ref{matrixfxx}) directly.
A remarkable feature of the Lebesgue integral measure (\ref{Lmeasure})
is that obtained eigenvectors (\ref{psiCg})
can be converted from $f$  to $\mathbf{x}$ space. The conversion
formula is (\ref{psiGclustersOpyatAZavor1e6}).
The $\psi_G^{[m]}(\mathbf{x})$, $m=0\dots D-1$
create the weights, that optimally separate $f$
in terms of $\Braket{f\psi^2}/\Braket{\psi^2}$ separation.
This is a typical setup of the technique we developed:
\begin{itemize}
\item For a large number $n$ of input attributes
  create the Lebesgue integral quadrature (\ref{lebesgueQ}).
\item Select the number of clusters $D\le n$. Using Lebesgue measure
  (\ref{Lmeasure})
  build Gaussian quadrature (\ref{gaussQ}) in $f$ space. It provides
  the optimal clustering of the dimension $D$.
\item Convert
  obtained results from $f$ to $\mathbf{x}$ space
  using  (\ref{psiGclustersOpyatAZavor1e6}),
  obtain the $\psi_G^{[m]}(\mathbf{x})$ classifiers.
\item One can also entertain a first order logic --like model
  using the attributes of Section \ref{firstOrderLogic}.
\end{itemize}

\begin{figure}[t]
  \includegraphics[width=16cm]{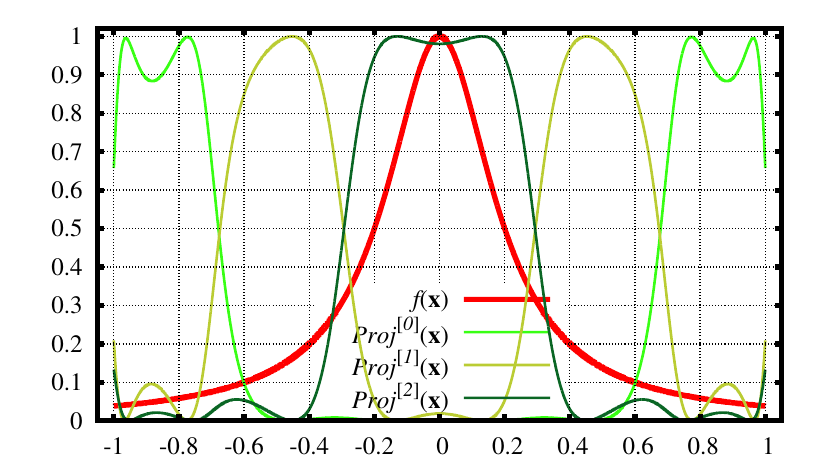}
  \caption{\label{RungeProj3}
 Runge function (\ref{rungeF}) data (\ref{mlproblem1D}) clustered to $D=3$.
 Corresponds to (\ref{quadrNumRungeN3}) data.
 The projections (\ref{distY}) to
 $\psi_G^{[m]}(\mathbf{x})$, $m=0\dots D-1$ are presented.
}
\end{figure}
\item The three function $\psi_G^{[m]}(\mathbf{x})$, corresponding
  to (\ref{quadrNumRungeN3}) nodes, are presented in Fig. \ref{RungeProj3}.
  The $\mathrm{Proj}^{[i]}(\mathbf{x})$ (this is squared and
  normalized $\psi_G^{[m]}(\mathbf{x})$  as (\ref{distY})).
  One can clearly see that the states $\psi_G^{[m]}(\mathbf{x})$
  are localized exactly near the $f^{[m]}$ nodes (\ref{quadrNumRungeN3}).
  This technique is a much more powerful one,
  than, say, \href{https://en.wikipedia.org/wiki/Support-vector_machine#Definition}{support--vector machine} linear separation.
  In the Radon--Nikodym approach
  the separation weights are the $\left[\psi_G^{[m]}(\mathbf{x})\right]^2$
  that are obtained without an introduction of a norm
  with subsequent
  minimization the difference between the result and a prediction
  with respect to the norm.
  The separation by the functions
  $\psi_G^{[m]}(\mathbf{x})$ is optimal
  among all $D$-- dimensional
  separations of $\left[\psi(\mathbf{x})\right]^2$ type.
  The cost is that the solution is now two--step\cite{2015arXiv151107085G}.
  On the first step the Lebesgue quadrature is built and the
  measure (\ref{Lmeasure}) is obtained. On the second step
  the Gaussian quadrature (\ref{gaussQ}) is built on this measure;
  the result is then converted 
  to $\mathbf{x}$ space (\ref{psiGclustersOpyatAZavor1e6}). The
  $\left[\psi_G^{[m]}(\mathbf{x})\right]^2$
  are the optimal separation weights.
\end{itemize}

\subsection{\label{NominalExample}Nominal Attributes Example}
In ML applications the attributes (\ref{mlproblem})
can be nominal. They may be of orderable (low, medium, high)
or unorderable (apple, orange, tomato) type.
A nominal attribute taking two values can be converted to $\{0,1\}$
binary attribute.
Orderable attributes (low, medium, high) can be converted
to $\{1,2,3\}$, or, say, $\{1,2,10\}$ this depends on the problem.
For unorderable attributes the conversion
is more difficult, however in some situations
it is straightforward:  a ``country'' attribute taking the value:
``country name from a list of eight countries'',
can be converted to three binary attributes.

The $f$, predicted by a ML system, is called class label.
It is often a binary attribute. This leads to
the nodes (\ref{fiLeb}) of the Lebesgue quadrature to
be grouped near two values of the class label.
We have tested a number of datasets  from
\href{https://archive.ics.uci.edu/ml/}{UC Irvine Machine Learning Repository},
\href{https://www.cs.waikato.ac.nz/ml/weka/datasets.html}{Weka datasets},
and other sources.
For direct comparison with
the existing software such as
\href{https://www.rulequest.com/see5-unix.html}{C5.0}
or
\href{https://www.cs.waikato.ac.nz/ml/weka/}{Weka 3: Machine Learning Software in Java}
a care should be taken of nominal attributes conversion
and class label representation.
We are going to discuss the details in a separate
publication, here we present only  qualitative
aspects of Radon--Nikodym approach
application
to ML problem with the \textsl{binary} class label.
Take \href{https://archive.ics.uci.edu/ml/machine-learning-databases/breast-cancer-wisconsin/}{breast-cancer-wisconsin}
database, the \texttt{\seqsplit{breast-cancer-wisconsin.data}}
dataset\cite{mangasarian1990cancer}
is of $699$ records,
we removed 16 records with unknown (``?'') attributes and split
the dataset
as 500:183
for training:testing. Obtained files are
\begin{verbatim}
wc breast-cancer-wisconsin_S.names \
   breast-cancer-wisconsin_S.data \
   breast-cancer-wisconsin_S.test 
  139   938  6234 breast-cancer-wisconsin_S.names
  500   500 14266 breast-cancer-wisconsin_S.data
  183   183  5182 breast-cancer-wisconsin_S.test
  822  1621 25682 total
\end{verbatim}
The data has nominal class label
\texttt{\seqsplit{2: Benign,{\enspace}4:Malignant}}.
\href{https://www.rulequest.com/see5-unix.html}{C5.0},
when run on this dataset produces a very good classifier:
\begin{verbatim}
c5.0  -f mldata/breast-cancer-wisconsin_S
Evaluation on training data (500 cases):
           (a)   (b)    <-classified as
          ----  ----
           293    10    (a): class 2
             3   194    (b): class 4
Evaluation on test data (183 cases):
           (a)   (b)    <-classified as
          ----  ----
           139     2    (a): class 2
             4    38    (b): class 4
\end{verbatim}
Now let us run the RN program to obtain the Lebesgue quadrature
\begin{verbatim}
java com/polytechnik/utils/RN --data_cols=11:1,9:10:-1:0 \
   --data_file_to_build_model_from=mldata/breast-cancer-wisconsin_S.data \
   --data_file_evaluation=mldata/breast-cancer-wisconsin_S.test
\end{verbatim}
The number of the nodes is 10, it is equal to
the number of input attributes $x_k$.
\begin{equation}
\begin{aligned}
f^{[0]}&= 2.090917684500027 & w^{[0]}&= 308.30166232236996 \\
f^{[1]}&= 3.198032991602546 & w^{[1]}&= 5.307371268658678 \\
f^{[2]}&= 3.344418191526764 & w^{[2]}&= 0.0189894231470068 \\
f^{[3]}&= 3.5619620739712725 & w^{[3]}&= 0.3341989402039986 \\
f^{[4]}&= 3.6221628167395497 & w^{[4]}&= 0.2549558854552573 \\
f^{[5]}&= 3.7509806530824346 & w^{[5]}&= 1.2339290581894928 \\
f^{[6]}&= 3.7939096228600513 & w^{[6]}&= 5.146789024450902 \\
f^{[7]}&= 3.8081118648848045 & w^{[7]}&= 0.16082536035874645 \\
f^{[8]}&= 3.8799894340830727 & w^{[8]}&= 50.25004460556501 \\
f^{[9]}&= 3.9574710127612613 & w^{[9]}&= 128.99123411160124
  \end{aligned}
  \label{quadrbreastcancer10}
\end{equation}
Then we calculate a joint distribution of realization/prediction
for $f_{RN}$ and $f_{RNW}$. The continuous to nominal
conversion for $f_{RN}$ and $f_{RNW}$ was performed by comparing
predicted value
with the average.
Evaluation without clustering on training data (\ref{BreastAllTraining}) (500 cases),
and on test data (\ref{BreastAllTesting}) (183 cases) is:
\begin{align}
\mathrm{Distribution}(f_{RN}) &:
\begin{matrix}
183 &    120 \\
0 &      197
\end{matrix}
&
\mathrm{Distribution}(f_{RNW}) &:
\begin{matrix}
294   &  9 \\
13    &  184
\end{matrix}
\label{BreastAllTraining} \\
\mathrm{Distribution}(f_{RN}) &:
\begin{matrix}
91  &    50 \\
0   &    42
\end{matrix}
&
\mathrm{Distribution}(f_{RNW}) &:
\begin{matrix}
140 &    1 \\
0   &    42
\end{matrix}
\label{BreastAllTesting}
\end{align}
We see that $f_{RN}$ that equally treats
the states with low and high prior probability often gives spurious
misclassifications. In the same time the $f_{RNW}$
that uses the projections adjusted to prior probability
gives a superior prediction. 

When we cluster to $D=2$:
\begin{verbatim}
java com/polytechnik/utils/RN --data_cols=11:1,9:10:-1:0 \
   --data_file_to_build_model_from=mldata/breast-cancer-wisconsin_S.data \
   --data_file_evaluation=mldata/breast-cancer-wisconsin_S.test \
   --clusters_number=2
\end{verbatim}
\begin{equation}
  \begin{aligned}
f^{[0]}&= 2.09463398432689 & w^{[0]}&=310.52326905818705 \\
f^{[1]}&=3.924320437715293 & w^{[1]}&=189.47673094181317
  \end{aligned}
  \label{quadrbreastcancer2}
\end{equation}
The evaluation with $D=2$ clustering
on training data (\ref{BreastD2Training}) (500 cases)
and on test data (\ref{BreastD2Testing}) (183 cases)
gives
joint distribution of realization/prediction
for $f_{RN}$ and $f_{RNW}$:
\begin{align}
\mathrm{Distribution}(f_{RN}) &:
\begin{matrix}
292  &   11 \\
7     &  190
\end{matrix}
&
\mathrm{Distribution}(f_{RNW}) &:
\begin{matrix}
295   &  8 \\
13    &  184
\end{matrix}
\label{BreastD2Training} \\
\mathrm{Distribution}(f_{RN}) &:
\begin{matrix}
141  &   0 \\
0    &   42
\end{matrix}
&
\mathrm{Distribution}(f_{RNW}) &:
\begin{matrix}
 141   &  0\\
1    &   41
\end{matrix}
\label{BreastD2Testing}
\end{align}
Now, after the states with low prior probabilities (\ref{Pprior})
are removed, both $f_{RN}$ and $f_{RNW}$ exhibit
a good classification. For $D=3$, however, we still get
a type of 
(\ref{BreastAllTraining}) and (\ref{BreastAllTesting})
behavior
of spurious misclassifications by $f_{RN}$
and no such behavior in  $f_{RNW}$.

This makes us to conclude that the $f_{RNW}$ answer
is the superior answer for
predicting a \textsl{probabilistic} $f$.
The posterior
distribution (\ref{Pposterior}) is Radon--Nikodym alternative to Bayes.

\section{\label{reproduce}\hyperref[RN]{RN Program} Application
With A Different Definition Of The Probability}
Besides a typical ML classification problem the \hyperref[RN]{RN Program}
can be used for a number of different tasks, e.g. it can be applied to
an interpolation problem.
The reason is simple: as an input Radon--Nikodym
only needs (\ref{matrixfxx}) matrices $F_{jk}$ and $G_{jk}$,
which are calculated from (\ref{mlproblem}) sample,
that is a file of $M$ rows and $n+2$ columns ($n$ for $x_k$ and two for $f$ and the weight $\omega$).
In the Appendix \ref{RN} the probabilities (\ref{matrixfxx}) 
were obtained as an
\href{https://en.wikipedia.org/wiki/Ensemble_average_(statistical_mechanics)}{ensemble average}, calculated from the data,
this is typical for a ML classification problem.

\begin{figure}[t]
  \includegraphics[width=16cm]{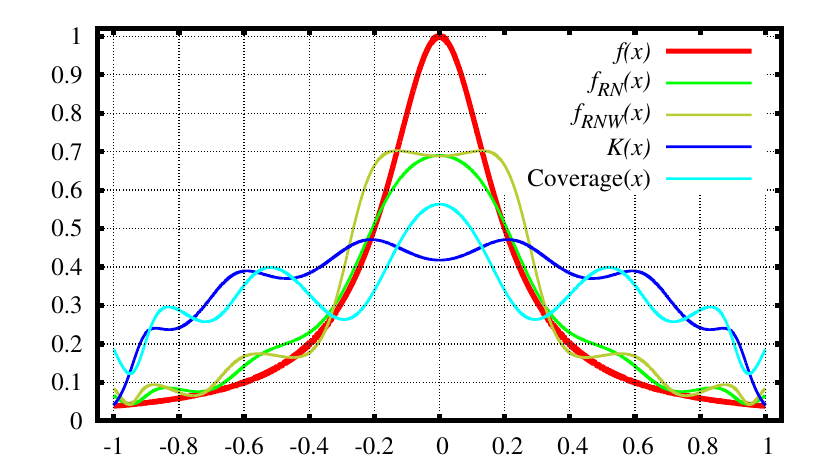}
  \caption{\label{RNRUNGE}
Runge function (\ref{rungeF}) interpolation result for $n=7$.
The input data (\ref{mlproblem})
was prepared (\ref{mlproblem1D})
in a way the classification problem solver from Appendix \ref{RN} 
to reproduce interpolation results of the 
Appendix D of
\cite{2015arXiv151005510G}.
The $f_{RNW}(\mathbf{x})$ (\ref{RNWfsolutionpsi}) (olive),
Christoffel function (blue) (\ref{ChristoffelLikepsi}),
and the $\mathrm{Coverage}(\mathbf{x})$ (sky) (\ref{coverage})
for the measure $\Braket{g}=\int_{-1}^{1}g(x)dx$ (\ref{totalmeasurerungeF})
are also calculated.
}
\end{figure}

Input file can be constructed in a way
that calculated averages represent a probability
of different kind, such as
\href{https://en.wikipedia.org/wiki/Ergodicity}{time average} probability.
Consider function interpolation problem,
the $\Braket{\cdot}$ now has a meaning of time--average
$\Braket{g}=\int g(x)\omega(x)dx$,
see Section II of \cite{2015arXiv151005510G}.
A one--dimensional interpolation problem\cite{2016arXiv161107386V}
can be reduced to (\ref{mlproblem}) data by converting
a two--columns sequence
$x^{(l)}\to f^{(l)}$, $l=1\dots M$ to:
\begin{align}
  (1,x,x^2,\dots,x^{n-1})^{(l)}&\to f^{(l)}& \text{weight $\omega^{(l)}$}
  \label{mlproblem1D}
\end{align}
Because the result is invariant relatively
any non--degenerated basis components linear transform
any polynomials (e.g. $P_m(x)$, $T_m(x)$, etc.) can be used instead of the $x^m$ in (\ref{mlproblem1D}).
For example: to reproduce
\href{https://en.wikipedia.org/wiki/Runge%27s_phenomenon}{Runge function}
  $d=1$ interpolation problem
\begin{align}
  f(x)&=\frac{1}{1+25x^2} \label{rungeF} \\
  d\mu&=dx \label{totalmeasurerungeF} \\
  x&\in [-1:1] \nonumber
\end{align}
for $n=7$, the result of the Appendix D of
\cite{2015arXiv151005510G},
take $x$ sequence with a small step about $dx=10^{-4}$, it will be about $M=1+2/dx$ total points
$x\in[-1,-1+dx,-1+2dx,\dots,1-2dx,1-dx,1]$
and create a comma--separated file of $M$ rows and $n+2$ columns:
$1,x,x^2,\dots,x^{n-1},f(x),\omega$. First $n$ columns are the $\mathbf{x}$ from (\ref{mlproblem1D}),
then $f(x)$ from (\ref{rungeF}) follows, and the last column is the
observation weight
$\omega=dx$ for all points except the $dx/2$ for the edges.
This file \texttt{\seqsplit{dataexamples/runge\_function.csv}}
is bundled with
\href{http://www.ioffe.ru/LNEPS/malyshkin/code_polynomials_quadratures.zip}{provided software}.
Run the program
\begin{verbatim}
java com/polytechnik/utils/RN --data_cols=9:0,6:7:8:1 \
      --data_file_to_build_model_from=dataexamples/runge_function.csv
\end{verbatim}
The output file \texttt{\seqsplit{runge\_function.csv.RN.csv}} has a few more columns,
four of them are: the $f_{RN}$ from (\ref{RNfsolutionpsi}),
the Christoffel function (\ref{ChristoffelLikepsi}),
the $f_{RNW}$ from (\ref{RNWfsolutionpsi}),
and the
$\mathrm{Coverage}(\mathbf{x})$ (\ref{coverage}).
The result is presented in Fig. \ref{RNRUNGE}. With
the data prepared as (\ref{mlproblem1D})
the Christoffel--like function (\ref{ChristoffelLikepsi})
is the regular Christoffel function for the measure (\ref{totalmeasurerungeF}).
The $f_{RNW}(x)$
is also presented in Fig. \ref{RNRUNGE}.
The $f_{RNW}(x)$, same as the $f_{RN}(x)$,  is a weighted
superposition (\ref{RNWfsolutionpsi}) of (\ref{GEVBracket}) eigenvalues,
but the weights are the \textsl{posterior weights} (\ref{Pposterior}),
that are the product of prior weights by the $\Ket{\psi^{[i]}}$ projections: $w^{[i]}\mathrm{Proj}^{[i]}$.
For Runge function in $n=7$ case only four prior weights (\ref{quadrNumRunge}) are non--zero,
thus in Fig. \ref{RNRUNGE} the $f_{RNW}(x)$ is a superposition of four eigenvalues.
As we discussed above in Section \ref{NashOtvetBayesy},
the $f_{RN}(x)$ should be used for a deterministic functions,
and the $f_{RNW}(x)$ is a solution to classification problem
for a probabilistic $f$;
it uses 
the posterior weights (\ref{Pposterior}).
Same result can be also obtained using multi--index multiplications
of Section \ref{firstOrderLogic}, take a single $x$ attribute and multiply
it by itself 6 times. The quadrature will be identical.
\begin{verbatim}
java com/polytechnik/utils/RN --data_cols=9:0,1:7:8:1 \
      --max_multiindex=6 \
      --data_file_to_build_model_from=dataexamples/runge_function.csv
\end{verbatim}

Radon--Nikodym interpolation \cite{2015arXiv151101887G} of an image
  ($d=2$ problem),
can be performed in a similar way.
Create a file of $M=d_x\times d_y$ rows and $n=n_x\times n_y +2$ columns.
Each row corresponds to a single pixel.
The last two columns are: pixel gray intensity and the weight (equals to 1). The first $n=n_x\times n_y$
columns are a function of pixel coordinate $(x_l\in 0\dots d_x-1,y_l\in 0\dots d_y-1)$ as $T_{j_x}(2\frac{x_l}{d_x-1}-1)T_{j_y}(2\frac{y_l}{d_y-1}-1)$, $j_x=0\dots n_x-1$,
$j_y=0\dots n_y-1$. The $T_m(x)$ is Chebyshev polynomial $T_0=1;T_1=x;\dots$,
they are chosen for numerical stability.
In \cite{2015arXiv151101887G}
the multi--index $\mathbf{j}=(j_x,j_y)$ has (\ref{lenMIconstraintX}) and (\ref{lenMIconstraintY}) constraints.
After running the \hyperref[RN]{RN Program} interpolated $f_{RN}$
and Christoffel function
columns
are added to output file,
the $f_{RN}(x_l,y_l)$ provides required interpolation.
While the Gaussian quadrature cannot be
obtained for $d\ge 2$,
the Christoffel function (\ref{ChristoffelLike})
can be easily calculated
not only in $d\ge 2$ case,
but also for an arbitrary $\mathbf{x}$ space
with a measure $\Braket{\cdot}$.

The input file can be also constructed
for $\mathbf{x}$ vector to represent a random variable.
For example a distribution regression
problem
where a ``bag'' of observations  is mapped to a single outcome $f$
can be approached\cite{2015arXiv151109058G} by using
the moments of the distribution of
a single ``observations bag'' as an input $\mathbf{x}$.
For every ``bag'', calculate it's distribution moments  (one can use any choice of polynomials),
then put these moments as
$\mathbf{x}$
(now the $x_k$ components are the moments of 
the distribution of a bag's instance), and use the $f$ as the outcome.

Similarly, temporal dependencies can be converted to (\ref{mlproblem}) type of data.
Assume $f$ has a $f(\mathbf{x}(t))$ form. Then each $x_k(t)$
can be converted to the moments $\Braket{Q_s(x_k)}_t$, $s=0\dots n_t$,
relatively some time--averaging $\Braket{\cdot}_t$ measure,
such as in the Section II of \cite{2015arXiv151005510G}.
Then the $n\times n_t$ input attributes
$\Braket{Q_s(x_k)}_t$, $k=0\dots n=1;s=0\dots n_t-1$,
are ``mixed'' moments:
time averaged $\Braket{\cdot}_t$ first
and then ensemble averaged in (\ref{matrixfxx}).
They can be used in (\ref{mlproblem}) data input.
Note, that ``combined'' averaging in (\ref{matrixfxx})
as $\Braket{\Braket{Q_s(x_j(t))|Q_{s^{\prime}}(x_k(t))}_t}$
produces different result than
``mixed'' one:
$\Braket{\Braket{Q_s(x_j(t))}_t|\Braket{Q_{s^{\prime}}(x_k(t))}_t}$.
Numerical experiments show that $\Braket{Q_s(x_k)}_t$
attributes typically show a better result
than using $(x_k(t),x_k(t-\delta),x_k(t-2\delta),\dots)$
as a ``vectorish'' $x_k$. With temporal (and spatial)
attributes the dimension of (\ref{mlproblem}) input can grow
very fast. In such a situation Section \ref{BasisReduction}
optimal clustering is of critical importance:
this way we can select only a few combinations of input attributes,
that optimally separate the $f$.

The strength of the Radon--Nikodym approach
is that it requires only two matrices (\ref{matrixfxx}) as an input,
and the average $\Braket{\cdot}$,
used to calculate the $F_{jk}$ and $G_{jk}$,
can be chosen with a different definition of the probability.
The input file (\texttt{\seqsplit{--data\_file\_to\_build\_model\_from=}} parameter) can be prepared in a form to represent any probability space
in any basis of any dimension. One row corresponds to
a single realization, all rows correspond to the entire sample.
After input datafile is prepared for the chosen probability space ---
the features introduced in this paper
$f_{RN}(\mathbf{x})$, $K(\mathbf{x})$,
$f_{RNW}(\mathbf{x})$, $\mathrm{Coverage}(\mathbf{x})$,
along with $\psi_G^{[m]}(\mathbf{x})$ clusters (\ref{psiGclustersOpyatAZavor1e6})
are calculated by the
\href{http://www.ioffe.ru/LNEPS/malyshkin/code_polynomials_quadratures.zip}{provided software}.

\section{\label{NumericalSilutionUnitaryMatrix}A Numerical Solution to
  Quadratic Form Maximization Problem in Unitary Matrix Space}
Consider a constrained optimization problem (\ref{optimmatrix})
\begin{align}
    {\mathcal F}=\sum\limits_{j,k,j^{\prime},k^{\prime}=0}^{n-1}
    {\mathcal U}_{jk}S_{jk;j^{\prime}k^{\prime}}{\mathcal U}^*_{j^{\prime}k^{\prime}}
    &\xrightarrow[{\mathcal U}]{\quad }\max  \label{optimmatrixAppendix}\\
\sum\limits_{k^{\prime}=0}^{n-1}{\mathcal U}_{jk^{\prime}} {\mathcal U}^*_{kk^{\prime}}&=\delta_{jk}
    \label{optimmatrixConstraintAppendix}
\end{align}
This is a problem of optimization
of scalar function (quadratic form with a Hermitian tensor $S_{jk;j^{\prime}k^{\prime}}$ from (\ref{SSymmetric}))  on the
\href{https://en.wikipedia.org/wiki/Unitary_group}{unitary group $U(n)$}.
It is equivalent to a problem of maximizing a quadratic form
with a Hermitian matrix
given multiple constraints (\ref{optimmatrixConstraintAppendix})
of quadratic form as well.
The constraint may be of more general 
``partial unitarity $D\le n$'' form
(\ref{constraintNU}); a slight
algorithm modification is then required,
see Appendix \ref{NumericalSilutionUnitaryMatrixNUDlen} below.
A regular eigenvalue problem has
a single quadratic form constraint, the problem in question has multiple.
We have already approached a problem with an extra
quadratic form
constraint in the Appendix F of \cite{MalMuseScalp},
the problem in question is of this type. Consider a ``simplified constraint'' (\ref{optimmatrixConstraintScalar})
\begin{align}
    &\sum\limits_{j,k=0}^{n-1}{\mathcal U}_{jk} {\mathcal U}^*_{jk}=n
    \label{optimmatrixConstraintScalarAppendix}
\end{align}
as a ``partial'' constraint for which optimization problem (\ref{optimmatrixAppendix})
can be readily converted to an eigenvalue problem to be directly solved.
The idea is then to adjust obtained solution
to satisfy full unitary constraints and
calculate new values for Lagrange multipliers.
Performing several iterations the process will converge
to (\ref{optimmatrixAppendix}) optimization problem solution
with the required constraints (\ref{optimmatrixConstraintAppendix}).

Consider Lagrange multipliers $\lambda_{jk}$
to optimize (\ref{optimmatrixAppendix})
with the constraints (\ref{optimmatrixConstraintAppendix})
\begin{align}
  &
  \sum\limits_{j,k,j^{\prime},k^{\prime}=0}^{n-1}
             {\mathcal U}_{jk}S_{jk;j^{\prime}k^{\prime}}{\mathcal U}^*_{j^{\prime}k^{\prime}}
             +
   \sum\limits_{j,k=0}^{n-1}        
   \lambda_{jk}\left[\delta_{jk}-\sum\limits_{k^{\prime}=0}^{n-1}{\mathcal U}_{jk^{\prime}} {\mathcal U}^*_{kk^{\prime}} \right]
   \xrightarrow[{\mathcal U}]{\quad }\max
   \label{lagrangetovariate}
\end{align}
and variate it over all ${\mathcal U}_{jk}$ components.
There are $2n^2$ real number coefficients defining ${\mathcal U}_{jk}=a_{jk}+i b_{jk}$,
only $n^2$ of them are independent for a unitary matrix.
One more coefficient is dropped as a common phase, so (\ref{optimmatrixAppendix}) optimization
with the constraints (\ref{optimmatrixConstraintAppendix})
is equivalent to an unconstrained optimization problem over $n^2-1$ independent real parameters.

It is typically more convenient  to variate (\ref{lagrangetovariate})
over
${\mathcal U}_{jk}$ and ${\mathcal U}^*_{jk}$
rather than over
 $a_{jk}$ and $b_{jk}$,
then take care of the constraints by adjusting
Lagrange multipliers $\lambda_{jk}$. The variations
\begin{subequations}
  \label{variatelagrangetovariate}
\begin{align}
 0&= \sum\limits_{j^{\prime},k^{\prime}=0}^{n-1}{\mathcal U}_{j^{\prime}k^{\prime}}S_{j^{\prime}k^{\prime};pq}
 -\sum\limits_{j^{\prime}=0}^{n-1}\lambda_{j^{\prime}p}{\mathcal U}_{j^{\prime}q} \label{variationUstar}\\
 0&= \sum\limits_{j^{\prime},k^{\prime}=0}^{n-1}S_{pq;j^{\prime}k^{\prime}}{\mathcal U}^*_{j^{\prime}k^{\prime}}
 -\sum\limits_{j^{\prime}=0}^{n-1}\lambda_{pj^{\prime}}{\mathcal U}^*_{j^{\prime}q}
\end{align}
\end{subequations}
are consistent only when $\lambda_{jk}$
is a Hermitian matrix
\begin{align}
  \lambda_{jk}&=\lambda^*_{kj}
  \label{lamdaHermitian}
\end{align}
From (\ref{variatelagrangetovariate}) also immediately follows:
the functional (\ref{optimmatrixAppendix}) extremal value is equal
to the spur of $\lambda_{jk}$:
\begin{align}
  {\mathcal F}^{(extr)}&=\sum\limits_{j=0}^{n-1}\lambda_{jj}
  \label{Fextremal}
\end{align}

An algorithm finding extremal (\ref{optimmatrixAppendix})
is a generalization of the one from
the Appendix F of \cite{MalMuseScalp} to multiple constraints:
\begin{enumerate}
\item
  Take initial $\lambda_{jk}$ and solve (\ref{lagrangetovariate}) optimization
  with partial constraint (\ref{optimmatrixConstraintScalarAppendix}).
  Solution method -- an eigenvalue problem of $n^2$ dimension
  in a vector space
  formed by writing all ${\mathcal U}_{jk}$ matrix elements
  in a vector,
   row by row.
   The result is: ${\mathcal F}$ and  ${\mathcal U}_{jk}$ matrix reconstructed
   back from the eigenvector corresponding to maximal eigenvalue, row by row.
\item
  Obtained from this solution matrix ${\mathcal U}_{jk}$  may not be unitary
  as the constraint (\ref{optimmatrixConstraintScalarAppendix}) is
  a subset of the full one (\ref{optimmatrixConstraintAppendix}). Expand ${\mathcal U}_{jk}$ in
  \href{https://en.wikipedia.org/wiki/Singular_value_decomposition}{SVD}
  \begin{align}
    {\mathcal U}_{jk}&=\sum\limits_{j^{\prime},k^{\prime}=0}^{n-1}
    U_{jj^{\prime}}\Sigma_{j^{\prime}k^{\prime}}V^{\dagger}_{k^{\prime}k}
    \label{SVD} \\
    \widetilde{{\mathcal U}}_{jk}&=
    \sum\limits_{j^{\prime}=0}^{n-1}
    U_{jj^{\prime}}V^{\dagger}_{j^{\prime}k}
    \label{svdadjusted}
  \end{align}
  and adjust all SVD numbers to $1$: $\Sigma_{jk}=\delta_{jk}$,
  obtained $\widetilde{{\mathcal U}}_{jk}$ is a unitary matrix,
  it is the next iteration of the solution. This matrix (\ref{svdadjusted})
  satisfies
  exact constraint (\ref{optimmatrixConstraintAppendix}),
  but the value of ${\mathcal F}$ is now increased.
  The $\widetilde{{\mathcal U}}_{jk}$ becomes a new ${\mathcal U}_{jk}$
  at this iteration.
\item Put this new ${\mathcal U}_{jk}$ to (\ref{variationUstar}),
  then multiply it by ${\mathcal U}^*_{jq}$ and sum over $q=0\dots n-1$.
  As the ${\mathcal U}_{jk}$ is unitary  
  $\lambda_{jk}=\sum_{p,q=0}^{n-1}\lambda_{jp}{\mathcal U}_{pq} {\mathcal U}^*_{kq}$
  obtain new values for Lagrange multipliers $\widetilde{\lambda}_{jk}$
  and take it's Hermitian part:
  \begin{align}
    \widetilde{\lambda}_{jk}&=
    \sum\limits_{j^{\prime},k^{\prime},q=0}^{n-1}{\mathcal U}_{j^{\prime}k^{\prime}}S_{j^{\prime}k^{\prime};kq}{\mathcal U}^*_{jq}
    \label{newLambdaSol} \\    
    \lambda_{jk}&=\frac{1}{2}
    \left[\widetilde{\lambda}_{jk}+\widetilde{\lambda}^*_{kj}\right]
    \label{newLambda}    
  \end{align}
  This  $\lambda_{jk}$ is 
  the next iteration of Lagrange multipliers.
  As iterations proceed -- the  $\widetilde{\lambda}_{jk}$
  should converge to a Hermitian matrix by itself,
  without (\ref{newLambda}) required.
  \item
    Put this new $\lambda_{jk}$ to (\ref{lagrangetovariate}) and repeat iterational process until converged.
    On the first iteration take initial values for Lagrange multipliers as $\lambda_{jk}=0$.
\end{enumerate}

\section{\label{nonunitarydynamics} Non--Unitary Dynamics}
In the previous section
an approach to numerical solution of
optimization problem (\ref{optimmatrixAppendix})
with unitary constraint (\ref{optimmatrixConstraintAppendix})
has been developed.
Whereas for quantum systems time evolution operator
${\mathcal U}_{jk}$ can be only unitary,
in data analysis it can possibly be of a non--unitary form.
The difference arises because in data analysis
wavefunction is directly ``observable'' (within a phase) with
the goal to construct a ``time evolution operator''
 (\ref{dyneqpsi}).

The first non--unitary matrix of this type to consider is
the (\ref{optimmatrixConstraintScalarAppendix}), having
a single constraint:
the sum of squared elements is equal to $n$.
With this matrix the problem can be easily solved.
It does not preserve the normalizing, but gives more weight
to correctly matched predictions. Regardless interpretation
difficulties the dynamics with
a matrix constrained the sum of squared elements being equal $n$
is the first one to try for the reasons of
computational simplicity (no iterational process required)
and mathematical interpretation simplicity (eigenvalue problem equivalence).

Another matrix of interest is a subspace-projection matrix.
This type of constraint
typically makes 
Lagrange multipliers $\lambda_{jk}$ calculation problematic,
however some results can be obtained analytically,
what makes a subspace-projection matrix the first
one to try for an analytic study.

In the considered above approach to dynamics
the $\mathbf{x}^{(l)}$ and $\mathbf{x}^{(l+1)}$
were belong to the same phase space.
It is of great interest to consider a situation where
$\Ket{\psi_{\mathbf{x}^{(l)}}}$ and
$\Ket{\psi_{\mathbf{x}^{(l+1)}}}$
belong to different vector spaces, e.g.
to use 
$\Ket{\psi_{\mathbf{f}^{(l)}}}$ instead of $\Ket{\psi_{\mathbf{x}^{(l+1)}}}$.
In this case in (\ref{dyneqpsi}) operator ${\mathcal U}$
is transforming $\Ket{\psi_{\mathbf{x}}}$ to
a different vector space
$\Ket{\psi_{\mathbf{f}}}$; this is not a true ``dynamics'' ($l$ is the same),
but such a transform can be applied to a
traditional ML
classification problem.

While a study of a general non--unitary  $\mathbf{x}\to\mathbf{f}$
homomorphism producing the most general form of
non--unitary dynamics is out of scope of this
work (see Appendix \ref{sysdecoherence} below for our first attempt),
let us consider a simple composition of a unitary transformation ${\mathcal U}$:
$\mathbf{x}\to\mathbf{x}$ followed by projection
of $\mathbf{x}$ on $\mathbf{f}$,
a ``projective''
form of non--unitary dynamics\footnote{
  Similar composition of a unitary transformation
$\mathbf{f}\to\mathbf{f}$ followed by transform projection
  on $\mathbf{x}$ can be constructed in exactly the same way;
  it looks, however, much less attractive.
  For isomorphic $\mathbf{f}$-space and $\mathbf{x}$-space (e.g. considered in
  Section \ref{DynamicEqPsi} above) the projection retains the full basis,
  thus $\mathbf{f}$ on $\mathbf{x}$ and $\mathbf{x}$ on $\mathbf{f}$
  inferences produce evolution operators ${\mathcal U}$
  in (\ref{dyneqpsi}) different only in time inverse.
  A promising direction for future research may be to consider two
  unitary transformation: ${\mathcal U}^{\mathbf{x}}$ acting
  $\mathbf{x}\to\mathbf{x}$
  and ${\mathcal U}^{\mathbf{f}}$ acting
  $\mathbf{f}\to\mathbf{f}$
  then do transforms projection, see Appendix \ref{Projection2spaces} below.
  }.
Let us apply it
to a vector--to--vector classification problem of Section \ref{ErrorF}.
Assume we have a problem with vector--valued class label
(\ref{mlproblemVector})
\begin{align}
  \mathbf{x}^{(l)}&\to\mathbf{f}^{(l)}
  & \text{weight $\omega^{(l)}$}; l=1\dots  M  \label{mlproblemVectorApp}
\end{align}
The choice of knowledge representation is the most important feature
of a ML approach. For example it can be a linear regression (\ref{fjregression}),
a ratio of two quadratic forms
(\ref{RNfsolutionVector})
or (\ref{RNWfsolutionVector}),
neural network weights, etc.
An important result of this appendix is to consider
not $\mathbf{x} \to \mathbf{f}$ mapping,
but instead to construct localized wavefunctions (\ref{psiYlocalized})
in $\mathbf{x}$- and $\mathbf{f}$- space:
$\psi_{\mathbf{y}}(\mathbf{x})$ and
$\psi_{\mathbf{g}}(\mathbf{f})$
to study $\psi_{\mathbf{y}}(\mathbf{x})$
mapping with a unitary operator ${\mathcal U}$ in $\mathbf{x}$-space
following by a projection of the transform
$\Ket{{\mathcal U} |\psi_{\mathbf{y}}}$
on $\mathbf{f}$-space outcome
$\psi_{\mathbf{g}}(\mathbf{f})$:
\begin{align}
&  \mathrm{Prob}(\mathbf{g}|\mathbf{y})=
  \left|\Braket{\psi_{\mathbf{g}}|{\mathcal U} |\psi_{\mathbf{y}}}\right|^2
  &1\ge\varpi(\mathbf{g})\ge\mathrm{Prob}(\mathbf{g}|\mathbf{y})
\label{vpierr}\\
&{\mathcal F}
    =\sum\limits_{l=1}^{M}
    \omega^{(l)}
    \left|\Braket{\psi_{\mathbf{f}^{(l)}}|{\mathcal U} |\psi_{\mathbf{x}^{(l)}}}\right|^2
    =\sum\limits_{l=1}^{M}
    \omega^{(l)}
    \mathrm{Prob}(\mathbf{f}^{(l)}|\mathbf{x}^{(l)})
          \label{mappingAppendix} \\
&   \mathrm{Error}=\Braket{1}-{\mathcal F}
  \label{fxerrorWProjectionAppendix}
\end{align}
Conditional probability (\ref{vpierr}) is
bounded by the value $\varpi(\mathbf{g})$
of full basis expansion (\ref{fxerrorWOnePointFullBasis}),
a situation without predictor available, this is the problem 
 we considered in Section \ref{ErrorF} above.
Because $\mathbf{x}$- and $\mathbf{f}$- space are different --
a projection of a wavefunction from one to another
gives
$1\ge\varpi(\mathbf{g})\ge\left|\Braket{\psi_{\mathbf{g}}|{\mathcal U} |\psi_{\mathbf{y}}}\right|^2$
in  (\ref{vpierr}).
This non--unitarity, however, does not create any practical difficulties
as we separated a ``unitary dynamics'' in $\mathbf{x}$-space
and a ``non--unitary projection'' to  $\mathbf{f}$-space.
The (\ref{fxerrorWProjectionAppendix}) error estimator
has the meaning of misclassified observations number,
it is bounded by 
considered above simple projective
estimator (\ref{fxerrorW});
it is zero if $\mathbf{f}$ is a subspace of $\mathbf{x}$
(in (\ref{sbasisX}) below consider $\Psi$ as a direct sum of $\Phi$
and the space orthogonal to $\Phi$, then in (\ref{probgypkAppSD})
numerator cancels denominator).

Given the expressions (\ref{psiYlocalized}) for $\psi_{\mathbf{y}}(\mathbf{x})$
and for $\psi_{\mathbf{g}}(\mathbf{f})$:
\begin{align}
  \psi_{\mathbf{g}}(\mathbf{f})&=
 \frac{\sum\limits_{j,k=0}^{m-1}g_jG^{\mathbf{f};\,-1}_{jk}f_k}
           {\sqrt{\sum\limits_{j,k=0}^{m-1}g_jG^{\mathbf{f};\,-1}_{jk}g_k}}
  \label{psiGflocalizedAppendix}
\end{align}
here $G^{\mathbf{f};\,-1}_{jk}$ is an inverse
of $G^{\mathbf{f}}_{jk}$ from (\ref{GfW1}),
we can write conditional probability (\ref{vpierr}) as:
\begin{align}
  \mathrm{Prob}(\mathbf{g}|\mathbf{y})&=
  \frac{
 \left|
    \sum\limits_{j,k,p=0}^{n-1}\sum\limits_{j^{\prime},k^{\prime}=0}^{m-1}
    y_jG^{\mathbf{x};\,-1}_{jk} u_{kp} G^{\mathbf{x}\mathbf{f}}_{pj^{\prime}} G^{\mathbf{f};\,-1}_{j^{\prime}k^{\prime}} g_{k^{\prime}}
    \right|^2
  }
       {
         \sum\limits_{j,k=0}^{n-1} y_{j}G^{\mathbf{x};\,-1}_{jk}y_{k}          
         \sum\limits_{j^{\prime},k^{\prime}=0}^{m-1} g_{j^{\prime}}G^{\mathbf{f};\,-1}_{j^{\prime}k^{\prime}}g_{k^{\prime}}          
       } \label{probgypkApp} \\
       \Ket{{\mathcal U}|x_k}&=\sum\limits_{p=0}^{n-1} u_{kp} x_p
       \label{uinxbasisApp}
\end{align}
The expression is very similar to (\ref{projestimation}),
the difference is that instead of $G^{\mathbf{x}\mathbf{f}}_{kj^{\prime}}$
we now have $\mathbf{x}$ transformed by a unitary operator ${\mathcal U}$
as
$\sum_{p=0}^{n-1}u_{kp} G^{\mathbf{x}\mathbf{f}}_{pj^{\prime}}$. This is the key difference:
instead of ``direct projection'' we now have a unitary transformation
and then a projection.
In
\begin{align}
{\mathcal F}&
=\sum\limits_{l=1}^{M}
 \omega^{(l)}
 \mathrm{Prob}(\mathbf{f}^{(l)}|\mathbf{x}^{(l)})
 =\sum\limits_{j,k,p,q=0}^{n-1}
 u_{jk}S_{jk;pq}u^*_{pq} \label{pql}
\end{align}
a Hermitian tensor $S_{jk;pq}$ is  readily obtained from (\ref{probgypkApp})
and (\ref{pql})
with simple algebra.
Thus we reduced  $\mathbf{x} \to \mathbf{f}$ classification problem
to a dynamic problem of finding a unitary matrix maximizing (\ref{pql}),
i.e. the problem considered in Section \ref{NumericalSilutionUnitaryMatrix}!
This is the most general solution to a vector class label
classification problem, it
finds a unitary transformation ${\mathcal U}$ (\ref{uinxbasisApp}),
producing the maximal
coverage in (\ref{pql}).

Note, that unitary operator ${\mathcal U}$ coefficients $u_{kp}$
are defined in (\ref{uinxbasisApp}) in a general, non--orthogonal basis
$x_k$, a one with real symmetric Gram matrix $G_{jk}^{\mathbf{x}}=\Braket{x_jx_k}$.
This makes unitarity constraint
more verbose:
\begin{align}
  G_{pq}^{\mathbf{x}}&=\sum\limits_{j,k=0}^{n-1}u_{pj}G_{jk}^{\mathbf{x}}u^*_{qk}
  \label{verbunitarity}
\end{align}
It is convenient to select orthogonal bases
$\Psi^{[i]}(\mathbf{x})$, $i=0\dots n-1$
and $\Phi^{[j]}(\mathbf{f})$, $j=0\dots m-1$ for input data,
we already did this in Eq. (\ref{projSil})
above:
\begin{align}
  \Psi^{[i]}(\mathbf{x})&=\sum\limits_{k=0}^{n-1}B^{\mathbf{x}}_{ik}x_k & i=0\dots n-1 \label{sbasisX} \\
   s_i^{(l)}&=\Braket{\psi_{\mathbf{x}^{(l)}}|\Psi^{[i]}}=
  \frac{\Psi^{[i]}(\mathbf{x}^{(l)})}
  {\sqrt{\sum\limits_{j=0}^{n-1}
    \left|\Psi^{[j]}(\mathbf{x}^{(l)})\right|^2
  }}
&1=\sum\limits_{i=0}^{n-1}\left|s_i^{(l)}\right|^2
  \nonumber \\
  \delta_{pq}&=\Braket{\Psi^{[p]}|\Psi^{[q]}}=
  \sum\limits_{j,k=0}^{n-1}B^{\mathbf{x}}_{pj}
  G_{jk}^{\mathbf{x}}B^{\mathbf{x}}_{qk} &p,q=0 \dots n-1  \nonumber \\ 
  \Phi^{[i]}(\mathbf{f})&=\sum\limits_{k=0}^{m-1}B^{\mathbf{f}}_{ik}f_k &i=0 \dots m-1
  \label{sbasisF} \\
  d_i^{(l)}&=\Braket{\psi_{\mathbf{f}^{(l)}}|\Phi^{[i]}}=
  \frac{\Phi^{[i]}(\mathbf{f}^{(l)})}
  {\sqrt{\sum\limits_{j=0}^{m-1}
    \left|\Phi^{[j]}(\mathbf{f}^{(l)})\right|^2
  }}
& 1=\sum\limits_{i=0}^{m-1}\left|d_i^{(l)}\right|^2
  \nonumber \\
  \delta_{pq}&=\Braket{\Phi^{[p]}|\Phi^{[q]}}=
  \sum\limits_{j,k=0}^{m-1}B^{\mathbf{f}}_{pj}
  G_{jk}^{\mathbf{f}}B^{\mathbf{f}}_{qk} &p,q=0 \dots m-1 \nonumber
\end{align}
As the solution is gauge--invariant relatively (\ref{gaugeXF})
we can use any basis.
An orthogonal basis choice
is also beneficial for computational complexity:
it takes $O(n)$ instead of $O(n^2)$ to calculate
a quadratic form
$\sum_{j,k=0}^{n-1} y_{j}G^{\mathbf{x};\,-1}_{jk}y_{k}$ in a basis
in which $G_{jk}^{\mathbf{x}}$ is diagonal.
The $\mathrm{Prob}(\Phi|\Psi)$ also takes a much simpler
form:
\begin{align}
  \mathrm{Prob}(\mathbf{f}|\mathbf{x})&=
  \mathrm{Prob}(\Phi|\Psi)=
 \frac{
 \left|
    \sum\limits_{j,k=0}^{n-1}\sum\limits_{i=0}^{m-1}
    \Psi^{[j]} {\mathcal U}_{jk}
    G^{\Psi\Phi}_{ki}
    \Phi^{[i]}
    \right|^2
  }
       {
         \sum\limits_{j=0}^{n-1} \left|\Psi^{[j]}\right|^2
         \sum\limits_{i=0}^{m-1} \left|\Phi^{[i]}\right|^2
       } \label{probgypkAppSD}  \\
G^{\Psi\Phi}_{ki}&=
       \Braket{\Psi^{[k]}\Phi^{[i]}}=
       \sum\limits_{j=0}^{n-1}\sum\limits_{j^{\prime}=0}^{m-1}
       B^{\mathbf{x}}_{kj}
       G^{\mathbf{x}\mathbf{f}}_{jj^{\prime}}
       B^{\mathbf{f}}_{ij^{\prime}} \label{GPsiPhi} \\
       S_{jk;pq}&=
       \sum\limits_{l=1}^{M}
       \omega^{(l)}
         \sum\limits_{r,t=0}^{m-1}
         s_{j}^{(l)} G^{\Psi\Phi}_{kr}d_{r}^{(l)}
           s_{p}^{(l)} G^{\Psi\Phi}_{qt} d_{t}^{(l)}       
       \label{Stensor}
\end{align}
The (\ref{Stensor}) corresponds to (\ref{Sdef})
when put formally
$s_{k}^{(l+1)}= \sum_{j=0}^{m-1}G^{\Psi\Phi}_{kj}d_{j}^{(l)}$
and swap tensor indexes (inverse time):
$\overset{\looparrowleft}{S}_{jk;pq}=S_{kj;qp}$.
A unitary operator ${\mathcal U}$ now has
a matrix  ${\mathcal U}_{jk}$ with
regular unitarity constraint (\ref{optimmatrixConstraintAppendix}).
As the result is basis--independent
it is practically convenient
to use input data
$x_k^{(l)}$ and $f_j^{(l)}$
to calculate the matrices (\ref{GfW1}) and (\ref{Gx}),
then build from them the
 bases (\ref{sbasisX})
and (\ref{sbasisF}), with possible regularization
of the Appendix \ref{regularization},
then finally use $\Psi^{[k]}(\mathbf{x}^{(l)})$ and $\Phi^{[j]}(\mathbf{f}^{(l)})$
as they were input data sample.
In new bases
the problem with Hermitian tensor (\ref{Stensor})
can be directly approached by (\ref{optimmatrixAppendix})
optimization with unitary constraint (\ref{optimmatrixConstraintAppendix}).
Obtained solution is independent on bases
$\Psi^{[k]}$ and $\Phi^{[j]}$
specific choice (gauge--invariant).
If contributing subspace is known explicitly
the solution of dimension $n$ can be reduced to $m$
using clustering approach (\ref{psicontributing}) of Appendix
\ref{ProjectiveDynamicsClustering} below;
there is also a general $D$-clusters solution
corresponding to a more general ``partial unitarity $D\le n$'' form
 of constraint (\ref{constraintNU}).

What is the main application of the approach of this appendix?
Most often -- it is a ``replacement'' of a regression in a problem
of recovering some hidden $\mathbf{x} \to \mathbf{f}$ relation.
Both theories take (\ref{mlproblemVectorApp}) data as input
and have zero error if $\mathbf{f}$ is a subspace of $\mathbf{x}$.
The differences can be summarized in the table:
\begin{center}
\begin{tabular}{||>{\raggedright}p{3cm}|p{5.9cm}|p{6.5cm}||}
\hline\hline
& Regression &   ``Dynamic'' theory \\[0.5ex]
\hline
The Result &
Function value
$\mathbf{f}(\mathbf{x})$ (\ref{fjregression});
diverges at $\mathbf{x}\to\infty$&
Conditional probability
$\mathrm{Prob}(\mathbf{f}|\mathbf{x})$ (\ref{probgypkAppSD});
does not diverge at $\mathbf{x}\to\infty$
\\[1ex]
Optimization &
$L^2$ norm (\ref{norm2regrf}) in $\mathbf{f}$-space&
The number of correctly classified
observations
(\ref{mappingAppendix}) \\
Mathematical
problem & Linear system solution
& Conditional optimization
(\ref{optimmatrixAppendix})
with unitary constraint (\ref{optimmatrixConstraintAppendix})\\
Outliers and fat tail sensitivity &
Very sensitive; a single ``several orders off'' outlier
completely invalidates the solution
&
Not sensitive;
a single outlier
may invalidate only a single observation point
\\
Symmetry $\psi\to - \psi$ &
Broken: observable is linear on $\mathbf{x}$;
$\psi$ is also linear on $\mathbf{x}$.
& Preserved:
$\psi$ is linear on $\mathbf{x}$,
but the probability (\ref{vpierr}) behaves as $\psi^2$,
invariant with:
$\psi_{\mathbf{x}}\to - \psi_{\mathbf{x}}$;
$\psi_{\mathbf{f}}\to - \psi_{\mathbf{f}}$
\\[0.5ex]
Physical world relation&
A model &
Most of dynamic equations in nature
are equivalent to a sequence of unitary transformations
(Newton, Maxwell, Schr\"{o}dinger equations) \\
\hline\hline
\end{tabular}
\end{center}

\section{\label{Projection2spaces}A Projective Non--Unitary Dynamics}
Considered in Section \ref{nonunitarydynamics} projective
dynamics consists in
a unitary transformation of $\mathbf{x}$ following by a projection
of the transform
on $\mathbf{f}$. The problem can be further generalized.
Consider input data (\ref{mlproblemVectorApp})
as vector spaces $\mathbf{x}$ and $\mathbf{f}$
(it is convenient to convert them
to $\Psi$ and $\Phi$ 
of Eqs.  (\ref{sbasisX}) and (\ref{sbasisF})).
The $\Psi$ and $\Phi$
are regular vector spaces of the dimensions $n$ and $m$
with a scalar product determined by
positively defined (otherwise apply Appendix \ref{regularization} regularization)
matrices (\ref{GfW1}) and (\ref{Gx})
calculated from the data sample (\ref{mlproblemVectorApp}).
In addition we have a ``cross--product'' $\Braket{\Psi|\Phi}$
  (\ref{GPsiPhi})
determined by the matrix $G^{\mathbf{x}\mathbf{f}}_{jk^{\prime}}$
(\ref{Gxf})
calculated from the same data sample.
These bases may not be full with respect to each other:
\begin{subequations}
  \label{psiphicomplete}
\begin{align}
  1&\ge\sum\limits_{j=0}^{m-1}\Braket{\Psi^{[i]}\Phi^{[j]}}^2 & i=0\dots n-1\\
  1&\ge\sum\limits_{j=0}^{n-1}\Braket{\Psi^{[j]}\Phi^{[i]}}^2 & i=0\dots m-1
\end{align}
\end{subequations}
In Section \ref{AttribsVectorF} we considered an approach
of various $\Psi \leftrightarrow \Phi$ projections.
In Appendix \ref{nonunitarydynamics}
we considered a composition of a unitary transformation ${\mathcal U}^{\Psi}$
$\Psi\to\Psi$ following by a projection
of the transform on $\Phi$.
In this appendix we consider the most general case, a composition of:
\begin{enumerate}
\item A $\Psi\to\Psi$ unitary transformation ${\mathcal U}^{\Psi}$,
  the transform is $\Ket{{\mathcal U}^{\Psi}|\Psi}$.
\item A $\Phi\to\Phi$ unitary transformation ${\mathcal U}^{\Phi}$,
  the transform is  $\Ket{{\mathcal U}^{\Phi}|\Phi}$.
\item Projection of these two transforms on each other: $\Braket{\Phi|{\mathcal U}^{\Phi}|{\mathcal U}^{\Psi}|\Psi}$
  using (\ref{GPsiPhi}) ``scalar product''.
\end{enumerate}
The number of ``covered'' observations is then:
\begin{align}
  \mathrm{Prob}(\mathbf{f}|\mathbf{x})&=
  \mathrm{Prob}(\Phi|\Psi)=
  \left|\Braket{\Phi|{\mathcal U}^{\Phi}|{\mathcal U}^{\Psi}|\Psi}\right|^2
\label{vpierrXFProj}\\
{\mathcal F}&
    =\sum\limits_{l=1}^{M}
    \omega^{(l)}
    \left|\Braket{\Phi_{\mathbf{f}^{(l)}}|{\mathcal U}^{\Phi}|{\mathcal U}^{\Psi}|\Psi_{\mathbf{x}^{(l)}}}\right|^2
    =\sum\limits_{l=1}^{M}
    \omega^{(l)}
    \mathrm{Prob}(\mathbf{f}^{(l)}|\mathbf{x}^{(l)})
    \label{mappingAppendixXFOProj}
\end{align}
These expressions are different from (\ref{vpierr}) and (\ref{mappingAppendix})
in a second unitary transformation $\|{\mathcal U}^{\Phi}\|$.
The problem is then:
Maximize (\ref{mappingAppendixXFOProj})
over ${\mathcal U}^{\Psi}_{jk}$ and ${\mathcal U}^{\Phi}_{jk}$
given \textsl{two}
unitary constraints:
\begin{subequations}
  \label{unitaryconstraints2}
  \begin{align}
\delta_{jk}&=
    \sum\limits_{i=0}^{n-1}{\mathcal U}^{\Psi}_{ji} {\mathcal U}^{\Psi\,*}_{ki}& j,k=0\dots n-1 \\
\delta_{jk}&=
\sum\limits_{i=0}^{m-1}{\mathcal U}^{\Phi}_{ji} {\mathcal U}^{\Phi\,*}_{ki}& j,k=0\dots m-1
  \end{align}
\end{subequations}
The optimization (\ref{mappingAppendixXFOProj})
with the constraints (\ref{unitaryconstraints2})
can be approached by Appendix \ref{NumericalSilutionUnitaryMatrix}
type of algorithm, however, as (\ref{mappingAppendixXFOProj})
is a quadratic form over matrix elements products ${\mathcal U}^{\Psi}_{jk}{\mathcal U}^{\Phi}_{qp}$
 (a ``two--particle'' system wavefunction basis
is a product of individual particles wavefunction), this makes the problem of dimensions product,
thus makes it impractical.
We expect that a heuristic algorithm, such as
alternately
optimize (\ref{mappingAppendixXFOProj})
over ${\mathcal U}^{\Psi}_{jk}$ and ${\mathcal U}^{\Phi}_{qp}$,
can be a better fit.
For isomorphic $\mathbf{f}$-space and $\mathbf{x}$-space
($n=m$ and all coefficients in (\ref{psiphicomplete}) are equal to $1$)
the dynamics is unitary and 
the problem itself becomes degenerated: It then depends on a single operator
$\|{\mathcal U}\|=\|{\mathcal U}^{\Psi}|{\mathcal U}^{\Phi}\|$
what is equivalent to the problem already considered
in Section \ref{DynamicEqPsi}.
This makes us to conclude that
considered in Section \ref{nonunitarydynamics} composition:
a unitary transformation of $\Psi$
following by a projection
of the transform on $\Phi$
is the most practical approach
to traditional ML
classification problem $\mathbf{x}\to\mathbf{f}$.

\section{\label{ProjectiveDynamicsClustering}On Clustering of a Dynamic System Phase Space}
In Appendix \ref{nonunitarydynamics}
a ``projective'' solution to dynamic system identification problem
has been developed.
The solution has the form of a unitary operator $\|{\mathcal U}\|$ in $\mathbf{x}$-space.
Conditional
probability given possible input/output
is determined by  (\ref{vpierr}) projection
of $\mathbf{x}$ vector transform
to a vector in $\mathbf{f}$-space. The dimension
of $\mathbf{x}$-space and  $\mathbf{f}$-space
can be quite different.
The $n$ is typically of hundreds, often thousands,
for a system with internal state (memory), see Appendix \ref{withinsernalstate} below,
it may reach millions. The $m$ is the dimension of $\mathbf{f}$,
the number of values of interest,
it is always below a few dozen.
From this relation
naturally arises
the problem of clustering:
to construct
a low dimension $D<n$ subspace of phase space $\mathbf{x}$
that captures most of the information about $\mathbf{f}$.
For a problem with vector class label
only the case $D=m$ is easy.

Consider some orthogonal basis $\Ket{\psi^{[i]}}$
in $\mathbf{x}$-space and expand $\mathbf{x}^{(l)}$-localized states $\psi_{\mathbf{x}^{(l)}}(\mathbf{x})$
in this basis:
\begin{align}
  \Ket{\psi_{\mathbf{x}^{(l)}}}&=
  \sum\limits_{i=0}^{n-1} \Braket{\psi_{\mathbf{x}^{(l)}}|\psi^{[i]}}\Ket{\psi^{[i]}}
  \label{psiXlfullbasisexpansionCluster}
\end{align}
then substitute to (\ref{mappingAppendix}), obtain
the number of covered observations:
\begin{align}
{\mathcal F}&
    =\sum\limits_{l=1}^{M}
    \omega^{(l)}
    \left|\Braket{\psi_{\mathbf{f}^{(l)}}|{\mathcal U} |\psi_{\mathbf{x}^{(l)}}}\right|^2
    \nonumber \\
    &=
    \sum\limits_{l=1}^{M}
    \omega^{(l)}
    \sum\limits_{i,j=0}^{n-1}
    \Braket{\psi^{[i]}|\psi_{\mathbf{x}^{(l)}}}
    \Braket{\psi^{[i]}|{\mathcal U}^{\dagger} |\psi_{\mathbf{f}^{(l)}}}
    \Braket{\psi_{\mathbf{f}^{(l)}}|{\mathcal U} |\psi^{[j]}}\Braket{\psi_{\mathbf{x}^{(l)}}|\psi^{[j]}}
    \label{mappingAppendixExpansion} 
\end{align}
Were we operate in terms of simple ``projective
paradigm'' of Section \ref{ErrorF}
this would correspond to
(\ref{ErrorKfsum}) error with
(\ref{ErrorKfsumSpectral}) spectral expansion.
Now, however, the problem is that sought basis $\Ket{\psi^{[i]}}$ enters
(\ref{mappingAppendixExpansion}) coverage  \textsl{four} times,
thus a direct eigenvalues expansion is no longer possible.
As the conditional probablities are 
 bounded 
(\ref{vpierr})
by direct projection to the entire $\mathbf{x}$-space
 by probabilities (\ref{fxerrorWOnePointFullBasis}),
obtain ${\mathcal F}$ upper bound:
\begin{align}
  {\mathcal F}^{DP}&=\sum\limits_{l=1}^{M}
 \omega^{(l)}
  \varpi(\mathbf{f}^{(l)})
   & {\mathcal F}\le {\mathcal F}^{DP}
  \label{dpfcompare}
\end{align}
The spectral expansion (\ref{ErrorKfsumSpectral}) 
has at most $m$ eigenvectors (\ref{GEVKftoXfxf})
contributing to coverage expansion with $\Ket{\psi_{\mathbf{f}^{(l)}}}$,
for (\ref{mappingAppendixExpansion}) this means
that
only these  $\Ket{\phi}$
contribute to coverage:
\begin{align}
  \Ket{\psi^{[i]}} &\in \Ket{{\mathcal U}|\phi} 
  \label{specI}
\end{align}
where $\Ket{\psi^{[i]}}$ belongs to (\ref{GEVKftoXfxf}) eigenvectors subset
having non--zero eigenvalue, there are at most $m$ out of total $n$.
From this follows that only vector space $\Ket{\phi^{[i]}}$
contribute:
\begin{align}
  \Ket{\phi^{[i]}}&= \Ket{{\mathcal U}^{\dagger}|\psi^{[i]}}
  \label{psicontributing}
\end{align}
where $i$ takes $m$ out of $n$ values such that $\lambda^{[i]}>0$ in (\ref{GEVKftoXfxf}).
The $\Ket{\phi^{[i]}}$ 
is the only $\mathbf{x}$-subspace contributing
to total coverage (\ref{mappingAppendixExpansion}).

Appendix \ref{NumericalSilutionUnitaryMatrix}
solution to maximization (\ref{mappingAppendixExpansion}) 
(which is a quality criterion)
finds unitary matrix $\|{\mathcal U}\|$ in $\mathbf{x}$-space of the dimension $n$.
However, as
quality criterion operates in
$\mathbf{f}$-space of the dimension $m$,
the transform (\ref{psicontributing})
allows to build $\mathbf{x}$-subspace of
the dimension $D=m$ as the only vector subspace contributing to quality criterion.

For a system with known contributing subspace
numerical optimization algorithm of
Appendix \ref{NumericalSilutionUnitaryMatrix}
can be optimized by
converting the basis to contributing subspace
and simplifying the constraints
to act in contributing subspace only,
 i.e. considering a subset of a full set of unitarity constraints.
The conversion back
from contributing subspace to
$\mathbf{x}$-space
then requires some algebra as the condition 
for unitary operators;
${\mathcal U}^{-1}={\mathcal U}^{\dagger}$  may no longer hold true
in full $\mathbf{x}$-space.

In practice
the problem of finding the contributing subspace  (\ref{GEVKftoXfxf})
is typically ``an extra step'',
thus it is sometimes more convenient to solve the problem directly
to avoid a non-unitary transformation between
contributing subspace and $\mathbf{x}$-space.
Whereas constructing a $\mathbf{f}$-predictor of given input dimension $D\le n$
creates the same problem as with (\ref{mappingAppendixExpansion})
(an expression with the fourth power of sought basis),
the problem of finding $\mathbf{x}$ subspace 
of the dimension $D\le n$ providing maximal coverage on $\mathbf{f}$,
can be directly reduced to a variant of Appendix \ref{NumericalSilutionUnitaryMatrix}
optimization problem.

Consider coverage maximization problem with constraints:
\begin{align}
  {\mathcal F}&=
  \sum\limits_{l=1}^{M}\omega^{(l)}
  \sum\limits_{j=0}^{D-1}
  \Braket{\psi_{\mathbf{f}^{(l)}}|\phi^{[j]}}^2
  \xrightarrow[{\phi}]{\quad }\max \label{uconstrNUcontributingSubspace}\\
  \delta_{jk}&=\Braket{\phi^{[j]}|\phi^{[k]}}  & j,k=0\dots D-1
  \label{constraintNU}
\end{align}
the goal is to find an orthogonal basis $\phi^{[j]}(\mathbf{x})$
of  dimension $D\le n$, $j=0\dots D-1$, providing maximal (\ref{uconstrNUcontributingSubspace})
coverage; the solution is non-unique,
it is (\ref{GEVKftoXfxf}) eigenvectors,
corresponding to $D$ largest eigenvalues
within an arbitrary unitary transformation of them.
The problem 
(\ref{fxerrorWOnePointFullBasis})
of above corresponds to  $D=n$ case;
(\ref{dpfcompare}) is the upper bound of (\ref{uconstrNUcontributingSubspace}).
Here $\psi_{\mathbf{g}}(\mathbf{f})$
is $\mathbf{f}=\mathbf{g}$ localized state
(\ref{psiGflocalizedAppendix})
in $\mathbf{f}$-space, and
$\phi^{[j]}(\mathbf{x})$ is $\mathbf{x}$-space linear function:
\begin{align}
  \phi^{[j]}(\mathbf{x})&=\sum\limits_{k=0}^{n-1}u_{jk}x_k & j=0\dots D-1
  \label{phisConstrainNU}
\end{align}
Substituting (\ref{phisConstrainNU}) to (\ref{uconstrNUcontributingSubspace})
obtain optimization problem with some $S_{jk;j^{\prime}k^{\prime}}$:
\begin{align}
    {\mathcal F}=\sum\limits_{j,j^{\prime}=0}^{D-1}\sum\limits_{k,k^{\prime}=0}^{n-1}
    u_{jk}S_{jk;j^{\prime}k^{\prime}}u^*_{j^{\prime}k^{\prime}}
    &\xrightarrow[u]{\quad }\max  \label{optimmatrixAppendixNU}\\
\sum\limits_{k,k^{\prime}=0}^{n-1}u_{jk}G^{\mathbf{x}}_{kk^{\prime}} u^*_{ik^{\prime}}&=\delta_{ji} & j,i=0\dots D-1
    \label{optimmatrixConstraintAppendixNU}
\end{align}
The problem: \textsl{to find $u_{jk}$ matrix
  of the dimensions $j=0\dots D-1,k=0\dots n-1$,
  providing maximal} (\ref{optimmatrixAppendixNU})
\textsl{subject to constraint} (\ref{optimmatrixConstraintAppendixNU}).
Obtained $u_{jk}$ matrix
defines
$\phi^{[j]}(\mathbf{x})$ basis (\ref{phisConstrainNU})
of the dimension $D\le n$
providing maximal coverage in (\ref{uconstrNUcontributingSubspace}).
This basis is then typically used
to construct in it a  unitary operator ${\mathcal U}$
providing
maximal coverage in (\ref{mappingAppendix}). 
Thus we need to solve \textbf{two} optimization problems: first (\ref{uconstrNUcontributingSubspace})
to construct a basis of lower dimension,
second (\ref{mappingAppendix}) to build a unitary operator in this basis.
If $D=m$ and $\mathbf{f}$ is a subspace of $\mathbf{x}$
then the sought basis is this
subspace and coverage is maximal ${\mathcal F}=\Braket{1}$.
Otherwise we modify Appendix \ref{NumericalSilutionUnitaryMatrix}
algorithm to $D\le n$ case, specifically:

\subsection{\label{NumericalSilutionUnitaryMatrixNUDlen}A Numerical Solution to
  Quadratic Form Maximization Problem
With Partial Unitarity Constraint}
Without loss of generality
let $G^{\mathbf{x}}_{kk^{\prime}}=\delta_{kk^{\prime}}$,
i.e. the problem is considered in bases (\ref{sbasisX}) and (\ref{sbasisF}).
Optimization problem is then:
\begin{align}
    {\mathcal F}=\sum\limits_{j,j^{\prime}=0}^{D-1}\sum\limits_{k,k^{\prime}=0}^{n-1}
    u_{jk}S_{jk;j^{\prime}k^{\prime}}u^*_{j^{\prime}k^{\prime}}
    &\xrightarrow[u]{\quad }\max  \label{optimmatrixAppendixNUDlen}\\
\sum\limits_{k=0}^{n-1}u_{jk} u^*_{ik}&=\delta_{ji} & j,i=0\dots D-1
    \label{optimmatrixConstraintAppendixNUDlen}
\end{align}
Consider Lagrange multipliers $\lambda_{jj^{\prime}}$, a matrix
of $D\times D$ dimension,
to optimize (\ref{optimmatrixAppendixNUDlen})
with the constraints (\ref{optimmatrixConstraintAppendixNUDlen})
\begin{align}
  &
  \sum\limits_{j,j^{\prime}=0}^{D-1}\sum\limits_{k,k^{\prime}=0}^{n-1}
             u_{jk}S_{jk;j^{\prime}k^{\prime}}u^*_{j^{\prime}k^{\prime}}
             +
   \sum\limits_{j,j^{\prime}=0}^{D-1}        
   \lambda_{jj^{\prime}}\left[\delta_{jj^{\prime}}-\sum\limits_{k^{\prime}=0}^{n-1}u_{jk^{\prime}} u^*_{j^{\prime}k^{\prime}} \right]
   \xrightarrow[u]{\quad }\max
   \label{lagrangetovariateNUDlen}
\end{align}
The variations are consistent only when $\lambda_{jj^{\prime}}$
is a Hermitian matrix.
The ``partial'' constraint is the squared Frobenius norm
condition:
\begin{align}
    &\sum\limits_{j=0}^{D-1}\sum\limits_{k=0}^{n-1}u_{jk} u^*_{jk}=D
    \label{optimmatrixConstraintScalarAppendixNUDlen}
\end{align}
with which (\ref{optimmatrixAppendixNUDlen}) optimization
can be reduced to a generalized eigenvalue problem.
Then repeat Appendix \ref{NumericalSilutionUnitaryMatrix}
iteration almost identically.
Generalized eigenvalue problem of the dimension $Dn$ is solved
with partial constraint (\ref{optimmatrixConstraintScalarAppendixNUDlen})
being wavefunction normalizing condition;
obtained with partiall constrained solution $u_{jk}$ 
requires an adjustment
to satisfy full constraint (\ref{optimmatrixConstraintAppendixNUDlen});
it is performed using
\href{https://en.wikipedia.org/wiki/Singular_value_decomposition}{SVD}
expansion:
  \begin{align}
    u_{jk}&=\sum\limits_{j^{\prime}=0}^{D-1}\sum\limits_{k^{\prime}=0}^{n-1}
    U_{jj^{\prime}}\Sigma_{j^{\prime}k^{\prime}}V^{\dagger}_{k^{\prime}k}
    \label{SVDNUDlen}
  \end{align}
followed by setting diagonal elements
of the rectangular diagonal matrix $\Sigma_{jk}$ to $1$;
new values for Lagrange multipliers $\lambda_{jj^{\prime}}$
are then calculated from adjusted $u_{jk}$ to perform a new iteration.
With these changes to Appendix \ref{NumericalSilutionUnitaryMatrix}
algorithm the iterational process produces $u_{jk}$ matrix
maximizing (\ref{optimmatrixAppendixNUDlen})
subject to partial unitarity  $D\le n$ constraint (\ref{optimmatrixConstraintAppendixNUDlen}).

\section{\label{withinsernalstate} The Dynamics of a System with Internal State}
The data (\ref{mlproblemVectorApp}) $\mathbf{x}^{(l)}\to\mathbf{f}^{(l)}$
is the form most frequently studied in ML, where
observations corresponding to different $l$ are considered
as independent observations.
Same data studied in signal processing
is typically considered as
 $l$--ordered (e.g. $l$ is time),
where the problem of timeserie prediction
corresponds to
$\mathbf{f}^{(l)}=\mathbf{x}^{(l+1)}$.
Such an embedding of timeserie data
to (\ref{mlproblemVectorApp})
implicitly selects a time--scale.
Real system have some internal state $\mathbf{z}$ (memory);
the output now depends not only on the input signals $\mathbf{x}$,
but also on the internal state $\mathbf{z}$:
\begin{align}
  \left(\mathbf{x}^{(l)},\mathbf{z}^{(l)}\right)&\to\mathbf{f}^{(l)}
  & \text{weight $\omega^{(l)}$}; l=1\dots  M  \label{mlproblemVectorMemory}
\end{align}
This produces a omnifarious dynamics, much richer
compared to systems without internal state.
An example of a system with memory
is a
\href{https://en.wikipedia.org/wiki/Finite-state_machine}{finite-state machine}.
From practical point of view it is convenient to classify
them as the systems with:
\begin{itemize}
\item  Completely observable internal state.
\item Partially observable internal state.
\end{itemize}
The same system (e.g. a vending machine)
can be completely observable to
a support team (have a full access to vending machine memory)
and partially observable to a customer (can only see whether
it is empty and not working).
In this appendix we will be only considering
the systems with completely observable internal state.

Consider a very simple finite-state machine:
\href{https://en.wikipedia.org/wiki/Flip-flop_(electronics)#Classical_positive-edge-triggered_D_flip-flop}{synchronous positive-edge-triggered D flip-flop} (D trigger);
it's circuit has
a positive feedback loop what creates
a bistable system. 
\href{https://www.ti.com/lit/ds/symlink/cd4013b.pdf}{CD4013}  chip
is a typical example of this device.
\tikzset{
  flipflop D/.style={flipflop, flipflop def={t1=D, t6=Q, c3=1, t3=C, t4=\ctikztextnot{Q}}}
  }
\begin{align}
  &\text{
\begin{circuitikz}
\draw 
  (0,0) node[flipflop D]{}
  ;
\end{circuitikz}
  }
\end{align}
It operates as following: on every $0\to 1$ transition on $C$
(on the rising edge \texttiming{LH} of the clock)
input $D$
is recorded and becomes immediately available on $Q$, the
$\overline{Q}$ is it's inverse.
Any changes on $D$ has no effect on the state unless
there is a rising edge on $C$:
\begin{align}
  &\text{
    \begin{tabular}{|c|c|c|}
      \hline
      \makebox[1cm][c]{C} & \makebox[1cm][c]{D} & Q \\
      \hline
      \texttiming{LH}
       & 0 & 0 \\
      \hline
      \texttiming{LH} & 1 & 1 \\
      \hline
      \makecell{0\\ 1\\
        \texttiming{HL}
      }& X & unchanged \\
      \hline
    \end{tabular}      
  }
  \label{transitionTable}
\end{align}
This device can be used as a 1-bit memory register,
pulses counter,
frequency divider by 2 (connect $D$ with $\overline{Q}$
to inverse the state on every \texttiming{LH} on $C$), etc.

Consider a simple problem of the dimensions $n=2$, $m=1$.
Take edge--triggered D flip-flop,
let $x_0=D$, $x_1=C$,
and output $f=Q$.
Also assume (to avoid
\href{https://en.wikipedia.org/wiki/Flip-flop_(electronics)#Timing_considerations}{timing considerations})
that on every tick $l$
the
$x_1^{(l)}$ takes the value slightly after $x_0^{(l)}$ was set.
The output $Q$ at $l$ now depends not only
on current input
$\mathbf{x}^{(l)}$ but also
on the previous state (and hence, previous inputs).
Now assume that all the input $\mathbf{x}^{(l)}$ are completely random.
For every new $l$--th input $\mathbf{x}^{(l)}$ coming (completely random) the system undergo transition:
\begin{align}
  f^{(l)}&=
  \begin{cases}
    x_0^{(l)}& \text{if $x_1^{(l-1)}=0$ and $x_1^{(l)}=1$} \\
    f^{(l-1)}& \text{otherwise}
  \end{cases}
  \label{dtriggereq}
\end{align}
It is clear that this D-trigger cannot be predicted by $n=2$, $m=1$
system corresponding to $D$, $C$, $Q$ trigger terminals
``connected'' to $x_0$, $x_1$ and $f$.
A system with (\ref{dtriggereq}) transition rules has a long--range dynamics\footnote{
  A more straightforward example of a system
  with long--range dynamics
is the aforementioned frequency divider by 2 (connect $D$ with $\overline{Q}$) and use $\mathbf{x}=C$, $\mathbf{f}=Q$;
this single input system
switches the state to the inverted
$f^{(l+1)}=\overline{f^{(l)}}$
for every $x_0^{(l)}=1$ such that $x_0^{(l-1)}=0$; this system has the state
completely determined by the initial state
and the number of \texttiming{LH} transition on $C$ input.}.

A typical result of interest for a study of such a system
is: given a long sequence of random $\mathbf{x}^{(l)}$  as input
be able
to tell: there is a D-trigger inside.
It is clear that
an approach
typical for signal processing:
take a finite number of previous
inputs $\mathbf{x}^{(l-1)}$, $\mathbf{x}^{(l-2)}$,
$\mathbf{x}^{(l-3)}$, \dots,
the length is determined by e.g. autocorrelation length of the signal,
is poorly applicable to a system with internal memory.

For a  system with completely observable internal state
the problem can be directly approached
by using $\mathbf{f}$
and
some previous $\mathbf{x}$ (like in signal processing)
as system memory:
put $\mathbf{z}^{(l)}=\left(f^{(l-1)},x_1^{(l-1)}\right)$
in (\ref{mlproblemVectorMemory}),
making
a system of the dimensions $n=4$, $m=1$.
Given this input almost any ML technique can build an accurate
predictor for D-trigger. The problem, however,
is that to apply obtained rules
an information about system current internal state is required
and this information is typically not available.
The approach of Appendix (\ref{nonunitarydynamics})
separates the system dynamics (in a form of unitary operator
$\|{\mathcal U}\|$ obtained from (\ref{mappingAppendix}) optimization)
and calculation of conditional
probability  (\ref{vpierr}) for a given input/output.
When applied to this problem only the first step
is straightforward:
 construct a unitary operator of dimension $4$
in
(\ref{mlproblemVectorMemory}) space
that can be selected as a subspace of
$\left(\mathbf{x}^{(l)},\mathbf{f}^{(l-1)},\mathbf{x}^{(l-1)},\mathbf{x}^{(l-2)},\dots\right)$
the transform then to be projected to $\mathbf{f}^{(l)}$;
the $\mathrm{Error}$ from (\ref{fxerrorWProjectionAppendix})
will be $0$.
However, the second step: it's application to a
prediction of future value of $\mathbf{f}$ is problematic
as the ``system current state'' is typically available
only for training data.
Nevertheless, obtained unitary operator
precisely identifies (\ref{transitionTable}) system dynamics 
and tells us exactly: there is a D-trigger inside!

\section{\label{sysdecoherence}Kraus Operators and State Decoherence Problem}
A dynamics considered so far was of either unitary
or unitary following by a projection forms.
The criterion (\ref{Unormax}) is the total coverage of a system
with an initial state (e.g. a localized pure state
$\Ket{\psi_{\mathbf{x}}}\Bra{\psi_{\mathbf{x}}}$;
it has a simple form in (\ref{sbasisX}) basis),
the initial state is
transformed to predicted state with a unitary transformation (\ref{Uevolution})
\begin{align}
  \|\rho_{\mathbf{x}}\|&=\Ket{\psi_{\mathbf{x}}}\Bra{\psi_{\mathbf{x}}}=
  \sum\limits_{i,k=0}^{n-1}
  \Ket{\Psi^{[i]}}
  \frac{\Psi^{[i]}(\mathbf{x}){\Psi^{[k]}}^{*}(\mathbf{x})}
  {\sum\limits_{j=0}^{n-1} \left|\Psi^{[j]}(\mathbf{x})\right|^2}
    \Bra{\Psi^{[k]}}
  \label{apurestate} \\
\|\widetilde{\rho}_{\mathbf{x}^{(l+1)}}\|&=\|{\mathcal U}|\rho_{\mathbf{x}^{(l)}}|{\mathcal U}^{\dagger}\| \label{Uevolution}
\end{align}
following a comparison of predicted and realized density matrices to obtain
the total coverage
by taking sum over all observations, exactly as we did
in Eq. (\ref{Unormax}) above:
\begin{align}
  {\mathcal F} &
  =\sum_{l=1}^{M}\omega^{(l)}\mathrm{Spur}\|\rho_{\mathbf{x}^{(l+1)}}|{\mathcal U}|\rho_{\mathbf{x}^{(l)}}|{\mathcal U}^{\dagger}\|
  =\sum_{l=1}^{M}\omega^{(l)}\mathrm{Spur}\|\rho_{\mathbf{x}^{(l+1)}}|\widetilde{\rho}_{\mathbf{x}^{(l+1)}}\|
  \label{coberagecomparison} \\
  {\mathcal U}^{\dagger}{\mathcal U}&=\mathds{1} \label{unitarityconstarintdensitymatrix}\\
  \mathrm{Error}&=\Braket{1}-{\mathcal F} \label{errcoverage}
\end{align}
This approach can be successfully applied to
a number of problems, e.g. to a deterministic finite-state machine
such as considered in the Appendix \ref{withinsernalstate} above.

An example of a system to which an application of unitary dynamics
has limitations is the data of
\href{https://en.wikipedia.org/wiki/Markov_chain}{Markov chain}
type.
Consider single boolean variable Markov chain with
a stationary
\href{https://en.wikipedia.org/wiki/Stochastic_matrix}{transition matrix} $P_{yz}$:
\begin{align}
  x^{(l)}&: \{0,1\} &\omega^{(l)}=1; l=1\dots M \label{Markovchain}  \\
  P_{yz}&=P(x^{(l+1)}=z|x^{(l)}=y)  & 1=\sum\limits_{z=0,1}P_{yz} \label{MarkovchainProbability}
\end{align}
For a boolean variable we can assume that
$x=0$ corresponds to $\Ket{\psi^{[0]}}$
and $x=1$ corresponds to $\Ket{\psi^{[1]}}$;
without loss of generality
we can also assume $\Braket{\psi^{[y]}|\psi^{[z]}}=\delta_{yz}$.
For $l=1\dots M$ observations Markov chain  model (\ref{Markovchain})
gives the transition:
if the value of $x^{(l)}$ is known then $x^{(l+1)}$ outcome probabilities can be
predicted according to (\ref{MarkovchainProbability}).
If  $x$  at $l$ is known and equal $x^{(l)}$
then $l \to l+1$ transition of $\Ket{\psi^{[x]}}$ state is:
\begin{align}
  \Ket{\psi^{[x]}}\Bra{\psi^{[x]}}&\to P_{x0}\Ket{\psi^{[0]}}\Bra{\psi^{[0]}}+
  P_{x1}\Ket{\psi^{[1]}}\Bra{\psi^{[1]}}
  \label{Ptransition}
\end{align}
Important, that Markov chain $l\to l+1$ transition
transforms pure state (given we know $x=x^{(l)}$ value)
to a mixed state according to transition matrix probabilities.
This type of transformation cannot be
obtained from unitary dynamics (\ref{Uevolution}).
A fundamental property of quantum dynamics is:
a pure state can be transformed only to a pure state.
Markov chain dynamics (\ref{Ptransition}) is different in
this sense as it possibly transforms pure state to a mixed state.

This problem
is known as
\href{https://en.wikipedia.org/wiki/Quantum_decoherence}{quantum decoherence}
and is a subject of active study\cite{zeh1970interpretation} since 
the inception of quantum theory
initially in application to
\href{https://en.wikipedia.org/wiki/Measurement_problem}{quantum measurement},
following by
quantum computing,
quantum field theory\cite{unruh1995evolution}, etc.; for example
as black hole radiates
as black body (Hawking radiation)
thus it should completely evaporate within
a finite time, and in this process an initially
pure quantum state should evolve to a mixed state\cite{wald1994quantum}.

The problem in hand is much less global. It is:
given the data (\ref{mlproblemVector})
to transform
localized pure state $\psi_{\mathbf{y}}(\mathbf{x})$ from
(\ref{psiYlocalized}) to a mixed state
to be subsequently used e.g. in (\ref{coberagecomparison})
coverage estimation
instead of $\|{\mathcal U}|\rho_{\mathbf{x}^{(l)}}|{\mathcal U}^{\dagger}\|$,
corresponding to regular quantum dynamics (\ref{unitaryPsiEvolution}).

Typically to obtain a mixed state from pure state one may
consider some other space $\Ket{\varphi}$, form
a composite system $\Ket{\varphi}\otimes\Ket{\psi}$,
then consider a pure state in the composite space;
as the $\Ket{\varphi}$ states are not observable
take the $\mathrm{Spur}$ over  $\Ket{\varphi}$ (partial spur)
and obtain a mixed state in $\Ket{\psi}$-space.
The difficulty is that with (\ref{mlproblemVector}) data there is
no other space $\Ket{\varphi}$,
only averaging over $l=1\dots M$ observations is available;
there is no ``second set of observations'' for a given $l$
(with possible exception of distribution regression problem\cite{2015arXiv151109058G} type of data).
For this reason we need other methods to construct
a mixed state.

Mathematically
the problem is equivalent to
constructing a
\href{https://en.wikipedia.org/wiki/Quantum_channel}{completely positive trace-preserving map} (quantum channel).
Considered in Appendix \ref{nonunitarydynamics} above
ML classification problem consists in
a unitary transformation in $\mathbf{x}$-space following by a
projection of the transform to $\mathbf{f}$-space; this is a 
\textsl{trace-decreasing map} (quantum operation)
as these two spaces are not necessary
full with respect to each other.

\href{https://en.wikipedia.org/wiki/Quantum_operation#Kraus_operators}{Kraus' theorem}
determines the most general form of this operation\cite{kraus1983states}:
\begin{align}
  \widetilde{\rho}&=\sum\limits_s B_s\rho B^{\dagger}_s \label{KrausOperator}
\end{align}
with Kraus operators $B_s$ satisfying
\begin{align}
  \sum\limits_s B_s^{\dagger}B_s&=\mathds{1} \label{constraintKrauss}
\end{align}
The number of terms in the $s$-sum is called Kraus rank.
The maximal number of terms is  $n^2$ (or $nm$
for (\ref{KrausOperator}) transformations between Hilbert spaces of different  dimensions),
in ML applications a good heuristic is to choose Kraus rank between $1$ and $3$, a value below $n$ fits most data analysis problems.
The transformation (\ref{KrausOperator}) subject to constraint (\ref{constraintKrauss})
is a generalization of regular quantum dynamics (\ref{Uevolution})
subject to unitary constraint  (\ref{unitarityconstarintdensitymatrix}).
A fundamental question is then: whether 
Appendix \ref{NumericalSilutionUnitaryMatrixNUDlen}
numerical optimization algorithm of a problem with partial unitarity
constraint (\ref{optimmatrixConstraintAppendixNUDlen})
can be modified to approach the problem
of finding Kraus operators $B_s$ maximizing
\begin{align}
  {\mathcal F} &=\sum\limits_{l=1}^{M}\omega^{(l)}\sum\limits_s\mathrm{Spur}\|\rho_{\mathbf{x}^{(l+1)}}|B_s|\rho_{\mathbf{x}^{(l)}}|B_s^{\dagger}\|
  \label{KraussCoberagecomparison}
\end{align}
subject to (\ref{constraintKrauss}) constraint;
the problem solution ``favors'' pure states as only for them
$\mathrm{Spur}\rho^2=1$ and the maximal coverage $\Braket{1}$
can be reached; for a series of mixed state density matrices $\rho_{(l)}$
maximal coverage is limited by the value
$\sum_{l=1}^{M}\omega^{(l)}\mathrm{Spur}\rho_{(l)}^2$,
which reaches $\Braket{1}$ only when all  $\rho_{(l)}$ are pure states.
This optimization problem, same as the one considered in the Appendix \ref{NumericalSilutionUnitaryMatrix}: maximize (\ref{coberagecomparison}) subject to
(\ref{unitarityconstarintdensitymatrix}),
has target function and constraints both
being quadratic functions on Kraus operators $B_s$ matrix elements.
Thus we can consider a ``wavefunction'' (of the dimension $n^2$ times the number of $B_s$ operators in (\ref{constraintKrauss}) sum)
constructed from $B_s$ matrix elements 
subject to ``partial'' constraint (a generalization of (\ref{optimmatrixConstraintScalarAppendix})): the sum of all $B_s$ matrix elements absolute value squared (the sum of all $B_s$ squared Frobenius norm) equals to $n$.
Optimization problem with partial constraint
can be easily solved as equivalent to a regular eigenvalue problem.
An iterational process involving an
update of obtained ``partial constraint'' solution to a full constraint
sub-optimal one
with subsequent
Lagrange multipliers recalculation is then repeated until
the required constraints (\ref{constraintKrauss}) are satisfied in full.
This treatment readily produces a numerical solution.
The solution is non--unique (take e.g. a permutation of $B_s$;
more generally -- $B_s$ are defined within a unitary transformation;
it is often convenient to work with orthogonal form
of Kraus operators (canonical form)
$\mathrm{Spur}B^{\dagger}_sB_t\sim\delta_{st}$,
that is especially useful for adjusting
 ``partial constraint'' solution to a full constraint
sub-optimal one)
but described numerical algorithm (contrary to na\"{\i}ve
\href{https://en.wikipedia.org/wiki/Newton%27s_method#k_variables,_k_functions}{Newtonian type}
  iterations) is expected to be non--sensitive to this degeneracy
 unless the number of terms
in (\ref{constraintKrauss}) sum
is chosen a very large; if there is just a single term in the sum
(Kraus rank one) --
then the problem is reduced to 
previously considered optimization problem (\ref{optimmatrixAppendix})
with unitary constraint (\ref{optimmatrixConstraintAppendix}),
a
\href{https://en.wikipedia.org/wiki/Quantum_channel#Pure_channel}{pure quantum channel}.

\bibliography{LD}

\end{document}